\documentclass[mnsc]{informs4}
\usepackage[T1]{fontenc}
\usepackage{lmodern}
\usepackage{eqndefns-left} 
\usepackage{xurl}
\usepackage{tikz}
\usetikzlibrary{positioning,arrows.meta,calc,fit,backgrounds}
\RequirePackage{tgtermes}
\RequirePackage{newtxtext}
\RequirePackage{newtxmath}
\RequirePackage{fix-cm}
\RequirePackage{endnotes}
\usepackage{tabularx}
\newcolumntype{Y}{>{\raggedright\arraybackslash}X}
\usepackage{booktabs}
\usepackage{array}
\usepackage{mathtools}
\usepackage{pgfplots}
\usepackage{float}  
\usepackage{placeins}
\OneAndAHalfSpacedXII 

\usepackage{algorithm}
\usepackage{algpseudocode}
\usepackage{tikz}
\usepackage{threeparttable}
\usepackage{algorithm}
\usepackage{algpseudocode}
\usepackage{enumerate}
\usepackage{lmodern}
\usepackage{clipboard}
\newclipboard{myclipboard}
\usepackage[T1]{fontenc}
\usepackage[font=small,labelfont=bf]{caption}
\usepackage[subrefformat=parens,labelformat=parens]{subcaption}
\usepackage{multirow}
\pgfplotsset{compat=1.18}
\usepackage{natbib}
 \bibpunct[, ]{(}{)}{,}{a}{}{,}%
 \def\bibfont{\small}%

\newcommand{\cS}{\mathcal{S}}
\newcommand{\cB}{\mathcal{B}}
\newcommand{\cA}{\mathcal{A}}
\newcommand{\cN}{\mathcal{N}}
\newcommand{\cM}{\mathcal{M}}
\definecolor{linkblue}{RGB}{28,82,145}
\definecolor{allocblue}{RGB}{31,78,140}
\definecolor{allocfill}{RGB}{219,231,247}
\definecolor{requestpur}{RGB}{112,48,160}
\definecolor{requestfill}{RGB}{236,224,247}
\definecolor{neutralgray}{RGB}{90,90,90}
\newtcolorbox{myquote}[1][]{%
    code={\OneAndAHalfSpacedXII},
    colback=black!5,
    colframe=black!5,
    notitle,
    sharp corners,
    borderline west={2pt}{0pt}{red!80!black},
    enhanced,
    breakable,
    coltext=blue,
}
\DeclareMathOperator{\E}{\mathbb{E}}

\newif\ifshowrev
\showrevtrue

\makeatletter
\@ifundefined{theARTICLEABSTRACT}{}{%
  \def\theARTICLEABSTRACT{%
    \HOOKb
    \vspace*{18pt}%
    \noindent
    \begin{minipage}[t]{\linewidth}
      \parindent1em
      \ABSfont
      \noindent\theABSTRACT\endgraf
      \vskip5pt
      \theFUNDING
      \theKEYWORDS
      \theSUBJECTCLASS
      \theAREAOFREVIEW
      \theMSCCLASS
      \theORMSCLASS
      \if@BLINDREV\else\theHISTORY\fi
      \noindent\hrulefill
    \end{minipage}%
    \par
    \vspace*{0pt}%
  }%
}
\makeatother
\newif\ifcompileletter \compilelettertrue
\newif\ifcompilemain   \compilemaintrue
\EquationsNumberedThrough    

\TheoremsNumberedThrough     
\ECRepeatTheorems  %
\RequirePackage{bm}
\RequirePackage[
    hypertexnames=false,
    plainpages=false
]{hyperref}
\MANUSCRIPTNO{}

\renewcommand{\theARTICLETOP}{}

\LRHSecondLine{}
\RRHSecondLine{}

\begin{document}


\RUNTITLE{Reinforcement Learning for Online Resource Allocation}

\TITLE{Reinforcement Learning for Online Resource Allocation Against an Adaptive Adversary}
\ARTICLEAUTHORS{%
\AUTHOR{Ziwei Wang}
\AFF{Peking University, \EMAIL{wangzwdl@pku.edu.cn}}

\AUTHOR{Hengzhou Li}
\AFF{Peking University, \EMAIL{lihengzhou@stu.pku.edu.cn
}}

\AUTHOR{Ningyuan Chen}
\AFF{University of Toronto, \EMAIL{ningyuan.chen@utoronto.ca}}

\AUTHOR{Jie Song}
\AFF{Peking University, \EMAIL{jie.song@pku.edu.cn}}
} 

\ABSTRACT{%
We study robust online resource allocation against an adaptive adversary, where future
request may depend on the resource state created by earlier allocation decisions.
The firm commits to an allocation policy, while an adversary chooses admissible request types as the resource state evolves. Observing request types before
allocation is the key structure: it lets the robust problem decompose by type
and makes the learning dynamics triangular across periods. Building on it, we
develop an adversary-in-the-loop reinforcement learning method in which adversary and allocation actors take natural policy steps from a critic fitted once per
epoch and reused. We prove global linear last-iterate convergence to the regularized robust equilibrium, up to an explicit error floor under tabular setting.
We also give sampling conditions for realizing the required accuracy. A deep reinforcement learning implementation with neural network design scales the method.
On personalized assortment and online linear programming, it attains
near-optimal robust revenue on small tabular instances and consistently outperforms
inventory balancing and dual mirror descent on large ones with neural policies.
}




\maketitle


\section{Introduction}
\label{sec:intro}
Online resource allocation concerns how firms allocate limited resources as requests arrive, before future requests are known. Applications include displaying product assortments under inventory constraints, accepting orders against shared capacity in online linear programming, and assigning jobs to available servers. Dating back to early work on airline seat inventory control \citep{littlewood1972forecasting}, the central challenge is to maximize total revenue by balancing current returns against future opportunities. The dominant approach designs an online algorithm and proves its competitive ratio against an adversary that chooses the arrival sequence \citep{mehta2007adwords,ball2009toward,golrezaei2014real,ma2020algorithms,balseiro2023best}. The resulting algorithms, such as protection levels, inventory balancing and primal--dual or dual-price policies, are elegant: they need little computation and no forecast, they apply to every instance, and many of their ratios are tight.

Competitive analysis typically assumes an oblivious adversary, which fixes the arrival sequence in advance and does not react to the resource state that the algorithm creates.
It leaves two questions open for a firm
operating a particular resource system. The first is how to evaluate policies
when allocations affect future arrivals. Stockouts can redirect customer
demand, and information about service delays can affect which customers stay
\citep{li2023estimating,yu2022delay,aksin2017impact}. An adaptive adversary
can respond to past decisions \citep{bendavid1994power}. In our setting, it
observes the resource state those decisions leave behind. An offline benchmark
evaluated on the resulting arrival sequence may then serve requests that its
own earlier decisions would never have induced. For any positive target
competitive ratio, we construct an instance and one adaptive adversary under
which every online policy falls below that target relative to this benchmark
(Proposition~\ref{pro:cr}). Depletion can also cause recovery losses that
grow linearly with the horizon (Proposition~\ref{prop:chase}). 

The second
question is how much a policy designed for the instance can improve on a
general-purpose rule. A tight class-wide ratio leaves room for improvement on
particular instances. Inventory balancing, for example, adjusts each
product's revenue using a prescribed function of its remaining inventory
fraction. A policy designed for the instance can also account for how the
mix of remaining products shapes later opportunities. On our small test
instances, inventory balancing and dual mirror descent achieve \(93.5\%\)
and \(75.0\%\) of the respective robust optima, against $99.9\%$ for our method (Table~\ref{tab:all-results}). Even on the competitive ratio computed for the same instance, which is the baselines' own metric, our policies do better in five of six comparisons.

We seek the highest expected revenue a firm can guarantee for a given
instance when requests may respond to the remaining resources. The instance
specifies the initial resources, feasible allocations, and the request
types available to the adversary in each period and resource state.
Business rules or demand forecasts can narrow those type sets. 
The firm commits to
an online policy that is evaluated against its own worst admissible
adversary policy. 

\begin{figure}
    \centering
    \includegraphics[width=0.8\linewidth]{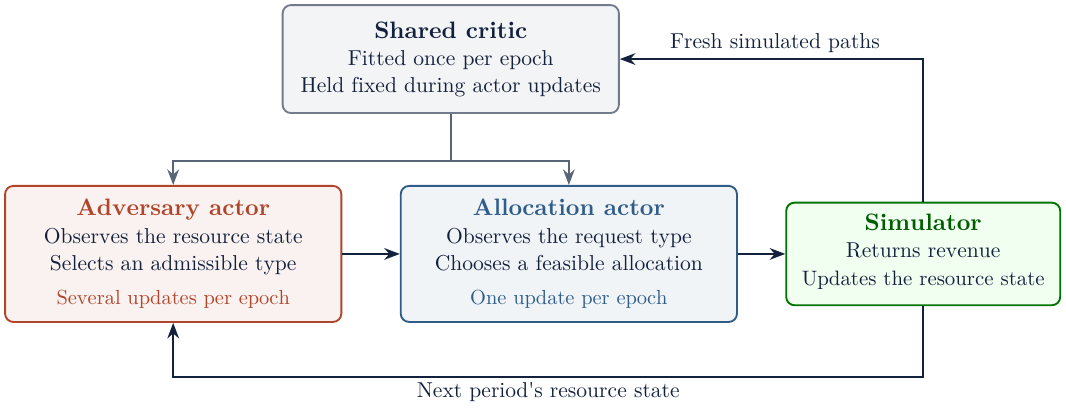}
    \caption{Adversary-in-the-Loop Actor-Lazy Critic Reinforcement Learning Framework}
    \label{fig:actorcritic}
\end{figure}

The timing of online allocation gives this robust problem a useful
computational structure. Our problem can be viewed as a finite horizon zero-sum Markov Stackelberg
game between the firm and an adversary. The firm is the leader: it
commits to an allocation policy that maps each resource state and observed
request type to an allocation. The adversary is the follower: knowing
this policy, it chooses at each resource state a request type from an
admissible set, to
minimize the firm's expected revenue. Within each period, the adversary moves
first and the firm observes the selected type before allocating. The
leader's equilibrium payoff is exactly the revenue the firm can guarantee
against every admissible adaptive adversary. Under rectangular request choices, the game admits an
\emph{s}-rectangular robust MDP representation with type-generated
uncertainty (Proposition~\ref{prop:lifting}). In this representation,
each action is a contingent plan that specifies an allocation for
every admissible request type. The number of deterministic plans can
grow exponentially in the number of types. Because the firm observes
the request type before allocating, the Bellman recursion can
optimize over feasible allocations one type at a time. Backward
induction gives exact solutions on small instances. The number of
resource states can still grow exponentially in the number of
resources, which motivates learning from simulated trajectories. Our main contributions are as follows:

\noindent\textbf{A reinforcement learning framework.}
Robust MDP policy gradient methods treat uncertainty as an adversarial
transition kernel and evaluate it inside each policy update, exactly,
approximately, or through a parameterized kernel with value approximation
\citep{wang2023policygradient,chen2024accelerated,li2025nonrectangular}.
Our uncertainty has more timing structure, so we learn both sides of the game
directly from simulated operating paths, without constructing worst-case
instances by hand (Figure~\ref{fig:actorcritic}). An allocation actor selects
a feasible action given the resource state and observed type, and an adversary
actor selects an admissible type given the resource state; their interaction
generates request paths that respond to earlier allocations. Because a change
in either actor alters the continuation values both use, re-evaluating after
every update is costly. A shared critic is therefore fitted once per epoch
and held fixed while the adversary actor takes several natural policy steps and
the allocation actor takes one, and the final adversary policy initializes the
next epoch. Neural policy and value networks extend the method to large
resource state spaces.

\noindent\textbf{Computational results and insights.}
We test the framework on personalized assortment with MNL choice and online
linear programming, using Inventory Balancing (IB) and Dual Mirror Descent (DMD) as
benchmarks. In exact instances, the learned policies attain \(99.90\%\) and
\(99.93\%\) of the robust optimum and improve certified worst case reward by
\(6.83\%\) and \(33.25\%\), respectively. On larger instances with ten
products or resources, the neural policies achieve mean payoff gains of
\(3.39\%\) and \(71.25\%\) against separately trained adversaries.
Across both applications, the adversary exploits premature depletion:
early requests encourage the use of resources that become more valuable
under later request. In assortment optimization, IB initially favors the
high revenue product over lower revenue alternatives; as depletion raises
its inventory penalty, IB becomes less willing to offer that product to
later customers who value it most. In online linear programming, DMD
exhausts bottleneck capacity early and must reject more profitable requests
later. The learned policies preserve greater flexibility: they offer
lower-revenue alternatives alongside scarce high-revenue products and
distribute capacity use across alternative resources. These decisions
retain a more balanced resource mix and sustain earning opportunities
later in the horizon.

\noindent\textbf{Theoretical contribution.}
Adversarial updates can cycle even when their average converges, so we
study the current policy pair, which is the allocation policy a firm would
deploy. Following the policy-optimization literature
\citep{agarwal2021theory,cen2022fast,cen2023faster}, we analyze the tabular
model, which raises two additional difficulties specific to online resource allocation problem: First, policy gradient analyses typically assume that every state is
visited with positive probability, through a full-support initial
distribution, irreducible dynamics, or a bounded distribution-mismatch
coefficient \citep{zeng2022regularized,wei2021last,chen2024accelerated}.
Inventory instead starts from a fixed level and only depletes, so these
conditions fail. Second, the reused
critic evaluates policies that differ from those being updated. Natural
preconditioning and entropy regularization address the first and accelerates convergence. The observed-type timing addresses the second: it makes the iteration
triangular across periods, so adversary affects allocation only through
continuation values, whose changes conditional KL identities bound by
squared policy-logit errors in later periods. The current pair then
converges geometrically to an explicit error neighborhood of the
equilibrium (Theorem~\ref{thm:sampled}), and reusing the critic adds no
separate limiting error (Corollary~\ref{cor:sampled-revenue}).


\subsection{Literature}
\label{subsec:lit}
\noindent\textbf{Online resource allocation and robust revenue management.}
Online resource allocation is usually studied under exogenous arrivals. For online
linear programming, a dynamic dual-price learning algorithm is near-optimal
in the random-permutation model \citep{agrawal2014dynamic}, and dual mirror
descent performs well across stochastic, adversarial, and nonstationary
inputs \citep{balseiro2023best}. Under adversarial arrivals, inventory
balancing gives competitive guarantees for personalized assortment
\citep{golrezaei2014real}, and \citet{ma2020algorithms} obtain tight
weight-dependent competitive ratios for matching, assortment, and pricing.
Robust revenue management guards against request misspecification through
competitive analysis of booking control \citep{ball2009toward}, robust
network controls \citep{perakis2010robust}, and robust assortment
optimization under uncertain multinomial logit parameters
\citep{rusmevichientong2012robust}. Other streams blend stochastic and
adversarial arrivals \citep{esfandiari2015online}, account for uncertainty
about the horizon or the available samples
\citep{balseiro2023online,golrezaei2022online}, or learn online algorithms
with strong worst-case behavior by adversarial training
\citep{zuzic2020learning,du2022adversarial}. In all of these models, the
arrival sequence or its distribution is fixed outside the resource path that policy creates, so every policy faces the same request. When arrivals adapt to the resource state, each allocation policy can
induce a different request sequence. The competitive ratio can be
arbitrarily small when the offline optimum is evaluated on the
sequence induced by the policy being assessed
(Proposition~\ref{pro:cr}).  Our robust objective measures the revenue
that a single allocation policy can guarantee across all admissible
adversary policies on a fixed problem instance.

\noindent\textbf{Adaptive and state-dependent adversary.}
Operations management has long recognized that request responds to
inventory. Displayed inventory can stimulate request
\citep{balakrishnan2004stack}, stockouts shift request to substitutes
\citep{honhon2010assortment}, and field data show sizable stockout-based
spillovers \citep{li2023estimating}. These models specify the request
response, and the firm optimizes against it. Allocation and queueing models
with Markovian or state-dependent arrivals likewise fix the state-to-request
map \citep{jia2023online,abouee2016state}. Stateful online learning allows
actions to affect future states but compares all policies under the same
reward and transition maps, and its vanishing-regret guarantees rely on
mixing, chasability, or bounded memory
\citep{evendar2009online,abbasiyadkori2013online,emek2020stateful,arora2012policy}.
We treat the request itself as uncertain: an adversary chooses among
admissible types after observing the resource state, and the firm seeks the
revenue it can guarantee against every such response. In addition, irreversible depletion in resource allocation can make recovery losses grow linearly with the
horizon, ruling out a uniform sublinear chasing bound over our model
class (Proposition~\ref{prop:chase}).
Hence, we optimize this worst-case value rather
than regret against a fixed mechanism.

\noindent\textbf{Robust MDPs and adversarial reinforcement learning.}
Robust dynamic programming replaces a known kernel by an ambiguity set and
takes a worst case inside the recursion
\citep{iyengar2005robust,nilim2005robust,wiesemann2013robust}; tractability
comes from rectangularity, and relaxing it is costly
\citep{goyal2023robust}. The value depends on when uncertainty is resolved
and on what the controller may condition on \citep{wang2023foundation}.
Algorithms accelerate the inner worst case for specific ambiguity sets
\citep{ho2018fast}, scale it by first-order methods
\citep{grandclement2021scalable}, or reach it through regularization
\citep{derman2021twice}. Policy gradient methods for robust MDPs place the
uncertainty in the transition kernel: \citet{wang2023policygradient} prove
global convergence of a selected best iterate with exact model gradients,
\citet{li2025nonrectangular} find a near-worst kernel before each policy
update and bound average suboptimality, and \citet{chen2024accelerated}
accelerate the $s$-rectangular case with entropy regularization and extend
it to sampled gradients with linear features. Robust adversarial
reinforcement learning trains a protagonist against an adversary that
perturbs the dynamics \citep{pinto2017robust}; its convergence is understood
in linear-quadratic systems \citep{zhang2020stability}, and a Stackelberg
formulation adds adaptive regularization \citep{huang2022robustrl}. Our
adversary actor chooses operational request
types on the same resource states rather than perturbing a kernel, and we
analyze the last iterate of the resulting learning dynamics.

\noindent\textbf{Policy optimization in zero-sum Markov games.}
Once the adversary is a policy, the robust problem is a zero-sum Markov game, and the closest analyses are last-iterate
results for regularized policy optimization in such games.
Our setting differs in
three ways. First, moves are sequential within a period: the adversary selects a request type, and the firm observes it before allocating. The firm's update at a
state and type therefore does not involve the same-period request
probabilities, which makes the learning dynamics triangular across periods.
In simultaneous-move games, each player's update depends on the other's
current strategy at the same state. Second, existing analyses refresh values
at every iteration, from an exact model in
\citet{cen2023faster,zeng2022regularized} and through a slowly updated
critic in \citet{wei2021last}. We fit one critic per epoch from simulated
paths and reuse it for several request updates and one allocation update; the
analysis treats this reuse as part of the iteration rather than as an error
that must vanish. Third, discounted analyses either update every state with
an exact model or require coverage: a full-support initial distribution
\citep{zeng2022regularized}, irreducible chains for sampled updates
\citep{wei2021last}, or a finite distribution-mismatch coefficient
\citep{chen2024accelerated}. 

For online resource allocation problem, inventory may deplete from a fixed initial level,
so the initial distribution could be a point mass, inventory never returns to a
higher level, and many states are rarely visited. None of these coverage conditions holds; we use only
structural reachability. We establish an explicit
global geometric bound under persistent critic and policy-update errors.
The iteration bound allows a point-mass initial inventory distribution
and requires no recurrence assumption.

\noindent\textbf{Stackelberg and bilevel learning.}
Because the firm commits to an allocation policy and request responds, our
problem can also be read as a Stackelberg game. From the firm's side, the
objective is then the envelope of Section~\ref{sec:framework}, which need not be concave in the allocation
parameters. Stackelberg and bilevel methods follow the gradient of such
envelopes, so they need second-order information about the follower
\citep{fiez2020implicit,huang2022robustrl}, global gradient dominance in
linear-quadratic systems \citep{zhang2020stability}, strong concavity of the
follower's value \citep{he2025stackelbergmarkov}, or a strongly convex lower
level
\citep{ghadimi2018approximation,hong2023twotimescale,goktas2023convexconcave}.
We never differentiate the envelope. In a zero-sum game, commitment does not
change the value, and entropy makes the equilibrium unique (Section~\ref{subsec:nested}). Both policies can therefore move
toward their soft best responses together, and the envelope enters only as
the yardstick for the equilibrium gaps.

\section{Model Formulation}\label{sec:model}
We consider a firm that allocates limited inventory or capacity as requests arrive over a finite horizon. The firm observes each request before deciding how to serve it. Its decisions change the remaining resources and may therefore change future requests. We represent uncertainty about this request by a set of admissible adversary policies and choose an allocation policy to maximize its worst case expected revenue.

A firm allocates resources over periods $h=1,\ldots,H$. At the beginning of period $h$, the system is in state $s_h\in\cS$, with initial distribution $s_1\sim\rho$.
The state typically records the remaining resource vector, possibly augmented by queue lengths, inventory mix, or accumulated request information. Let $\cB$ be a finite set of request types and $\cB_h(s)\subseteq\cB$ the types admissible at state $s$ in period $h$. The adversary observes $s_h$ and selects $b_h\in\cB_h(s_h)$. 

\begin{figure}[htbp]
    \centering
    \includegraphics[width=0.6\linewidth]{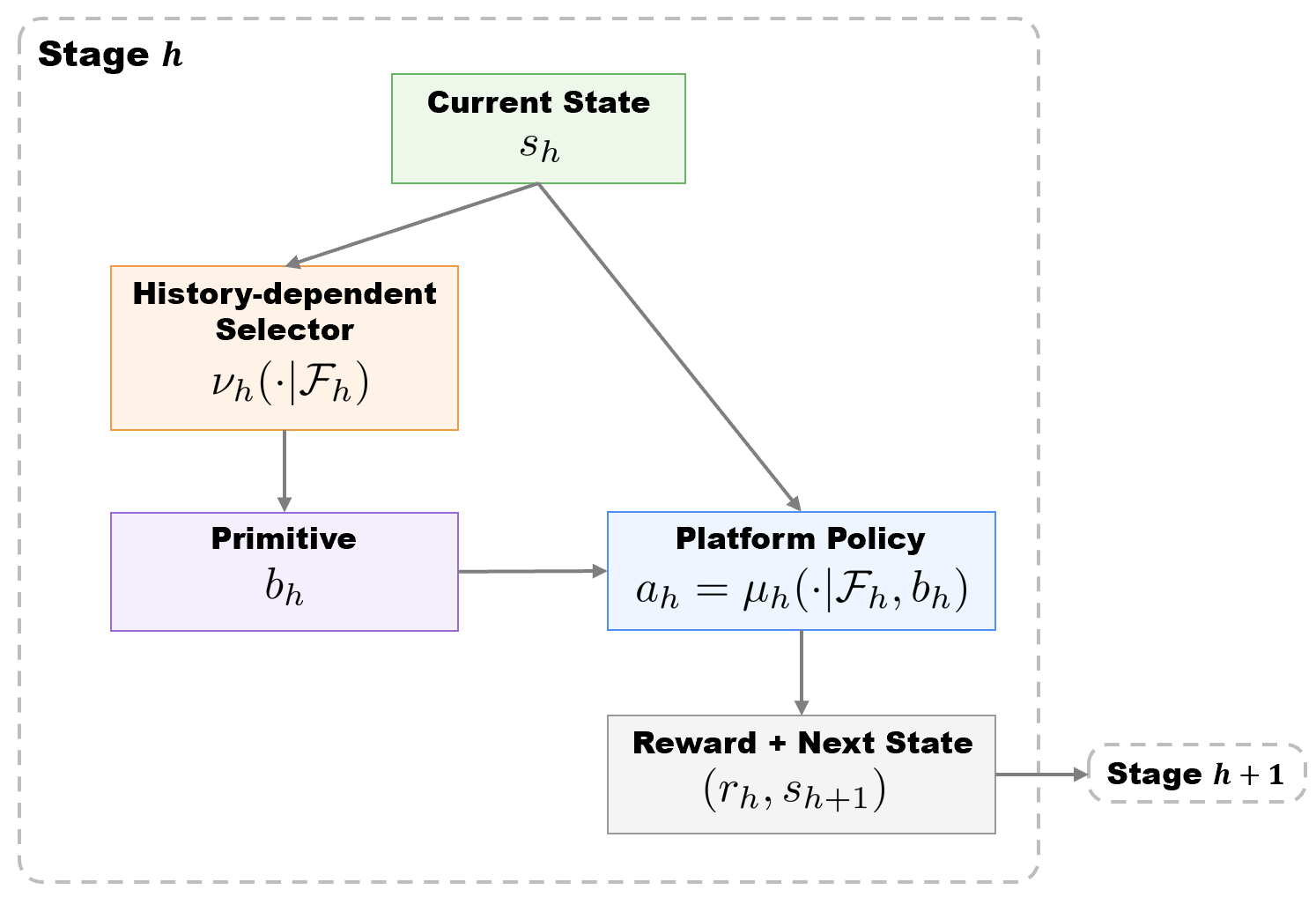}
    \caption{Within-stage timing with observed request type.}
    \label{fig:timing}
\end{figure}

The set $\mathcal B_h(s)$ is an uncertainty set of allowed request types for robust
planning. The firm can
limit the conservatism of worst-case planning by calibrating this set
using historical request frequencies at similar resource states, a
forecast supplemented with plausible alternative types, or business
rules governing substitution after stockouts. When a forecast also
provides type probabilities, Appendix~\ref{app:kl-extension} gives a
distributionally robust extension that uses a KL penalty to control
how far the adversary can profitably deviate from the forecast. The firm then observes $b_h$, chooses an
allocation $a_h\in\cA_h(s_h,b_h)\subseteq\cA$, earns revenue $R_h$ with $\E[R_h\mid s_h,a_h,b_h]=r_h(s_h,a_h,b_h)$, and moves to
$s_{h+1}\sim P_h(\cdot\mid s_h,a_h,b_h)$. All sets are finite and nonempty and revenues are
bounded by $R_{\max}$. Infeasible types and actions always receive zero probability. Figure~\ref{fig:timing} shows the order of moves.

This timing separates two sources of uncertainty. The adversary policy determines which observed request
arrives. Conditional on it, the operating model determines purchase, service, and other random outcomes. Thus an observed customer trajectory may be included in $b_h$, whereas an
unobserved preference shock cannot be used by the allocation policy. Define $\Delta(\mathcal X)$ as the simplex over a finite set $\mathcal X$ and $\mathcal F_h=\sigma(s_1,b_1,a_1,r_1,\ldots,s_h)$ for the history before the current request is selected. History-dependent policies satisfy $\nu_h(\cdot\mid\mathcal F_h)\in\Delta(\cB_h(s_h))$ for the history-dependent adversary and $\mu_h(\cdot\mid\mathcal F_h,b_h)\in\Delta(\cA_h(s_h,b_h))$ for the allocation policy. Their policy classes are denoted as $\cN_{\rm H},\cM_{\rm H}$.
The Markov subclasses $\cN_{\rm M},\cM_{\rm M}$ contain policies of the form $\nu_h(b\mid s)$ and $\mu_h(a\mid s,b)$ that depend only on the current state. 

For a policy pair $(\mu,\nu)$ with $s_1\sim\rho$, define expected revenue, robust revenue, and
the optimal robust value by
\begin{equation}
\begin{aligned}
J(\mu,\nu)
&=\E_{\rho,\mu,\nu}\!\left[\sum_{h=1}^H r_h(s_h,a_h,b_h)\right],\\
\operatorname{Rev}(\mu)
&=\inf_{\nu\in\cN_{\rm H}}J(\mu,\nu),\qquad
V^*=\sup_{\mu\in\cM_{\rm H}}\operatorname{Rev}(\mu).
\end{aligned}
\label{eq:objective}
\end{equation}
We call $\operatorname{Rev}(\mu)$ the \emph{robust revenue} of the allocation policy $\mu$. Accordingly, the robust value is 
$$V^*\coloneqq \sup_\mu\inf_\nu J(\mu,\nu),$$ 
i.e., the largest revenue the firm can guarantee by committing to an allocation policy before the adversary responds to it. 

\begin{example}{Online personalized assortment}\label{exp:gol}
There are $n$ products with inventories $s=(s_1,\ldots,s_n)$ and unit
revenues $r_i$. The observed request type $b$ identifies a customer segment.
An action $a$ is an offered assortment, so $\cA_h(s,b)=\{a:a\subseteq\{i:s_i>0\}\}$. The empty assortment is feasible. Under multinomial logit choice, let
$w_{bi}>0$ and $w_{b0}>0$ be the attraction parameters of product $i$ and the
outside option for segment $b$. The purchase probabilities are
\[
p_i(a,b)=\frac{w_{bi}}{w_{b0}+\sum_{j\in a}w_{bj}},\quad i\in a,
\qquad
p_0(a,b)=\frac{w_{b0}}{w_{b0}+\sum_{j\in a}w_{bj}}.
\]
Expected revenue is $r_h(s,a,b)=\sum_{i\in a}r_i p_i(a,b)$. A purchase of
product $i$ generates revenue $r_i$ and changes the inventory to $s-e_i$,
where $e_i$ is the $i$th unit vector. The outside option generates zero
revenue and leaves $s$ unchanged.
\end{example}

\begin{example}{Online linear programming}\label{exp:san}
A firm manages \(m\) resources with remaining capacity
\(s\in\cS\subseteq\mathbb R_+^m\). Request type \(b\) identifies a class and a set
\(\mathcal X_b\) of service options. Choosing \(x\in\mathcal X_b\) earns
reward \(q_b^\top x\) and consumes \(A_bx\) units of resources. Hence,
\[
\mathcal A_h(s,b)
=
\{x\in\mathcal X_b:A_bx\le s\},
\qquad
r_h(s,x,b)=q_b^\top x,
\]
and the next resource state is \(s-A_bx\). The reject option \(x=0\) is always
feasible. The standard accept-or-reject model is obtained by taking
\(\mathcal X_b=\{0,1\}\).
\end{example}

\subsection{Markov policies and dynamic programming}

\citet{wang2023foundation} show how admissible policy classes and the
structure of the ambiguity set affect the validity of robust dynamic
programming. In our model, the admissible request types are determined
by the current period and resource state, and the adversary may specify
a request distribution separately at each history. This rectangularity
allows backward induction on the resource state.

\begin{remark}[Markov policies and dynamic programming]
\label{rem:markov-reduction}
Every Markov allocation policy admits a deterministic, stage-dependent
Markov worst response.
Consequently, for every $\mu\in\cM_{\rm M}$,
\[
\operatorname{Rev}(\mu)
=\inf_{\nu\in\cN_{\rm H}}J(\mu,\nu)
=\min_{\nu\in\cN_{\rm M}}J(\mu,\nu).
\]
Moreover, a deterministic Markov policy pair is a saddle point over
the full history-dependent policy classes. Its value is
$V^*=\E_\rho[V_1(s_1)]$, where $V_h$ satisfies the backward recursion
\begin{equation}
V_h(s)=\min_{b\in\cB_h(s)}\max_{a\in\cA_h(s,b)}
\left\{r_h(s,a,b)+
\sum_{s'}P_h(s'\mid s,a,b)V_{h+1}(s')\right\},
\qquad V_{H+1}\equiv0.
\label{eq:bellman}
\end{equation}
The order of optimization reflects the allocation policy's ability to
respond to the observed request realization.
\end{remark}

The reduction gives a finite-horizon, zero-sum Markov Stackelberg
representation of the commitment problem, with the firm as the leader
and the adversary as the follower.

\subsection{Relation to robust Markov decision processes}
\label{subsec:rmdp}
In a robust MDP representation with one decision per period at resource
state $s$, the firm's action at $(h,s)$ is a \emph{conditional plan}
$\varphi=(\varphi_b)_{b\in\cB_h(s)}$. The plan assigns a randomized
allocation $\varphi_b\in\Delta(\cA_h(s,b))$ to every possible request type
$b$, so the realized type determines which allocation distribution the
firm uses. Let $\Phi_h(s)=\prod_{b\in\cB_h(s)}\Delta(\cA_h(s,b))$ be
the set of all plans. A type distribution $\eta\in\Delta(\cB_h(s))$
gives the one-step model
\begin{equation}
\bar r_h(s,\varphi;\eta)=\sum_{b,a}\eta(b)\varphi_b(a)\,r_h(s,a,b),\qquad
\bar P_h(s'\mid s,\varphi;\eta)=\sum_{b,a}\eta(b)\varphi_b(a)\,P_h(s'\mid s,a,b).
\label{eq:plan-model}
\end{equation}
We define an ambiguity set as type-generated if, at every stage--state
pair $(h,s)$, it is constructed entirely from these models for
$\eta\in\Delta(\cB_h(s))$. A single distribution $\eta$ generates the
models for all plans at a stage--state pair. Allowing the generating
distributions to be chosen separately across stage--state pairs gives
an $s$-rectangular ambiguity set \citep{wiesemann2013robust}.

\begin{proposition}[Robust MDP representation]
\label{prop:lifting}
At each stage--state pair $(h,s)$, a Markov policy pair $(\mu,\nu)$
defines a conditional plan $\varphi$ with $\varphi_b=\mu_h(\cdot\mid s,b)$
and a type distribution $\eta=\nu_h(\cdot\mid s)$. Conversely, statewise
choices of plans and type distributions define a Markov policy pair.
The two representations have the same conditional expected revenue and
next-state law under \eqref{eq:plan-model}. By
Remark~\ref{rem:markov-reduction}, they have the same optimal robust value.
\end{proposition}

Explicitly enumerating deterministic plans can be costly. A deterministic
plan fixes one allocation for each type, so there are
$\prod_{b\in\cB_h(s)}|\cA_h(s,b)|$ plans at $(h,s)$. In
Example~\ref{exp:gol} with $n=10$ products and six request types, each
type has $2^{10}=1{,}024$ feasible assortments when all products are in
stock. A deterministic plan is a list of six assortments, giving
$(2^{10})^6=2^{60}\approx10^{18}$ lists at a single state. The sequential
formulation uses one allocation distribution per type, with
$6\times2^{10}$ probability entries per state. The Bellman recursion
in \eqref{eq:bellman} optimizes separately for each observed type.

\paragraph{Inventory information.}
With an adaptive adversary, an allocation policy can change both
future capacity and future requests. Proposition~\ref{prop:inventory-information}
shows that, even when the same request types are admissible at every
state, averaging over inventory before evaluating the worst case can
overstate the revenue the firm can guarantee.

\begin{proposition}[Value of inventory information to the adversary]
\label{prop:inventory-information}
Consider a subgame starting in period $h$ with $s_h\sim\rho_h$,
where $\rho_h$ is known to both players. Suppose
$\cB_h(s)=\cB_h$ for every $s$ with $\rho_h(s)>0$. Using the continuation
value $V_{h+1}$ in \eqref{eq:bellman}, define
\[
q_h(s,b)=\max_{a\in\cA_h(s,b)}
\left\{r_h(s,a,b)+
\sum_{s'}P_h(s'\mid s,a,b)V_{h+1}(s')\right\}.
\]
Let $V_h^{\mathrm{obs}}(\rho_h)$ and $V_h^{\mathrm{hid}}(\rho_h)$ denote
the firm's robust values when the adversary observes $s_h$ and
when its period-$h$ request choice uses only $\rho_h$, respectively.
In both cases, the firm
observes $(s_h,b_h)$ before choosing $a_h$, and the adversary observes
the state from period $h+1$ onward. Then
\begin{equation}
\label{eq:inventory-information-gap}
\begin{aligned}
V_h^{\mathrm{obs}}(\rho_h)
&=\E_{\rho_h}\!\left[\min_{b\in\cB_h}q_h(s_h,b)\right]\\
&\le \min_{b\in\cB_h}\E_{\rho_h}[q_h(s_h,b)]
=V_h^{\mathrm{hid}}(\rho_h).
\end{aligned}
\end{equation}
Equality holds if and only if a single type $b\in\cB_h$ minimizes
$q_h(s,b)$ at every state $s$ with $\rho_h(s)>0$.
\end{proposition}

\subsection{Competitive ratios and regret}
\label{subsec:cr}

For a fixed adversary sequence $U$, competitive analysis compares a
policy's revenue with $\mathrm{OPT}(U)$, the revenue attainable if that
sequence were known in advance. With an adaptive adversary, the
allocation and adversary policies jointly generate the sequence
$U^{\mu,\nu}$. The offline benchmark $\mathrm{OPT}(U^{\mu,\nu})$
can therefore vary across allocation policies.

\begin{example}[Two periods, one unit]
\label{ex:two}
The firm has one unit of inventory and faces request values in
$\{0,\varepsilon,1\}$, where $0<\varepsilon<1$. A fixed Markov adversary
policy presents value $\varepsilon$ in period one. In period two it presents
value one if the unit has been sold and value zero otherwise. If $p$ is the
probability of accepting the first request, then
\begin{equation}
J(\mu,\nu)=p\varepsilon,\qquad
\E[\mathrm{OPT}(U^{\mu,\nu})]=p+(1-p)\varepsilon,\qquad
\frac{J(\mu,\nu)}{\E[\mathrm{OPT}(U^{\mu,\nu})]}\le\varepsilon.
\label{eq:cr-example}
\end{equation}
The denominator is positive even when $p=0$, in which case the ratio is zero.
\end{example}

\begin{proposition}[The policy-dependent competitive ratio can vanish]
\label{pro:cr}
For every $c>0$, there is a finite two-period instance with request values
in $[0,1]$ and a fixed Markov adversary policy $\nu$ such that every randomized
online resource allocation policy $\mu$ satisfies
$J(\mu,\nu)/\E[\mathrm{OPT}(U^{\mu,\nu})]<c$.
\end{proposition}

When the adversary sequence is fixed independently of the policy,
$\mathrm{OPT}(U)$ gives a common offline benchmark. We report
fixed-sequence ratios in Section~\ref{sec:experiments} to compare
performance under exogenous requests.

A fixed request mechanism also permits a regret comparison. For a fixed policy $\nu$ and a finite
class $\Gamma\subseteq\cM_{\rm M}$, let
\begin{equation}
\operatorname{Reg}_H(\mu;\nu,\Gamma)=\max_{\bar\mu\in\Gamma}J(\bar\mu,\nu)-J(\mu,\nu),
\label{eq:regret}
\end{equation}
where $\nu$ responds to each policy's own resource path. Regret
measures the shortfall relative to the best policy for one mechanism.
The robust objective measures the revenue a policy guarantees across
all admissible mechanisms. Regret bounds for stateful problems often
rely on a chasing condition, under which the decision maker can follow
a comparison policy from any state at a controlled loss
\citep{emek2020stateful}. Irreversible resource depletion can make
this recovery loss grow linearly in the horizon.

\begin{proposition}[Uniform chasability fails]
\label{prop:chase}
For every $H\ge2$, there is a deterministic instance with revenues in $[0,1]$ and one
admissible type per period in which, after one period, the first of two Markov allocation
policies reaches a state from which every continuation earns $H-1$ less than the second
policy earns over periods $2,\ldots,H$. Hence any recovery loss bound that holds uniformly over this model
class must be at least $H-1$ and cannot be $o(H)$.
\end{proposition}

\section{A Reinforcement Learning Framework for Online Robust Allocation}
\label{sec:framework}
As the number of resources grows, exact robust dynamic programming requires evaluating a large inventory state space. We therefore use an actor–critic framework that learns an allocation policy conditional on the observed request type and a state-dependent adversary policy from simulated trajectories. A shared critic is fitted once per epoch and supports several request updates and one allocation update. 
Following common practice in RL theory
\citep{agarwal2021theory,cen2023faster}, we first analyze tabular policies to quantify how errors in critic evaluation and
policy implementation affect the current allocation policy and request
response. 

\subsection{Regularization and equilibrium errors}
\label{subsec:nested}
The adversary and allocation policies are softmax policies,
$\nu_{\phi,h}(\cdot\mid s)\in\Delta(\mathcal B_h(s))$ and
$\mu_{\theta,h}(\cdot\mid s,b)\in\Delta(\mathcal A_h(s,b))$, where $\Delta$
denotes the probability simplex over feasible choices. With temperature
$\alpha>0$, define the regularized one-step reward
\[
\widetilde r_{\alpha,h}^{\mu,\nu}(s,a,b)
=r_h(s,a,b)-\alpha\log\mu_{\theta,h}(a\mid s,b)
+\alpha\log\nu_{\phi,h}(b\mid s).
\]
The opposite signs encourage both actors to retain mixed policies, since
allocation maximizes the objective and request minimizes it. In the complete
tabular class, regularization gives unique, smooth responses on reachable
decisions and supplies the damping term in the natural updates below.
Finite softmax logits give every feasible choice positive probability;
the bounded-trajectory argument keeps these probabilities uniformly
positive along training. The regularized return and robust envelope are
\begin{equation}
J_\alpha(\theta,\phi)
=\mathbb E_{\rho,\mu_\theta,\nu_\phi}
\left[\sum_{h=1}^H\widetilde r_{\alpha,h}^{\mu,\nu}(s_h,a_h,b_h)\right],
\qquad h_\alpha(\theta)=\min_\phi J_\alpha(\theta,\phi).
\label{eq:regularized-policy-objective}
\end{equation}
Let $A_{\max}=\max_{h,s,b}|\mathcal A_h(s,b)|$ and
$B_{\max}=\max_{h,s}|\mathcal B_h(s)|$. For fixed policies and
$0\le\alpha'\le\alpha$,
\[
-(\alpha-\alpha')H\log B_{\max}
\le J_\alpha(\theta,\phi)-J_{\alpha'}(\theta,\phi)
\le(\alpha-\alpha')H\log A_{\max}.
\]
Thus the regularized value approaches the robust objective of Section~2
as $\alpha\downarrow0$; Proposition~\ref{prop:regularization-bias} gives
the resulting revenue comparison.

For the tabular analysis, we define the inventory states that some
feasible policy pair can reach from arbitrary initial distribution $\rho$:
$\mathcal S_h^{\rm reach}
=\{s:\Pr_{\rho,\mu,\nu}(s_h=s)>0$
for some feasible $(\mu,\nu)\}$.
These states and their feasible decisions are \emph{structurally reachable}.
Demand has a separate logit vector $\phi_h(s)$ at each such state, and
allocation has one $\theta_h(s,b)$ for each admissible type $b$.
Since adding a common constant leaves softmax probabilities unchanged,
we impose $\mathbf1^\top\phi_h(s)=\mathbf1^\top\theta_h(s,b)=0$.
Decisions with one feasible choice have no free parameter; norms are
Euclidean on the remaining centered logits. Every finite-logit pair
reaches these decisions, even from a single initial inventory. Write $z=(\theta,\phi)$. The regularized game has a unique
saddle $z^\star=(\theta^\star,\phi^\star)$ on these coordinates, with value
$V_\alpha^\star$ (Proposition~\ref{ass:local}).

Write $\phi_\alpha(\theta)$ for the worst adversary policy to $\theta$ and
$d_h^{\theta,\phi}(s)$ for the probability that inventory is $s$ in period
$h$. The soft performance difference identity gives
\begin{equation}
J_\alpha(\theta,\phi)-h_\alpha(\theta)
=\alpha\sum_{h,s}d_h^{\theta,\phi}(s)
\operatorname{KL}\!\left(\nu_{\phi,h}(\cdot\mid s)\,\middle\|\,
\nu_{\phi_\alpha(\theta),h}(\cdot\mid s)\right).
\label{eq:kl-gap}
\end{equation}
The left side vanishes exactly when request is a worst response at every
reachable state; deviations at rarely reached states receive less weight.
Two gaps measure equilibrium error:
\begin{equation}
\Delta_A(\theta)=V_\alpha^\star-h_\alpha(\theta),\qquad
\Delta_R(\theta,\phi)=J_\alpha(\theta,\phi)-h_\alpha(\theta).
\label{eq:gaps}
\end{equation}
The allocation gap is the loss in regularized robust value. The response
gap is how much the current adversary policy overstates that value by falling
short of a worst response. The adversary policy seeks to reduce the response
gap, whereas an allocation update changes the response problem and can
enlarge it. Observed revenue alone therefore does not establish robustness.
Theorem~\ref{thm:sampled} controls both gaps for the current policy pair,
without requiring either to decrease at every update.

\subsection{Actor--Lazy Critic Learning with Natural Policy Gradient}
\label{subsec:sampled}
To limit the cost of policy evaluation at scale, we fit a reference
critic once per epoch and reuse it for several adversary updates and
one allocation update. The question is whether these updates still improve the current policies when the critic and the policy steps are approximate. In the tabular model, natural preconditioning
cancels visitation weights in the population update directions, and
entropy regularization damps the logit updates. Because the firm
observes the request type before allocating, the exact allocation
direction depends on the adversary policy only through later-period
continuation values. This ordering gives a triangular error recursion.
Under its step-size and uniform accuracy conditions,
Theorem~\ref{thm:sampled} bounds the last-iterate logit error by a
geometrically decaying term plus an explicit term for critic and
policy-update errors.

\paragraph{One critic fit per epoch.}
The critic estimates the regularized return from the current decision to
the end of the horizon. For a policy pair $z=(\theta,\phi)$ and tuple
$X_h=(s_h,a_h,b_h)$, this return separates into the current reward and
the continuation value:
\begin{equation}
Q_h^z(X_h)=\widetilde r_{\alpha,h}^z(X_h)
+\mathbb E_z[Q_{h+1}^z(X_{h+1})\mid X_h],\qquad Q_{H+1}^z=0.
\label{eq:fixedpair}
\end{equation}
We fit this recursion backward for the epoch's reference pair
$z_t=(\theta_t,\phi_{t,0})$. At stage $h$, a fresh batch of $B_{Q,t}$
complete trajectories supplies targets consisting of the observed
regularized reward plus the already fitted continuation:
\begin{align}
Y_{t,h}^i
&=\widetilde R_{\alpha,h}^i
+Q_{\xi_{t,h+1},h+1}(X_{h+1}^i),
\qquad Q_{\xi_{t,H+1},H+1}=0,
\label{eq:targets}\\
\mathcal L_{t,h}(\xi)
&=\frac1{B_{Q,t}}\sum_{i=1}^{B_{Q,t}}
\{Q_{\xi,h}(X_h^i)-Y_{t,h}^i\}^2.
\label{eq:critic-fit}
\end{align}
Here $\xi$ denotes the current stage's critic parameters. The reward
uses the reference policies; the continuation is evaluated at the next
tuple in the same trajectory and held fixed during regression.
Independent stage batches keep its fit independent of the current
regression. The critic is fitted by minimizing the regression loss
\eqref{eq:critic-fit}.
The resulting critic $\widehat Q_t=Q_{\xi_t}$
is shared by both actors and held fixed throughout the epoch.

\paragraph{Natural policy updates.}
Write $\mu_t=\mu_{\theta_t}$ and $\nu_{t,k}=\nu_{\phi_{t,k}}$.
The conditional value of a type is
$\bar V^\theta_{t,h}(s,b)
=\sum_a\mu_{t,h}(a\mid s,b)\widehat Q_{t,h}(s,a,b)$ and $\bar{V}^\phi_{t,k,h}(s)
=\sum_b\nu_{t,k,h}(b\mid s)\bar V^\theta_{t,h}(s,b)$.
Using the frozen critic, define
\begin{equation}
\begin{aligned}
\widehat A^\theta_{t,h}(s,a,b)
&=\widehat Q_{t,h}(s,a,b)-\bar V^\theta_{t,h}(s,b),\\
\widehat A^\phi_{t,k,h}(s,b)
&=\bar V^\phi_{t,k,h}(s)-\bar V^\theta_{t,h}(s,b)
\end{aligned}
\label{eq:neural-ppo-advantages}
\end{equation}
Positive advantage favors the corresponding actor: allocation prefers
actions with higher value, while adversary prefers types with lower value
to the firm. The conditional means provide baselines for comparing
choices at the same inventory state.

Each actor maximizes its expected signed advantage with a KL penalty
for departing from its current policy. With penalty weights $1/\eta$
for adversary and $1/\lambda$ for allocation, the solutions are
\begin{equation}
\begin{aligned}
\nu_{t,k+1,h}(b\mid s)
&\propto\nu_{t,k,h}(b\mid s)
\exp\{\eta\widehat A^\phi_{t,k,h}(s,b)\},\\
\mu_{t+1,h}(a\mid s,b)
&\propto\mu_{t,h}(a\mid s,b)
\exp\{\lambda\widehat A^\theta_{t,h}(s,a,b)\}.
\end{aligned}
\label{eq:allocation-mirror}
\end{equation}
Each row is normalized over its feasible choices. Both actors therefore
shift probability toward positive-advantage choices without solving
their full response problems.
For a policy block with $m$ feasible choices, define the centering operator
$\Pi\coloneqq I_m-\frac1m\mathbf1_m\mathbf1_m^\top$,
where $\mathbf1_m$ is the vector of ones. The same notation applies within allocation blocks
$(h,s,b)$ and adversary blocks $(h,s)$. In centered logits,
\eqref{eq:allocation-mirror} becomes
\begin{equation}
\begin{aligned}
\phi_{t,k+1}&=\phi_{t,k}-\eta G^\phi_{t,k},
&G^\phi_{t,k}&=-\Pi\widehat A^\phi_{t,k},\\
\theta_{t+1}&=\theta_t+\lambda G^\theta_t,
&G^\theta_t&=\Pi\widehat A^\theta_t.
\end{aligned}
\label{eq:natural-logit-update}
\end{equation}
\begin{remark}[A soft bid-price rule]
For one resource, acceptance earns fare $q_b$ and consumes one unit;
rejection preserves capacity. At the regularized equilibrium, the acceptance
probability is
$[1+\exp\{-(q_b-\Delta V_{h+1}^\star(s))/\alpha\}]^{-1}$ for $s\ge1$,
where $\Delta V_{h+1}^\star(s)=V_{h+1}^\star(s)-V_{h+1}^\star(s-1)$
is the robust marginal value of capacity under the regularized value
function. This is an entropy-smoothed bid-price rule: the robust marginal value
of capacity serves as the opportunity cost against which the fare is
compared, following the bid-price interpretation of
\citet{talluri1998analysis}.
\end{remark}

\paragraph{Sampled directions and the algorithm.}
For large problems, actors estimate the ideal directions from $B_{j,t}$ fresh
paths of the current pair. At a current pair $z=(\theta_t,\phi_{t,k})$, let
$j\in\{\theta,\phi\}$ index allocation and adversary policy, respectively.
Actor $j$ draws $B_{j,t}$ fresh trajectories under $z$.
The stage scores $\psi_h^\theta =\nabla_\theta\log\mu_{\theta,h}(a_h\mid s_h,b_h)$, and $\psi_h^\phi =\nabla_\phi\log\nu_{\phi,h}(b_h\mid s_h)$ are evaluated at $z$; superscript $i$ denotes their values on trajectory $i$. Define the sampled score estimates by
\begin{equation}
\widehat S^j(z)\coloneqq 
\begin{cases}
\displaystyle
\frac1{B_{\theta,t}}\sum_{i=1}^{B_{\theta,t}}\sum_{h=1}^H
\psi_h^{\theta,i}\widehat A^\theta_{t,h}(X_h^i),
& j=\theta,\\[8pt]
\displaystyle
-\frac1{B_{\phi,t}}\sum_{i=1}^{B_{\phi,t}}\sum_{h=1}^H
\psi_h^{\phi,i}\widehat A^\phi_{t,k,h}(s_h^i,b_h^i),
& j=\phi.
\end{cases}
\label{eq:advantages}
\end{equation}
For either actor, its population Fisher matrix is
$F_j(z)\coloneqq \sum_{h=1}^H\mathbb E_z[\psi_h^j(\psi_h^j)^\top]$.
The empirical matrix and estimated natural direction are
\begin{equation}
\begin{aligned}
\widehat F_j(z)
&\coloneqq \frac1{B_{j,t}}\sum_{i=1}^{B_{j,t}}\sum_{h=1}^H
\psi_h^{j,i}(\psi_h^{j,i})^\top+\varepsilon_{\rm rid}I,\\
\widehat G^j(z)
&\coloneqq \widehat F_j(z)^{-1}\widehat S^j(z),
\qquad \varepsilon_{\rm rid}>0.
\end{aligned}
\label{eq:main-empirical-fisher}
\end{equation}
Here $I$ acts on actor $j$'s centered reachable tabular coordinates.
 Write $G^\theta(z)=G^\theta_t$ and
$G^\phi(z)=G^\phi_{t,k}$ for the ideal directions in
\eqref{eq:natural-logit-update}. Then condition on the current policies, frozen critic, and preceding updates:
\begin{equation}
\mathbb E_z[\widehat S^j(z)]
=\bar S_j(z;\widehat Q_t)
\coloneqq \sum_{h=1}^H\mathbb E_z[\psi_h^j\widehat Q_{t,h}(X_h)]
=F_j(z)G^j(z).
\label{eq:frozen-score-identity}
\end{equation}
Fisher preconditioning therefore recovers the ideal direction from its
population score mean. The ridge stabilizes the empirical inverse when decisions are poorly sampled, but introduces bias.

Algorithm~\ref{alg:sampled} uses $\widehat G^\phi_{t,k}
\coloneqq \widehat G^\phi(\theta_t,\phi_{t,k})$ and $\widehat G^\theta_t
\coloneqq \widehat G^\theta(\theta_t,\phi_{t,K})$.
Neural actors may instead approximately solve the conditional KL problems
in \eqref{eq:allocation-mirror}. For these actors, the same symbols denote
the induced centered output-logit changes divided by $-\eta$ and
$\lambda$, respectively. Their accuracy is measured against the ideal
directions in \eqref{eq:natural-logit-update}.

\begin{algorithm}[htbp]
\caption{Actor--lazy critic learning with approximate natural policy updates}
\label{alg:sampled}\label{alg:ac}
\begin{algorithmic}[1]
\Require Initial policies; temperature $\alpha$; steps $\eta,\lambda$; epochs $T$; number of adversary updates $K$; batches $B_{Q,t},B_{\phi,t},B_{\theta,t}$
\For{$t=0,\ldots,T-1$}
  \State Fit $\widehat Q_t$ for $z_t=(\theta_t,\phi_{t,0})$ by \eqref{eq:targets}--\eqref{eq:critic-fit}; freeze it
  \Comment{one critic fit per epoch}
  \For{$k=0,\ldots,K-1$}
    \State Draw fresh paths under $(\theta_t,\phi_{t,k})$; estimate the request direction $\widehat G^\phi_{t,k}$
    \State $\phi_{t,k+1}\gets\phi_{t,k}-\eta\widehat G^\phi_{t,k}$
  \EndFor
  \State Draw fresh paths under $(\theta_t,\phi_{t,K})$; estimate the allocation direction $\widehat G^\theta_t$
  \State $\theta_{t+1}\gets\theta_t+\lambda\widehat G^\theta_t$;\quad $\phi_{t+1,0}\gets\phi_{t,K}$
  \Comment{same frozen critic; warm start}
\EndFor
\State \Return $\mu_{\theta_T}$, $\nu_{\phi_{T,0}}$
\end{algorithmic}
\end{algorithm}

\paragraph{Evaluation effort.}
An epoch uses
$H^2B_{Q,t}+H\{KB_{\phi,t}+B_{\theta,t}\}$
simulated transitions: $H$ stage batches for one backward critic fit
and fresh paths for each actor update. Refitting before every actor
step would multiply the critic term by $K+1$ at the same batch sizes.
This saving is per epoch; total effort also depends on the required
accuracy and number of epochs. The fit--then--update schedule gives both actors a common value estimate during an epoch and separates critic fitting from policy optimization.

We first give the convergence guarantee, then translate its accuracy
requirements into sampling and ridge choices. The ideal directions $G^j$
use the frozen critic in \eqref{eq:natural-logit-update}; the reference
pair is $z_t=(\theta_t,\phi_{t,0})$.

\begin{theorem}[Global last-iterate convergence]
\label{thm:sampled}
Consider the complete finite tabular model with bounded realized rewards,
$\alpha>0$, and any initial distribution $\rho$. Fix $K\ge1$ and constant
steps satisfying
\begin{equation}
0<\alpha\lambda\le1,\qquad 0<\alpha K\eta\le1.
\label{eq:global-step-conditions}
\end{equation}
From any finite centered-logit initialization, run
Algorithm~\ref{alg:sampled}. For any $T\ge1$ and $\delta\in(0,1)$,
suppose that, with probability at least $1-\delta$, the bounds
$\|\widehat Q_t-Q^{z_t}\|_\infty\le\varepsilon_Q$,
$\|\widehat G^j-G^j\|_\infty\le\varepsilon_{\rm act}$, $j\in\{\theta,\phi\}$,
hold simultaneously at every evaluation in epochs $0,\ldots,T-1$.
The critic supremum covers reachable feasible tuples, and the direction
suprema cover centered reachable coordinates. Put
$\varepsilon_{\rm est}=\varepsilon_Q+\varepsilon_{\rm act}$ and suppose
\begin{equation}
\varepsilon_{\rm est}\le\frac{\alpha}{16H}.
\label{eq:global-accuracy-cap}
\end{equation}
Define
$E_t=\|z_t-z^\star\|_\infty$,
$\varrho_{\rm gl}
=1-\frac{\alpha}{2}\min\{\lambda,K\eta\}$.
On this event, the generated logits remain bounded and, for
$0\le t\le T$,
\begin{equation}
\begin{aligned}
E_t
&\le C_0\varrho_{\rm gl}^{\,t}E_0
       +\frac{4\varepsilon_{\rm est}}{\alpha},\\
\Delta_A(\theta_t)
&\le\frac{\alpha H}{2}E_t^2,\qquad
\Delta_R(z_t)\le\alpha H E_t^2,
\end{aligned}
\label{eq:global-error-floor-main}
\end{equation}
where $C_0$ is given explicitly in
\eqref{eq:global-rate-prefactor}.
If the accuracy event holds over all epochs, these bounds hold for
every $t\ge0$, and
$\limsup_{t\to\infty}E_t\le4\varepsilon_{\rm est}/\alpha$.
\end{theorem}

\begin{corollary}[Finite-time revenue guarantee]
\label{cor:sampled-revenue}
On the accuracy event of Theorem~\ref{thm:sampled}, for $0\le t\le T$,
\begin{equation}
\begin{aligned}
V^*-\operatorname{Rev}(\mu_{\theta_t})
\le{}&
\underbrace{\frac{\alpha H}{2}
\left(
C_0\varrho_{\rm gl}^{\,t}E_0
+\frac{4\varepsilon_{\rm est}}{\alpha}
\right)^2}_{\text{learning error}}\\
&+\underbrace{\alpha H\log(A_{\max}B_{\max})}
_{\text{regularization bias}}.
\end{aligned}
\label{eq:sampled-revenue}
\end{equation}
Under all-time accuracy, the limiting learning error is at most
$8H\varepsilon_{\rm est}^2/\alpha$.
\end{corollary}
Here $V^*$ is the unregularized robust value. A smaller temperature reduces
regularization bias but requires greater learning accuracy through both
\eqref{eq:global-accuracy-cap} and the factor $1/\alpha$.

\paragraph{Why one critic fit per epoch suffices.}
Entropy turns the natural updates into damped moves toward bounded
conditional targets, keeping logits bounded from any finite start
(Lemma~\ref{lem:global-boundedness}). The observed-type timing then makes
the iteration triangular across periods: allocation at period $h$ depends
on later-period values, while request also depends on the current allocation.
Conditional KL identities make continuation-value errors quadratic in
later-period logit errors. Final-period allocation settles first, followed
by request, and backward induction gives $\limsup_{t\to\infty}\|z_t-z^\star\|_\infty \le4\varepsilon_{\rm est}/\alpha$.

\paragraph{Why natural policy update under resource depletion.}
Starting from a fixed inventory, depletion creates many possible
resource states, some of which are rarely visited but remain relevant
for serving valuable later requests. Natural preconditioning removes
the visitation and choice probability weights from the policy update
at these decisions. Theorem~\ref{thm:sampled} gives a global convergence
bound from any finite centered-logit initialization, with factor
$\varrho_{\rm gl}=1-\frac{\alpha}{2}\min\{\lambda,K\eta\}$ per critic
fit and logit error floor at most
$4\varepsilon_{\rm est}/\alpha$.
The factor contains no visitation probabilities.

\begin{figure}[htbp]
    \centering
    \includegraphics[width=0.72\linewidth]{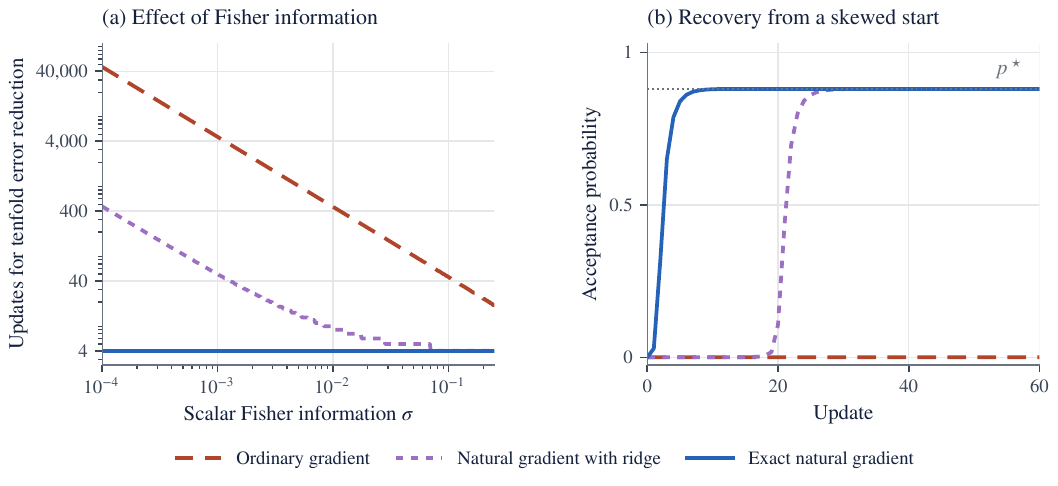}
    \caption{Natural preconditioning under exact values and expectations.
    (a) Local updates needed for a tenfold reduction in logit error as
    scalar Fisher information varies. (b) Acceptance probabilities,
    starting from log-odds $-9$ toward optimal log-odds $2$.
    Both panels use $\alpha\lambda=0.5$; the ridged update uses
    $\varepsilon_{\rm rid}=10^{-2}$.}
    \label{fig:natural-conditioning}
\end{figure}

Ordinary gradients multiply each conditional natural direction by
its Fisher block. Let
$\sigma_j^\star\coloneqq\lambda_{\min}(F_j(z^\star))$,
$j\in\{\theta,\phi\}$, where the population Fisher matrices are
restricted to centered reachable tabular coordinates.
With exact values and population scores, the same frozen-critic
schedule gives ordinary-gradient updates the local linearized factors
$1-\alpha\lambda\sigma_\theta^\star$ and
$1-\alpha K\eta\sigma_\phi^\star$ per epoch; their maximum governs
the linearized contraction
(Appendix~\ref{app:fisher-interpretation}).
For adversary, $\sigma_\phi^\star$ lies within a factor of two of the
smallest equilibrium probability of an inventory--type pair among
blocks with at least two admissible types
(Proposition~\ref{prop:sigma}).
In an ample-stock assortment problem with ten products and ten periods,
$\sigma_\phi^\star\le2/92{,}378$.
These probabilities can decay exponentially with the horizon, slowing
ordinary updates at rarely visited decisions. In practice, a positive ridge makes
the empirical Fisher matrix invertible but also damps natural updates. 
In the population linearization, the adversary factor becomes
$1-\alpha K\eta\,\sigma_\phi^\star/
(\sigma_\phi^\star+\varepsilon_{\rm rid})$
(Appendix~\ref{app:fisher-interpretation}). It still outperforms ordinary gradient. Figure~\ref{fig:natural-conditioning} compares the convergence speeds of ordinary, natural, and ridged natural updates.

\paragraph{Why multiple request updates per epoch.}
Multiple request updates allow a larger cumulative correction before
the allocation policy changes, while reusing the fitted critic.
This accommodates the small proximal steps used in the later neural
implementation. Within~\eqref{eq:global-step-conditions}, and for fixed
$\eta$ and accuracy tolerances, increasing $K$ improves the global
contraction bound until $K\eta$ reaches $\lambda$; thereafter, the
allocation step limits the bound. 

\paragraph{Accuracy and sampling budgets.}
Theorem~\ref{thm:sampled} needs small critic and direction errors at every reachable decision. In the tabular model, each
inventory--type cell is estimated only from its own visits, so the required batches grow with the inverse of the rarest reachable cell probability (Proposition~\ref{prop:ridge-accuracy} in
Appendix~\ref{app:sampled-accuracy}), the familiar coverage cost of tabular estimation \citep{azar2013minimax,agarwal2021theory}. Neural networks reduce this cost by sharing information across states. A critic shared across inventory levels needs small Bellman residuals only on the states the policies visit, and the parameterized Fisher matrix pools information across states, so its conditioning need not deteriorate with the rarest inventory level. The price is approximation error: accuracy now holds on average over
visited decisions rather than at every decision. Section~\ref{subsec:practical}
develops this implementation and the guarantees it retains
(Lemmas~\ref{lem:neural-reference-natural-error}
and~\ref{lem:neural-relative-fisher-solve}).

\subsection{Neural network implementation}
\label{subsec:practical}

Tabular logits do not scale. In the larger instances of
Section~\ref{sec:experiments}, inventory alone takes up to
$5^{10}\approx10^7$ values per period in assortment and
$7^{10}\approx3\times10^8$ in online linear programming, and the logits of an
unsampled state never move. Neural networks let what is learned at frequent
inventory states carry over to rare ones. Here $\theta$, $\phi$, and $\xi$
denote network weights. Training retains the epoch structure of Algorithm~\ref{alg:sampled}.

\begin{figure}[htbp]
    \centering
    \includegraphics[width=\linewidth]{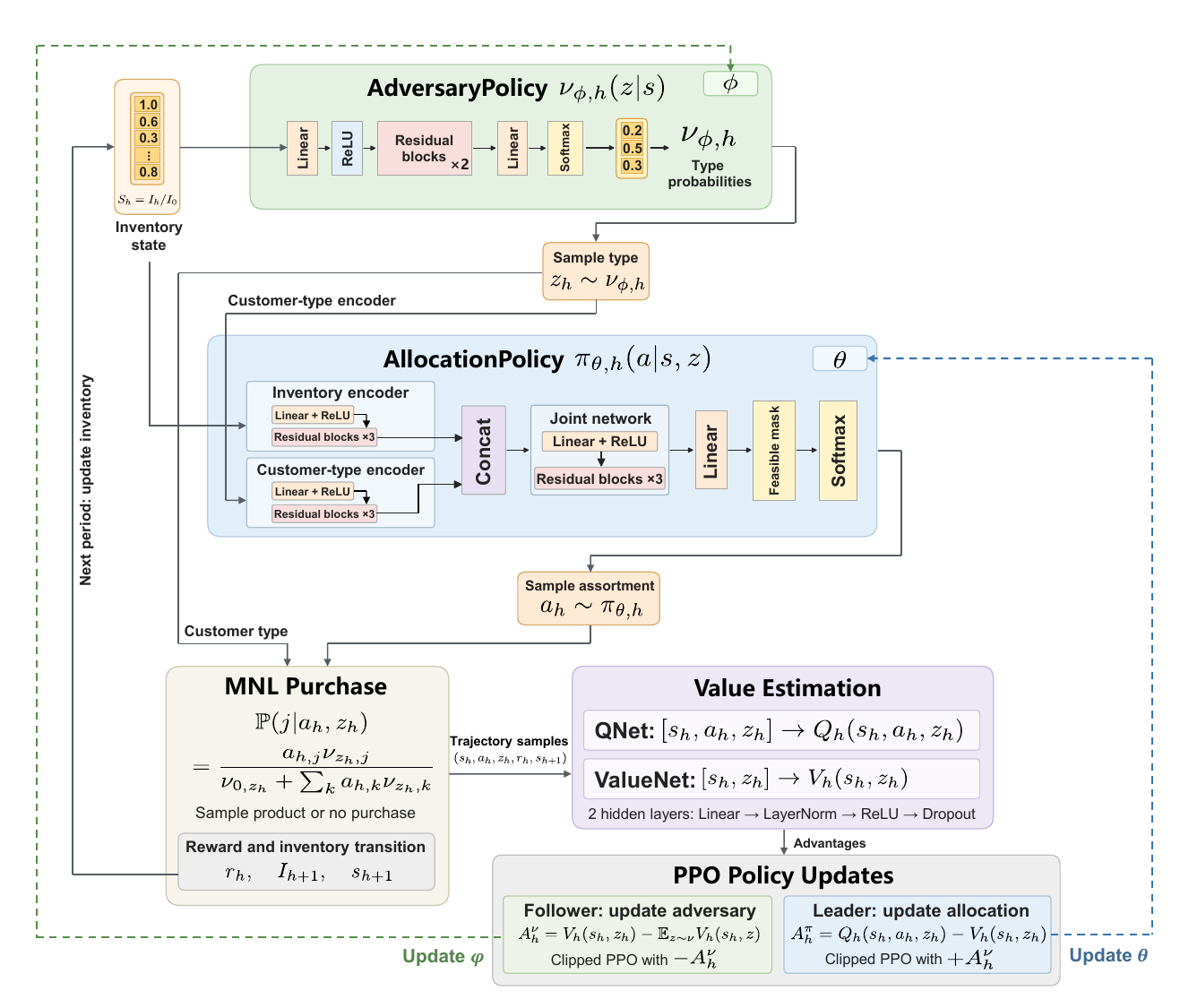}
    \caption{Neural implementation for personalized assortment. The request
    actor selects a customer type, the allocation actor selects a feasible
    assortment, and the simulator generates purchases and inventory
    transitions. Both actors use the same fitted critic during an epoch.}
    \label{fig:model}
\end{figure}

\paragraph{Architecture.}
Weights are \emph{not} shared across periods: each period $h$ has its own
request network, allocation network, and critic, with weights $\phi_h$,
$\theta_h$, and $\xi_h$ (Figure~\ref{fig:model}). Separate weights fit
nonstationary decisions, since the same inventory is worth more early than
late, and keep the backward critic fit from changing the period-$(h+1)$
values behind its targets \eqref{eq:targets}. All networks read inventory
normalized by its initial level. The request network maps it through a ReLU
layer and two residual blocks to a softmax over admissible types. The
allocation network conditions on the observed type, as the timing of
Section~\ref{sec:model} requires: it encodes inventory and type separately,
each by a ReLU layer and three residual blocks, concatenates the encodings,
and passes them through a joint network of the same form, because the same
inventory calls for different offers to different customers. A linear layer
then gives one logit per candidate assortment, and a mask removes
assortments with an out-of-stock product before the softmax. Residual
connections keep these deep networks stable to train, and masks give
infeasible choices exactly zero probability, with no penalty weight to tune.
The critic has two networks, each with two hidden layers (linear, layer
normalization, ReLU, dropout). The Q-network maps $(s,a,b)$ to
$\widehat Q_{t,h}(s,a,b)$ and is fitted by \eqref{eq:critic-fit}. The value network maps $(s,b)$ to $\bar V^\theta_{t,h}(s,b)$. Together they give both advantages \eqref{eq:neural-ppo-advantages}.
Online linear
programming uses the same structure, with request classes as types and
service options as actions.

\paragraph{PPO loss and training.}
Each actor is updated by clipped proximal policy optimization (PPO) with
the advantages \eqref{eq:neural-ppo-advantages}. For period-$h$ tuples
$X_h^i=(s_h^i,a_h^i,b_h^i)$ drawn under the current pair, let
$\gamma_h^i(\theta)
=\mu_{\theta,h}(a_h^i\mid s_h^i,b_h^i)/\mu_{t,h}(a_h^i\mid s_h^i,b_h^i)$
be the probability ratio and $\epsilon_{\rm clip}\in(0,1)$ the clip range.
The allocation network maximizes over $\theta_h$
\begin{equation}
\frac1{B_{\theta,t}}\sum_{i=1}^{B_{\theta,t}}
\min\Bigl\{\gamma_h^i(\theta)\,\widehat A^\theta_{t,h}(X_h^i),\;
\operatorname{clip}\bigl(\gamma_h^i(\theta),1-\epsilon_{\rm clip},
1+\epsilon_{\rm clip}\bigr)\widehat A^\theta_{t,h}(X_h^i)\Bigr\}.
\label{eq:ppo-loss}
\end{equation}
The request objective replaces the ratio by
$\nu_{\phi,h}(b_h^i\mid s_h^i)/\nu_{t,k,h}(b_h^i\mid s_h^i)$ and the advantage
by $\widehat A^\phi_{t,k,h}(s_h^i,b_h^i)$. 

The KL-penalized updates underlying \eqref{eq:allocation-mirror}
yield natural-gradient steps under a linear approximation of the
objective and a quadratic approximation of KL.
PPO provides a practical surrogate for these updates, with clipping
suppressing gains from large favorable probability-ratio changes.
This connection also motivates the request inner loop with $K>1$:
several small proximal updates can accumulate a larger request
adjustment against the fixed allocation policy while reusing the
same fitted critic.
Each inner step uses fresh paths and the current adversary policy
$\nu_{t,k}$ as its ratio reference.
The clip range and optimizer settings tune the size of each policy
update, while $K$ controls how many request updates precede the
allocation update.

\paragraph{Sampling example.}
Parameter sharing is useful when different inventory states have common
continuation-value structure. To illustrate the sampling gain, condition on the reference policies
and fitted continuation in one backward regression, and freeze an encoder
mapping feasible tuples into
$G$ groups. Let $p_g$ be the sampling probability of group $g$,
$p_{\min}=\min_gp_g>0$, and $m(x)=\mathbb E[Y\mid X=x]$ for targets
bounded by $|Y|\le M$. Suppose $m(x)$ differs from its group mean by at
most $a_{\rm rep}$. For $u>0$ and $\delta\in(0,1)$, fitting one head
coefficient per group from $B$ independent paths gives
\begin{equation}
B\ge\frac1{p_{\min}}
\max\left\{8,\frac{4M^2}{u^2}\right\}\log\frac{4G}{\delta}.
\label{eq:shared-group-sampling}
\end{equation}
With probability at least $1-\delta$, $\|\widehat Q_h-m\|_\infty\le a_{\rm rep}+u$, where $a_{\rm rep}$ measures the approximation cost.
The sampling cost depends on $G$, even if an
individual inventory tuple is arbitrarily rare. Here, $G$ is the width of a hidden representation with one indicator
feature per group. 
Appendix~\ref{app:neural-shared-sampling} gives the proof.

In practice, choose encoder width jointly with the simulation budget:
start with a compact representation that pools inventories with similar
continuation values, and increase its width only when finer distinctions
improve value prediction on held-out trajectories.
In the balanced-group example, keeping the sampling term fixed requires
$B$ to grow proportionally to $G\log(4G/\delta)$.
Thus, when a wider encoder improves training fit but worsens validation
accuracy, retain the smaller encoder or collect more trajectories.

\paragraph{Convergence guarantees for neural actors.}
The last-iterate bound in Theorem~\ref{thm:sampled} extends to the
policies produced by the neural actors if the reference critic is
uniformly accurate and each realized update is close to the
corresponding tabular natural step computed from that critic.
We measure this discrepancy in centered output logits, divide by the
corresponding step size, and require the result to be at most
$\varepsilon_{\rm act}$ on every reachable decision block. This
tolerance includes the effect of errors in the type-value network.
Applying the theorem to clipped PPO requires this accuracy condition
in addition to the theorem's step-size conditions.

For exact population natural-gradient updates with an exact critic frozen
within each epoch, Proposition~\ref{prop:neural-natural} gives local
linearized factors $1-\alpha\lambda$ for allocation and $1-\alpha K\eta$
for demand. This result requires exact representation of the regularized
tabular saddle, $C^2$ centered logits, and positive definite population
Fisher matrices on identifiable parameter coordinates.

Appendix~\ref{app:neural-natural} gives weighted accuracy certificates
using Bellman residuals, value-net accuracy, and Fisher-solve errors.
For fixed-opponent response probes, the analyzed projected refinement
bounds the average squared gradient mapping by an $O(T^{-1})$ term plus
the average squared gradient error
(Proposition~\ref{thm:neural-response-stationarity}).

\section{Computational Studies and Operational Implications}
\label{sec:experiments}

We test the method on personalized assortment allocation and online linear programming. In each application, we run both the tabular method (Algorithm~\ref{alg:ac}) and the neural method in Figure~\ref{fig:model} on a small instance, where exact robust evaluation certifies performance and reveals the mechanism behind the gains, and only the neural method on large instances, which tests whether the gains persist at scale and could beat heuristic baselines.

\subsection{Experiment Settings}
\label{subsec:experiment-settings}
\paragraph{Personalized assortment allocation.}
The retailer observes each customer's type, offers an assortment of in-stock products, and the customer chooses by multinomial logit. The small instance has nine periods, a scarce high-revenue product (revenue $10$, three units), an abundant low-revenue product (revenue $4$, six units), and three customer types: one prefers the high-revenue product, one likes both, and one prefers the low-revenue product. The large instance has ten products, six customer types, and ten periods. The baseline is Inventory Balancing (IB) of \citet{golrezaei2014real}, which discounts each product's revenue by the fraction of its inventory that remains.

\paragraph{Online linear programming.}
Each request is either served with one of two options, each with a reward and resource requirements, or rejected. The small instance has six periods and two resources with three units each. Two request types are \emph{flexible}: a lucrative option uses two units of one resource and a cheaper option one unit of the other. The third, \emph{hybrid} type needs the first resource, which we call the \emph{core resource}, in both of its options. The large instance has ten resources, twenty request types with twenty options each, and ten periods. The baseline is Dual Mirror Descent (DMD) of \citet{balseiro2023best}, which prices resources by dual variables updated after each request. Appendix~\ref{app:experiment-config} reports all parameters.

In both applications, the admissible type set is fixed within each instance: $\mathcal B_h(s)=\mathcal B$ for every period and reachable inventory state. Even so, request arrivals remain adaptive because the adversary can adapt its type choice to the current resource state, which is shaped by past allocation decisions.

\paragraph{Evaluation.}
On small instances, each allocation policy is evaluated against its exact worst adversary policy computed by backward induction. On large instances, we freeze every allocation policy, including the baselines, and train five adversaries from independent initializations using the same training budget. Each policy--adversary pair is evaluated on independent simulation paths, and we report the mean revenue across the five adversaries.

For the competitive-ratio diagnostic, we randomly generate $10{,}000$ request sequences. On each sequence, we divide the policy's revenue by the corresponding offline optimum and report the smallest ratio. We refer to this sampled minimum as the empirical competitive ratio.

Three diagnostics track whether the training approaches best response: the follower and leader residuals FR and LR, which are the response and allocation gaps in~\eqref{eq:gaps} divided by $|V_\alpha^\star|$, and envelope stability ES, the range of the last $m$ envelope estimates divided by the magnitude of their mean. They are exact for the tabular method and estimated by response probes for the neural method. We select checkpoints at which every applicable diagnostic is at most $1\%$.

\subsection{Experimental Results}
\label{subsec:experimental-results}

Table~\ref{tab:all-results} summarizes the results: Panel~A reports performance and Panel~B the training diagnostics. The competitive ratio in Panel~A is the smallest ratio, over fixed arrival sequences, of a policy's revenue to the offline optimum for that sequence.

\begin{table}[htbp]
\centering
\caption{Revenue, robustness, and training diagnostics by application and instance size.}
\label{tab:all-results}
\begingroup
\fontsize{9.5}{11.5}\selectfont
\setlength{\tabcolsep}{4pt}
\renewcommand{\arraystretch}{1.12}
\newcommand{\expstat}[2]{$\vcenter{\hbox{\shortstack{$#1$\\[-2pt]{\fontsize{7.2}{8}\selectfont\color{black!55}$(#2)$}}}}$}
\begin{tabularx}{\textwidth}{@{}p{0.22\textwidth}p{0.20\textwidth}*{4}{>{\centering\arraybackslash}X}@{}}
\toprule
& & \multicolumn{2}{c}{\textbf{Assortment allocation}}
& \multicolumn{2}{c}{\textbf{Online linear programming}} \\
\cmidrule(lr){3-4}\cmidrule(lr){5-6}
Metric & Policy & Small & Large & Small & Large \\
\midrule
\multicolumn{6}{@{}l}{\textbf{Panel A. Policy performance}} \\[3pt]
Revenue & Robust optimum & $41.2507$ & --- & $20.0000$ & --- \\[2pt]
 & Tabular & \expstat{41.2100}{0.0001} & --- & \expstat{19.9869}{0.0008} & --- \\[2pt]
 & Neural & \expstat{40.2242}{0.9018} & \expstat{6.4300^{\dagger}}{0.0626} & \expstat{19.7924}{0.2236} & \expstat{205.3113^{\dagger}}{15.9716} \\[2pt]
 & Baseline & $38.5738$ & \expstat{6.2195^{\dagger}}{0.0528} & $15.0000$ & \expstat{121.0097^{\dagger}}{17.8686} \\[2pt]
\addlinespace[5pt]
Gain over baseline (\%) & Tabular & \expstat{6.83}{0.00} & --- & \expstat{33.25}{0.01} & --- \\[2pt]
 & Neural & \expstat{4.28}{2.34} & \expstat{3.39^{\dagger}}{1.59} & \expstat{31.95}{1.49} & \expstat{71.25^{\dagger}}{15.22} \\[2pt]
\addlinespace[5pt]
\shortstack[l]{Share of robust\\optimum (\%)} & Tabular & \expstat{99.90}{0.00} & --- & \expstat{99.93}{0.00} & --- \\[2pt]
 & Neural & \expstat{97.51}{2.19} & --- & \expstat{98.96}{1.12} & --- \\[2pt]
\addlinespace[5pt]
\shortstack[l]{Competitive ratio,\\all sequences} & Tabular & $0.8838$ & --- & $0.6200$ & --- \\[2pt]
 & Baseline & $0.8547$ & --- & $0.5882$ & --- \\[2pt]
\addlinespace[3pt]
\shortstack[l]{Competitive ratio,\\sampled sequences} & Neural & $0.9461$ & $0.8824$ & $0.6688$ & $0.4053$ \\[2pt]
 & Baseline & $0.8547$ & $0.8820$ & $0.5938$ & $0.4471$ \\[2pt]
\midrule
\multicolumn{6}{@{}l}{\textbf{Panel B. Training diagnostics}} \\[3pt]
FR (\%) & Tabular & \expstat{0.00}{0.00} & --- & \expstat{0.01}{0.00} & --- \\[2pt]
 & Neural & \expstat{0.02}{0.03} & \expstat{0.09}{0.11} & \expstat{0.00}{0.00} & \expstat{0.00}{0.00} \\[2pt]
 & Baseline & \expstat{0.18}{0.16} & \expstat{0.00}{0.00} & \expstat{0.00}{0.00} & \expstat{0.08}{0.17} \\[2pt]
\addlinespace[5pt]
LR (\%) & Tabular & \expstat{0.08}{0.00} & --- & \expstat{0.19}{0.00} & --- \\[2pt]
 & Neural & \expstat{0.01}{0.02} & \expstat{0.05}{0.12} & \expstat{0.00}{0.00} & \expstat{0.11}{0.25} \\[2pt]
\addlinespace[5pt]
ES (\%) & Tabular & \expstat{0.00}{0.00} & --- & \expstat{0.00}{0.00} & --- \\[2pt]
 & Neural & \expstat{0.59}{0.30} & \expstat{0.66}{0.24} & \expstat{0.45}{0.15} & \expstat{0.08}{0.09} \\[2pt]
 & Baseline & \expstat{0.87}{0.13} & \expstat{0.74}{0.25} & \expstat{0.18}{0.24} & \expstat{0.36}{0.42} \\[2pt]
\bottomrule
\end{tabularx}
\par\vspace{4pt}
\begin{minipage}{\textwidth}
\fontsize{8.5}{10.5}\selectfont
\textit{Notes.} Tabular and Neural are our method with exact values and with neural networks; Baseline is IB in assortment and DMD in online linear programming. Small-instance revenues are exact worst-case revenues;
$^{\dagger}$marks mean revenue across independently trained adversaries. Gain is the relative improvement over the baseline, averaged across training runs. For small instances, CR uses exact expected revenues and DP OPT, with minima taken over all sequences for the tabular comparison and sampled sequences for the neural comparison. Parentheses give standard deviations across training runs; dashes mark results that are unavailable or do not apply.
\end{minipage}
\endgroup
\end{table}

\paragraph{Near-optimal revenue on certified instances.}
On the small instances, exact worst-response evaluation shows that the tabular method attains $99.90\%$ and $99.93\%$ of the robust optimum, and the neural method attains $97.51\%$ and $98.96\%$, respectively.
On the large instances, we report the mean revenue across five independently trained adversaries against each frozen policy. 

\paragraph{Gains are largest where depletion bites.}
Against exact worst responses, our method improves on IB by $6.83\%$ (tabular) and $4.28\%$ (neural) and on DMD by $33.25\%$ and $31.95\%$; on the large instances, the gains are $3.39\%$ and $71.25\%$. They are larger in online linear programming because committing the core resource early strands the remaining capacity (Section~\ref{subsec:operational}).

\paragraph{Robustness costs little when arrivals are exogenous.}
On fixed arrival sequences, the exogenous setting of Section~\ref{subsec:cr}, our policies attain the higher competitive ratio in five of six comparisons. The exception, large online linear programming ($0.4053$ against $0.4471$), is the revenue that protecting capacity costs on sequences that do not exploit it.

\begin{figure}[!h]
\centering
\begin{subfigure}[]{0.48\textwidth}
\centering
\includegraphics[width=\linewidth]{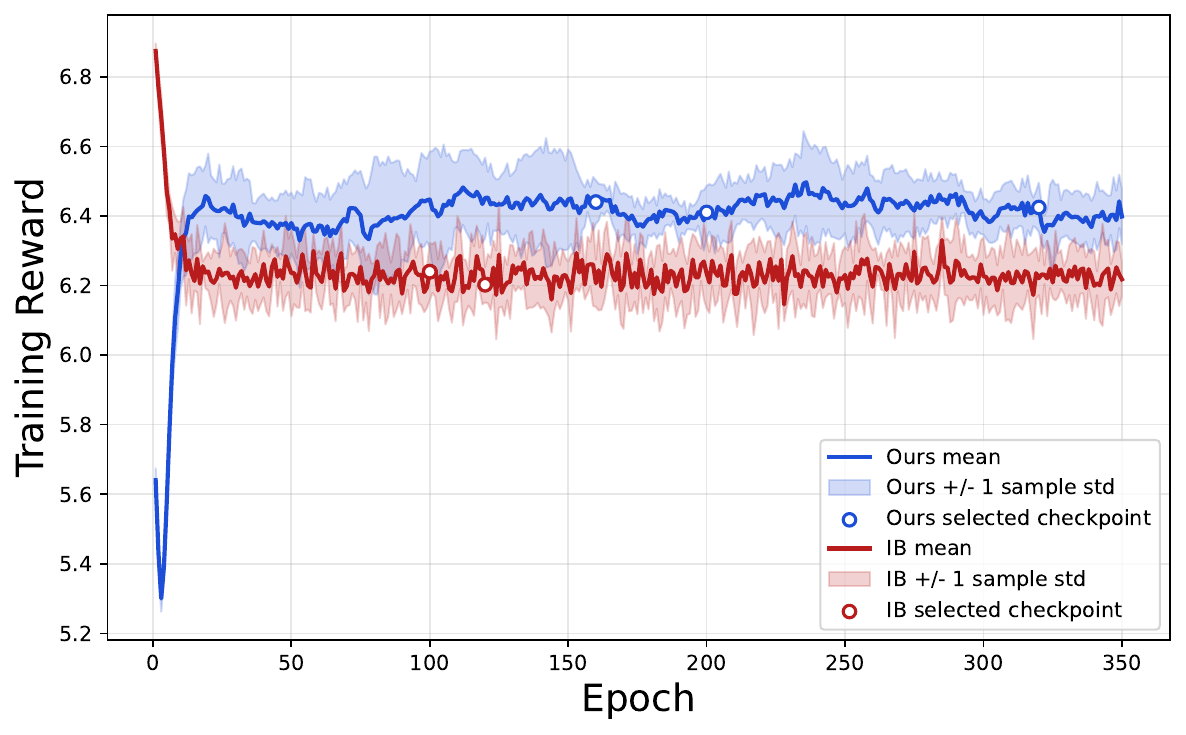}
\caption{Assortment.}
\end{subfigure}
\hfill
\begin{subfigure}[]{0.48\textwidth}
\centering
\includegraphics[width=\linewidth]{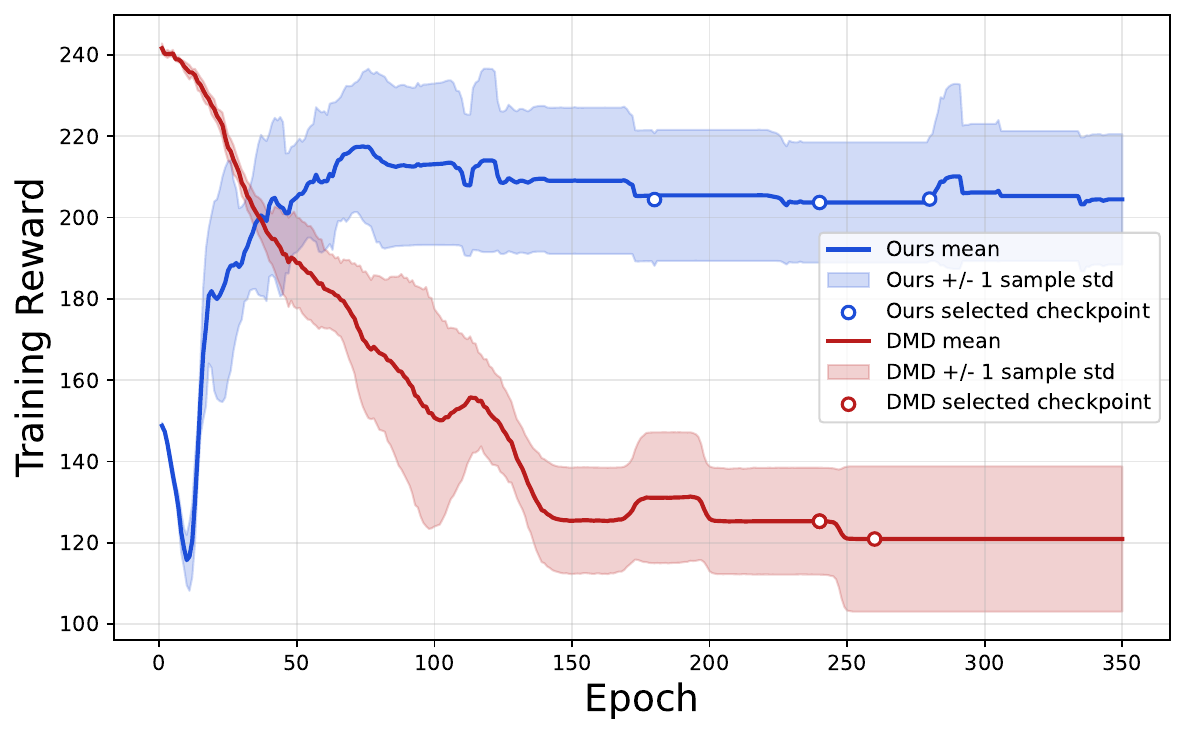}
\caption{Online linear programming.}
\end{subfigure}
\caption{Large instances, neural method: training revenue by outer epoch,
with each policy facing the adversary that learns against it. Lines, bands,
and circles are as in Figure~\ref{fig:exact-training}.}
\label{fig:function-training}
\end{figure}

\begin{figure}[!h]
\centering
\begin{subfigure}[]{0.48\textwidth}
\centering
\includegraphics[width=\linewidth]{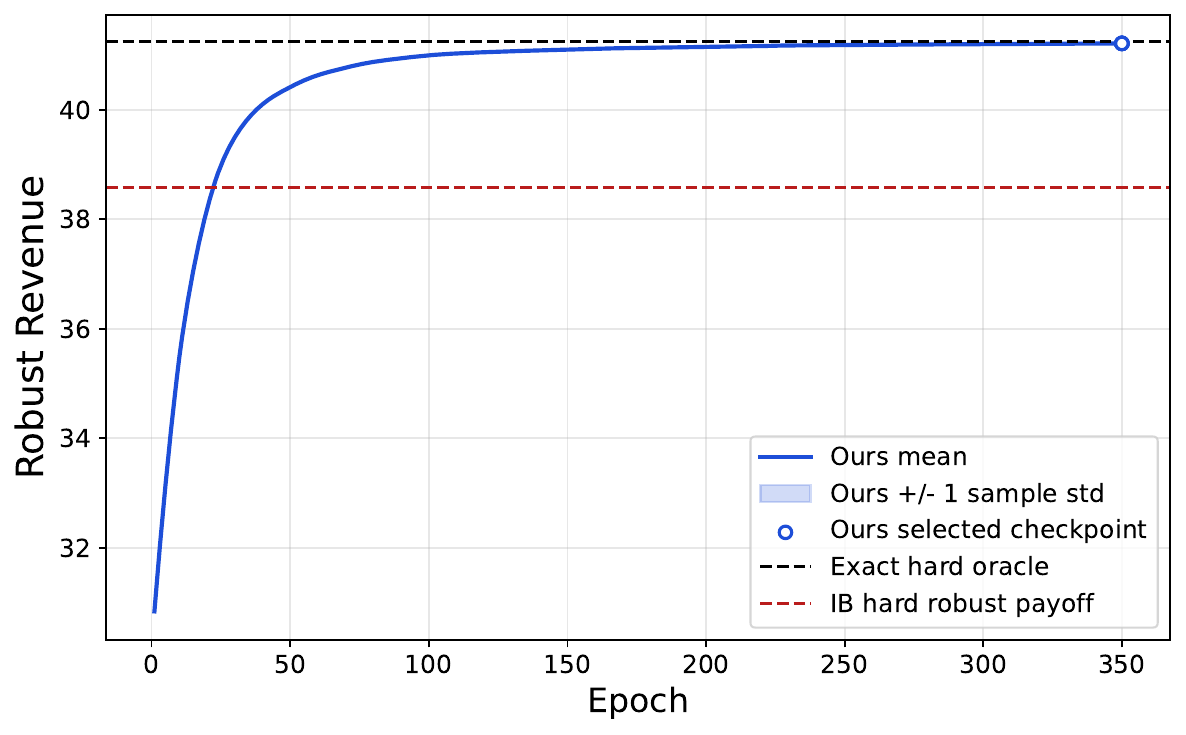}
\caption{Assortment: exact robust revenue.}
\end{subfigure}
\hfill
\begin{subfigure}[]{0.48\textwidth}
\centering
\includegraphics[width=\linewidth]{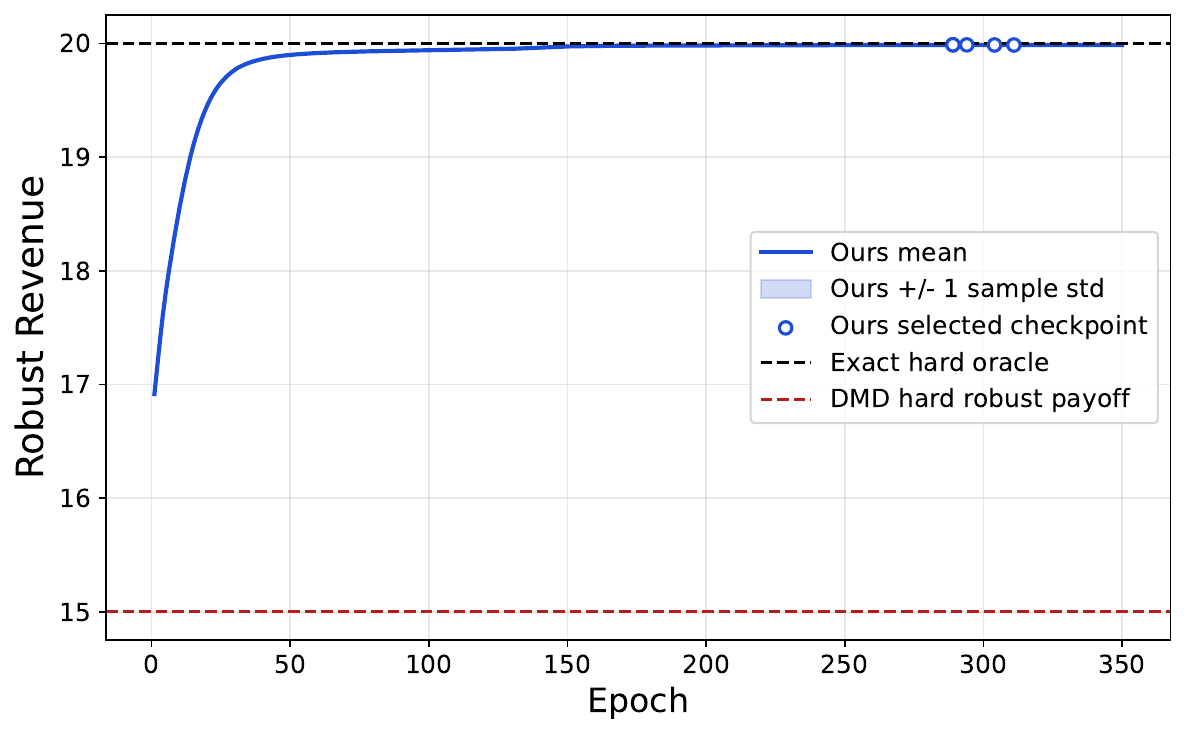}
\caption{Online LP: exact robust revenue.}
\end{subfigure}
\par\medskip
\begin{subfigure}[]{0.48\textwidth}
\centering
\includegraphics[width=\linewidth]{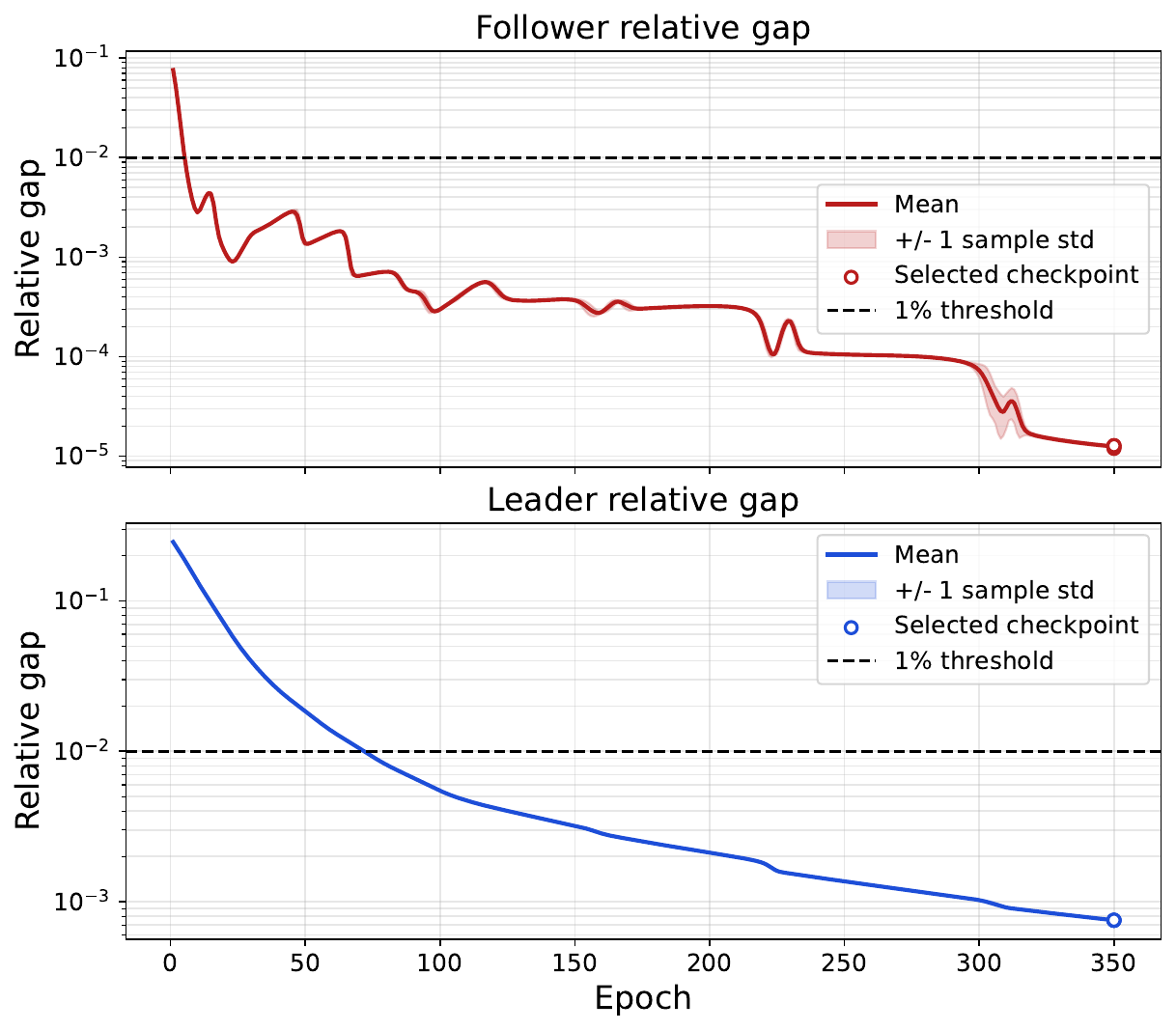}
\caption{Assortment: response and allocation gaps.}
\end{subfigure}
\hfill
\begin{subfigure}[]{0.48\textwidth}
\centering
\includegraphics[width=\linewidth]{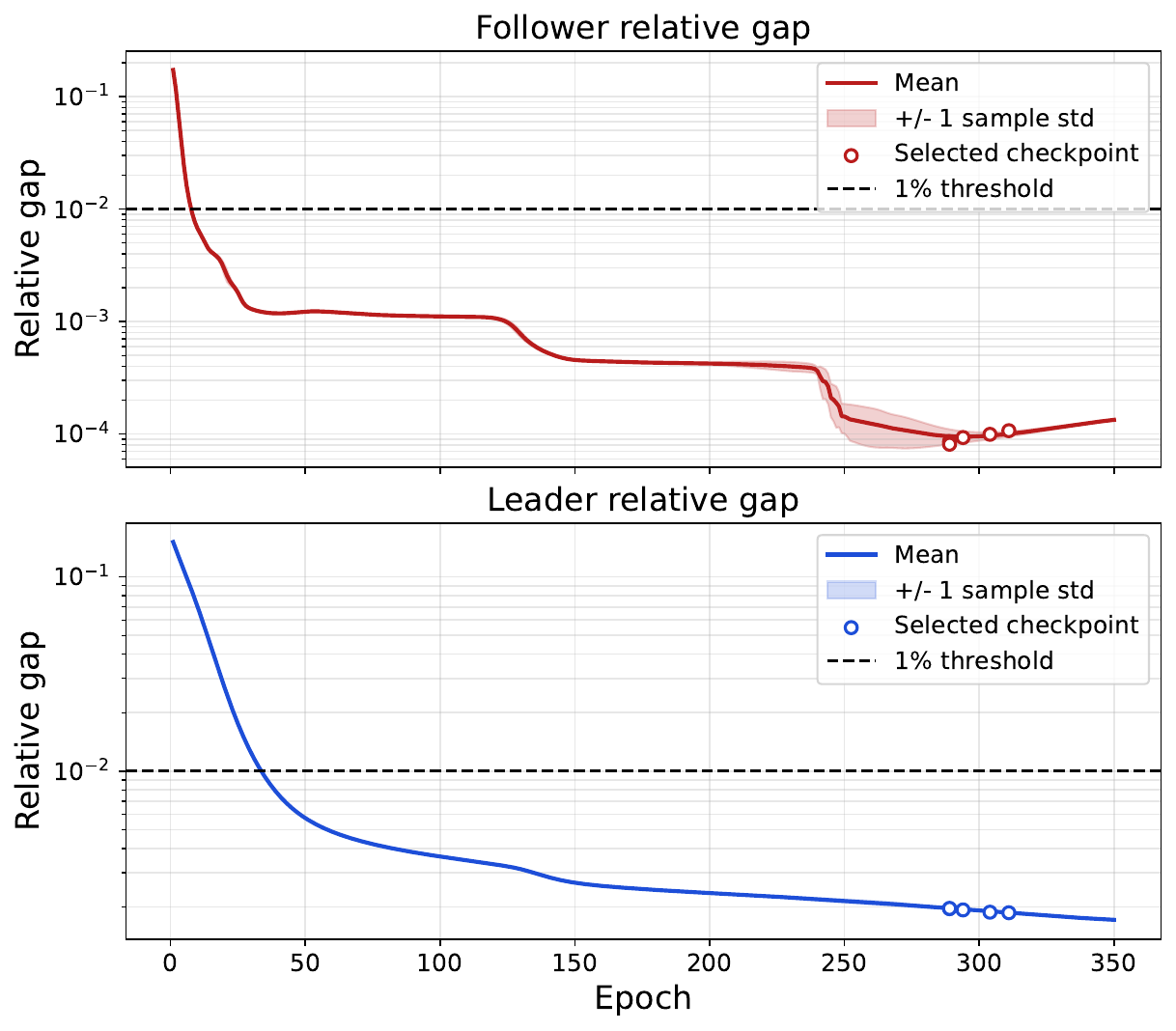}
\caption{Online LP: response and allocation gaps.}
\end{subfigure}
\caption{Small instances, tabular method. (a)--(b) Exact robust revenue of the
current policy by epoch; dashed lines mark the robust optimum (black) and the
baseline's robust revenue (red). (c)--(d) FR (top) and LR (bottom) on a
logarithmic scale, with the $1\%$ threshold dashed. Lines are means across
training runs, bands $\pm1$ standard deviation, and circles selected
checkpoints.}
\label{fig:exact-training}
\end{figure}

The tabular method's exact robust revenue comes within $1\%$ of the optimum in about $100$ epochs in assortment and $25$ in online linear programming, and FR and LR fall below the $1\%$ threshold within about $100$ epochs (Figure~\ref{fig:exact-training}). This verifies that our algorithm converges to the equilibrium.
On the large instances (Figure~\ref{fig:function-training}), the baseline's revenue falls to a lower plateau as its adversary learns, whereas ours dips briefly and then stabilizes above it.

\FloatBarrier

\subsection{Operational Implications}
\label{subsec:operational}

We examine how allocation choices translate into revenue under the request
responses that each policy induces. We simulate each neural policy against an
adversary trained against it and average its behavior by period over
simulated trajectories. Each insight combines a small instance, which reveals
the mechanism, with a large instance, which shows how it plays out over the
horizon.

\begin{figure}[htbp]
\centering
\begin{subfigure}[t]{0.32\textwidth}
\centering
\includegraphics[width=\linewidth]{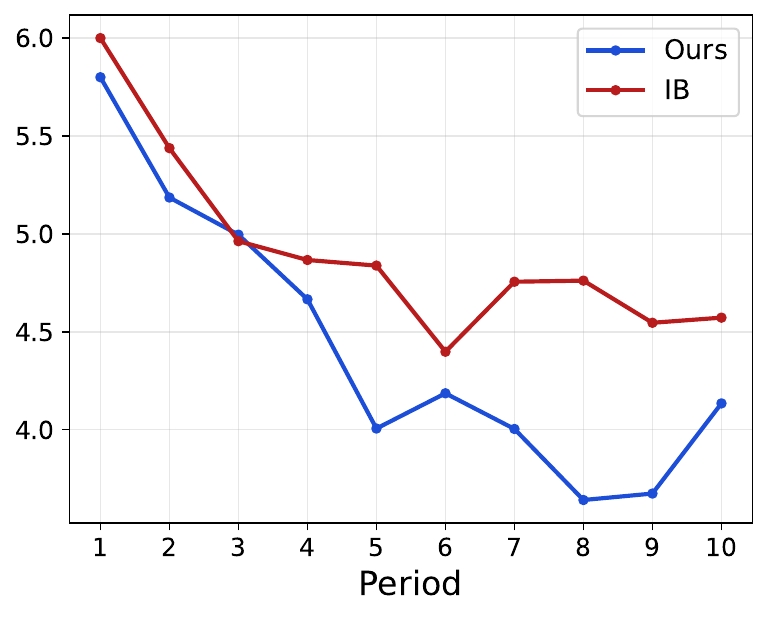}
\caption{Assortment size.}
\label{fig:study2-assortment-size}
\end{subfigure}
\hfill
\begin{subfigure}[t]{0.32\textwidth}
\centering
\includegraphics[width=\linewidth]{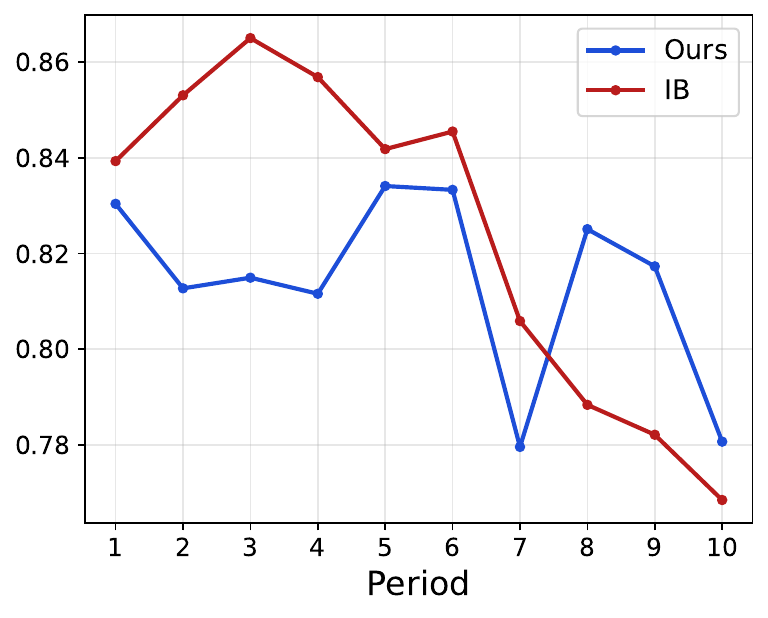}
\caption{Revenue of offered products.}
\label{fig:study2-offered-unit-revenue}
\end{subfigure}
\hfill
\begin{subfigure}[t]{0.32\textwidth}
\centering
\includegraphics[width=\linewidth]{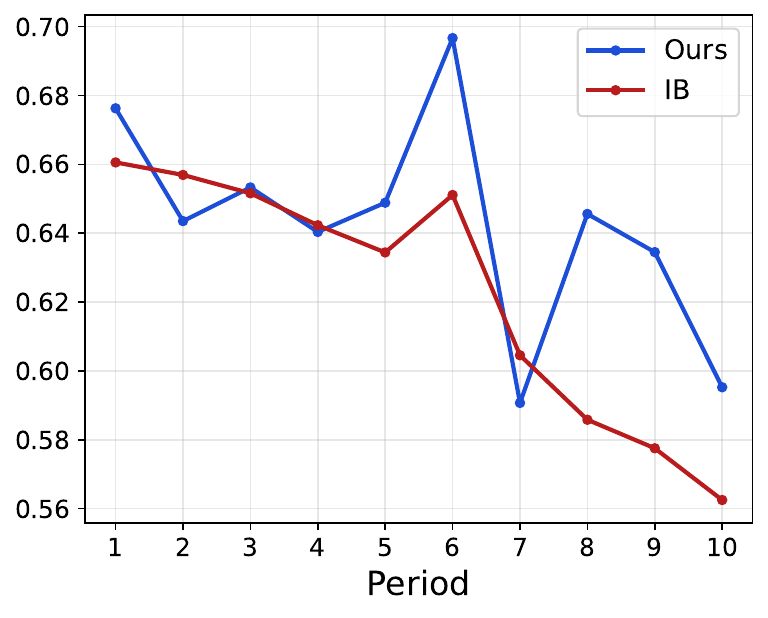}
\caption{Revenue.}
\label{fig:study2-assortment-revenue}
\end{subfigure}
\par\medskip
\begin{subfigure}[t]{0.32\textwidth}
\centering
\includegraphics[width=\linewidth]{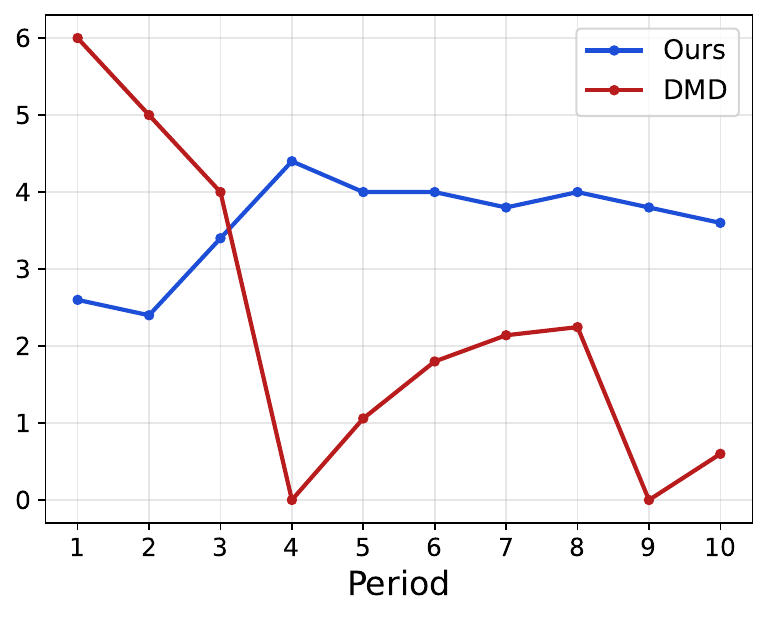}
\caption{Resource consumption.}
\label{fig:study2-resource-consumption}
\end{subfigure}
\hfill
\begin{subfigure}[t]{0.32\textwidth}
\centering
\includegraphics[width=\linewidth]{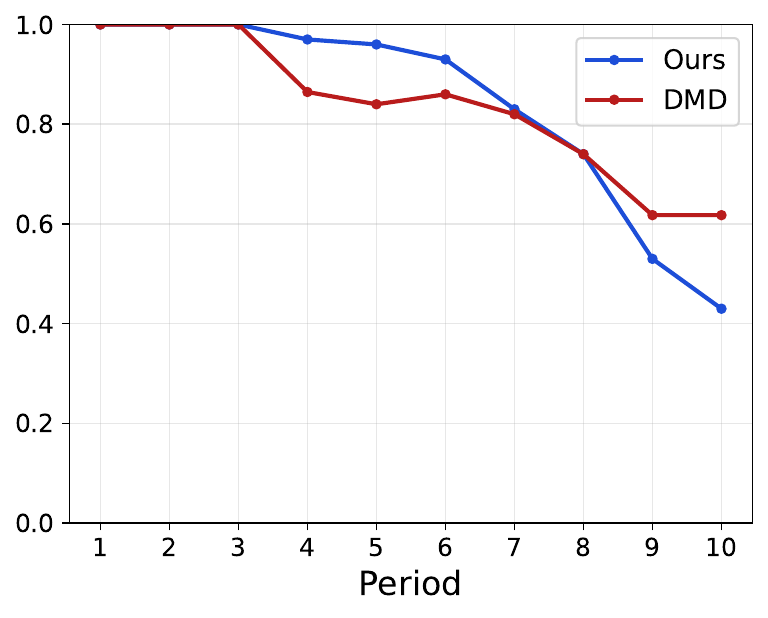}
\caption{Feasible allocation ratio.}
\label{fig:study2-feasible-allocation-ratio}
\end{subfigure}
\hfill
\begin{subfigure}[t]{0.32\textwidth}
\centering
\includegraphics[width=\linewidth]{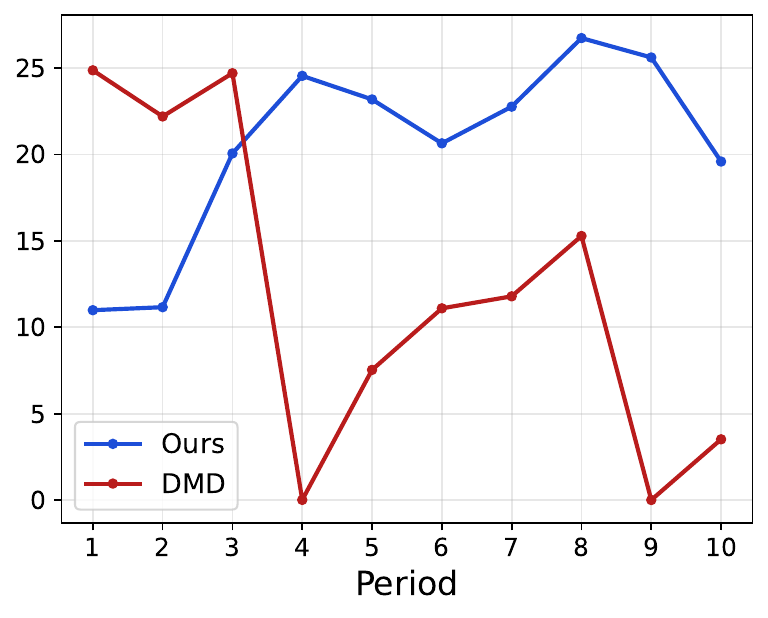}
\caption{Reward.}
\label{fig:study2-olp-reward}
\end{subfigure}
\caption{Large instances: per-period averages for our method (blue) and the
baseline (red), each under its trained adversary. (a)--(c) Assortment;
(d)--(f) online linear programming. The feasible allocation
ratio is the share of options that remaining capacity still permits.}
\label{fig:study2-stage-behavior}
\end{figure}

\paragraph{Insight 1: Assortment composition, not breadth, protects scarce inventory.}
Our policy earns more than IB while offering smaller menus and, early on,
lower-revenue products. In the large assortment instance
(Figure~\ref{fig:study2-stage-behavior}\subref{fig:study2-assortment-size}--\subref{fig:study2-assortment-revenue}), it offers fewer products from
period~4 on, and in periods~1--6 its offered products have lower average unit
revenue. It nevertheless earns as much as IB in total over periods~1--6 and
more in periods~8--10, which account for about three quarters of its revenue
advantage. Neither menu size nor the unit revenue of offered products
therefore explains the gap.


\begin{figure}[htbp]
\centering
\begin{subfigure}[t]{0.96\textwidth}
\centering
\includegraphics[width=\linewidth]{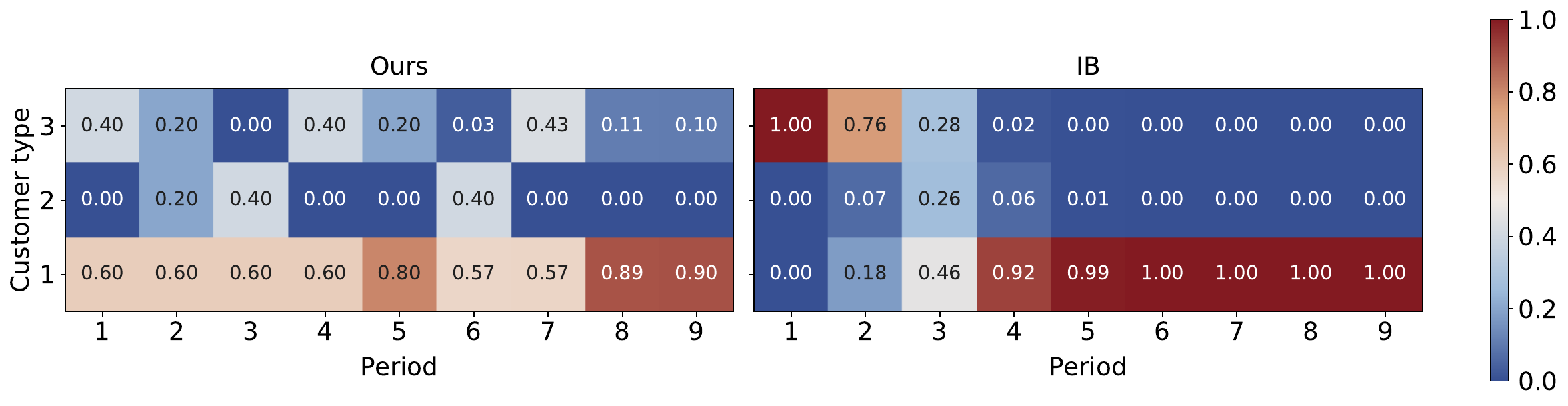}
\caption{Assortment: customer types.}
\end{subfigure}
\par\smallskip
\begin{subfigure}[t]{0.96\textwidth}
\centering
\includegraphics[width=\linewidth]{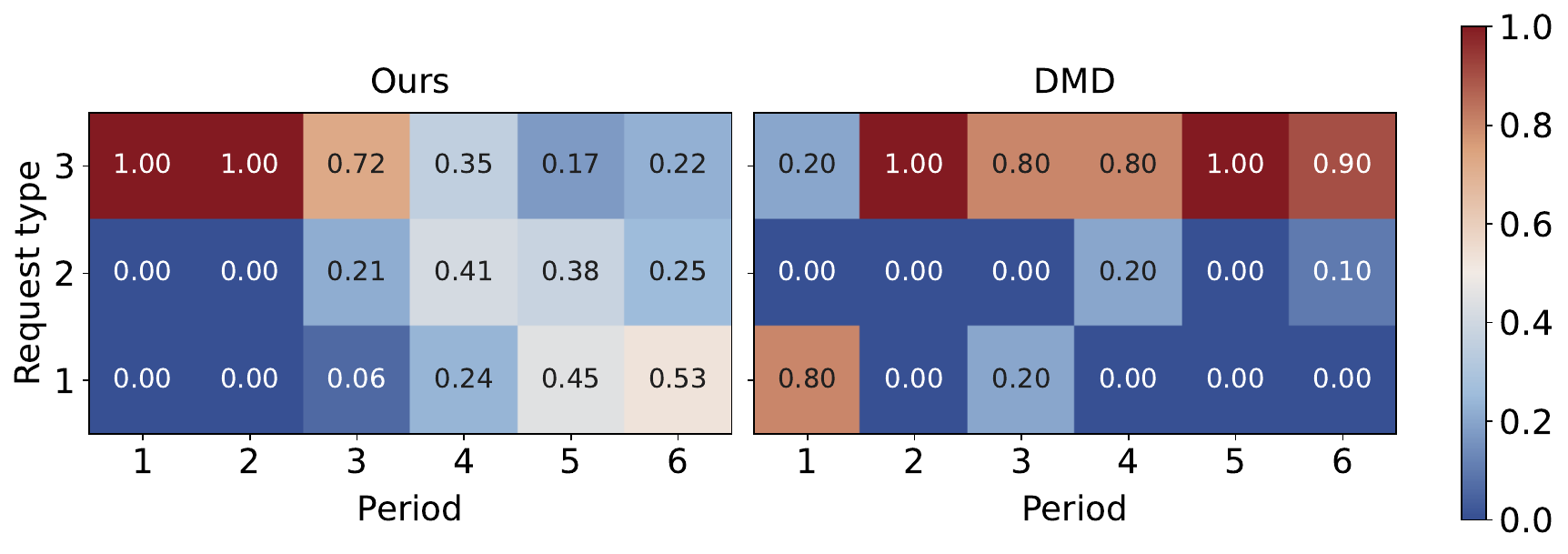}
\caption{Online linear programming: request types.}
\end{subfigure}
\caption{Request types selected by the adversary in the small instances
with neural policies. Each cell gives the frequency of a request type
(row) in a period (column) for our method (left) or the baseline (right).
(a) Type~1 prefers the scarce high-revenue product, type~2 likes both
products, and type~3 prefers the abundant low-revenue product.
(b) Types~1 and~2 are flexible requests. Type~3 is a hybrid request
that needs the core resource under either service option.}
\label{fig:adversarial-arrivals}
\end{figure}

\begin{figure}[htbp]
\centering
\begin{subfigure}[t]{0.96\textwidth}
\centering
\includegraphics[width=\linewidth]{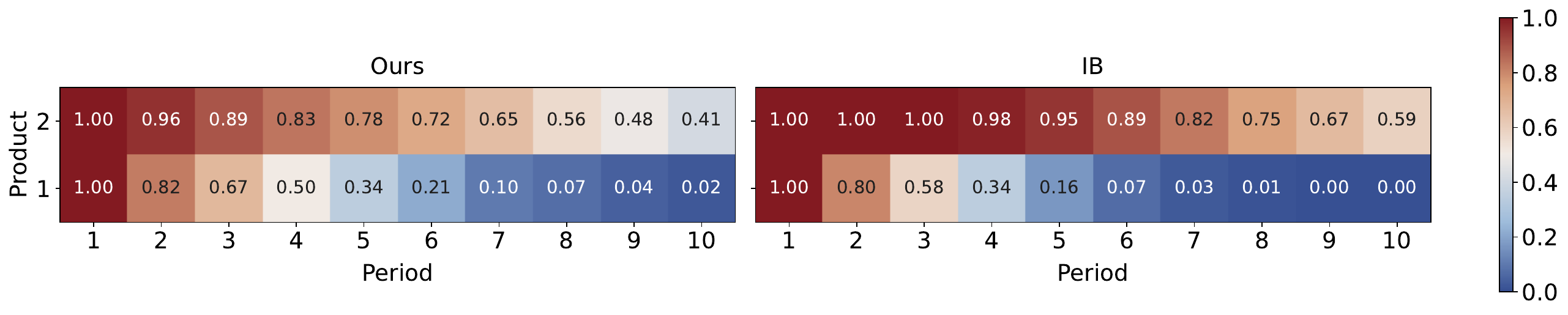}
\caption{Remaining inventory as a fraction of initial inventory.}
\end{subfigure}
\par\smallskip
\begin{subfigure}[t]{0.96\textwidth}
\centering
\includegraphics[width=\linewidth]{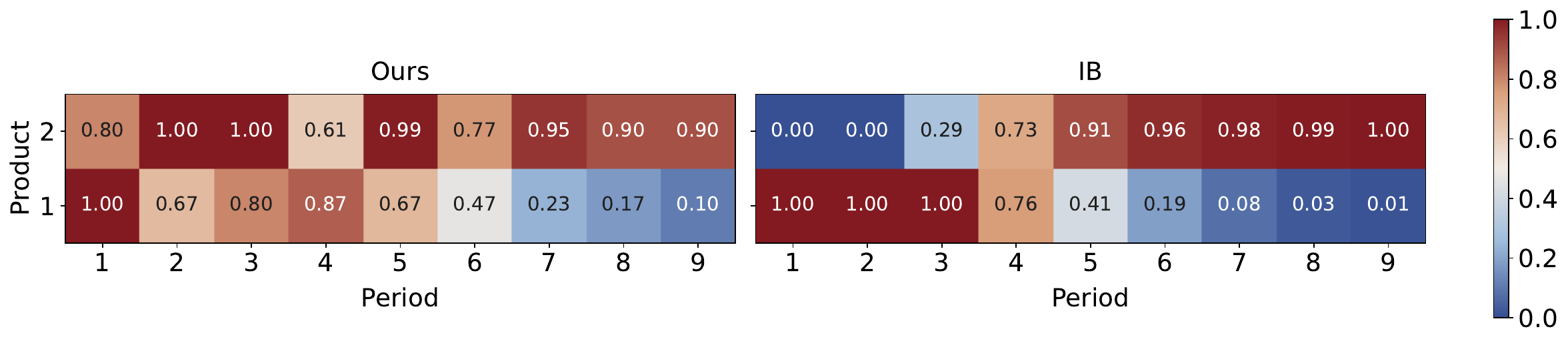}
\caption{Probability that each product is offered.}
\end{subfigure}
\caption{Remaining inventory (a) and offer probabilities (b) by period
in the small assortment instance, for our method (left) and IB (right).
Product~1 is the scarce high-revenue product, and product~2 is the
abundant low-revenue product. The final column in panel~(a) shows
the inventory remaining at the end of the horizon.}
\label{fig:resource-preservation-case1}
\end{figure}

The small instance shows that the explanation is substitution
(Figures~\ref{fig:adversarial-arrivals}a
and~\ref{fig:resource-preservation-case1}). IB offers only the scarce
high-revenue product in periods~1 and~2, and the adversary responds by
sending customers who prefer the abundant low-revenue product. These
customers either take scarce inventory that later customers value more or
leave; the adversary then switches to customers who prefer the scarce
product. Our policy offers the low-revenue product from period~1, so current
purchases shift toward abundant inventory, and $21\%$ of the scarce product
remains at the start of period~6, against $7\%$ under IB. Under MNL choice,
assortment composition determines which product a customer buys: a cheaper
alternative can reduce immediate revenue while preserving a more valuable
future sale. The learned policy uses this substitution deliberately,
retaining scarce products for later request without withholding attractive
offers altogether.

\begin{figure}[htbp]
\centering
\begin{subfigure}[t]{0.96\textwidth}
\centering
\includegraphics[width=\linewidth]{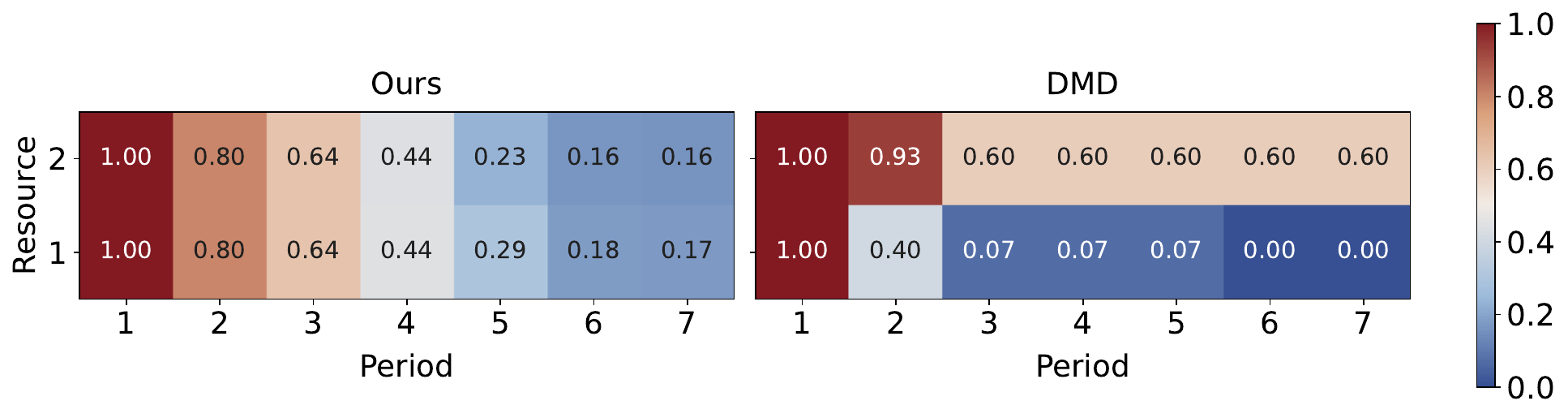}
\caption{Remaining capacity as a fraction of initial capacity.}
\end{subfigure}
\par\smallskip
\begin{subfigure}[t]{0.96\textwidth}
\centering
\includegraphics[width=\linewidth]{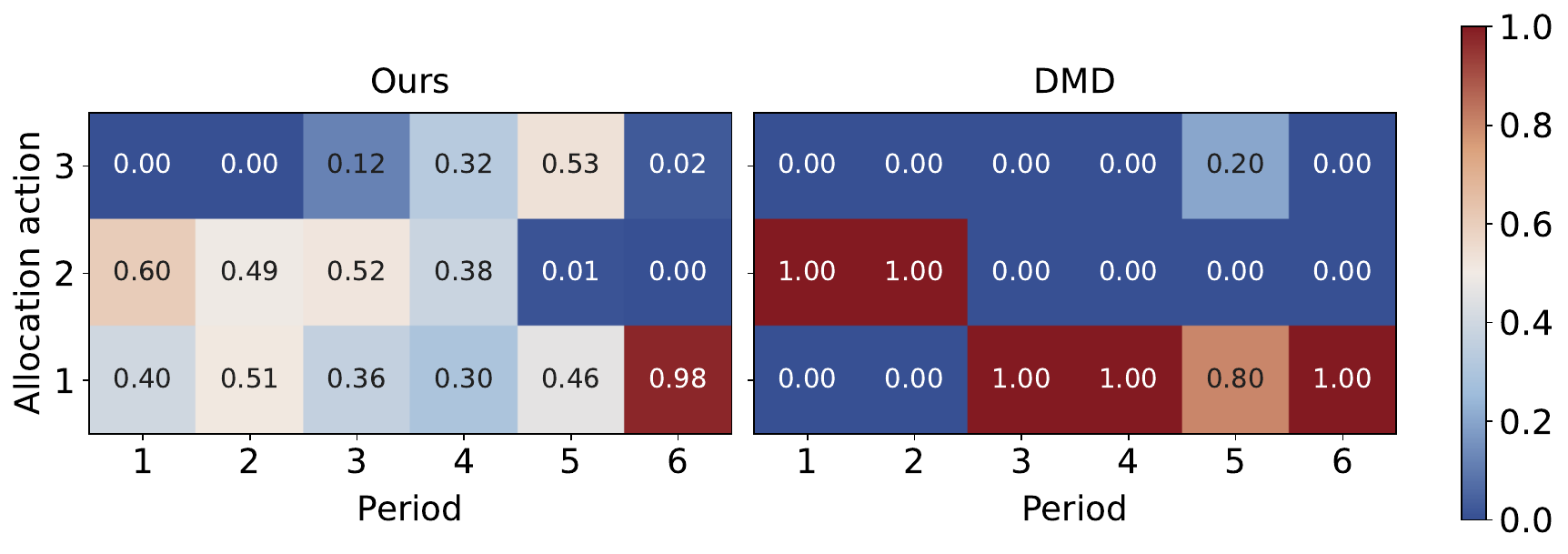}
\caption{Probability of each action.}
\end{subfigure}
\caption{Remaining capacity (a) and action probabilities (b) by period
in the small online linear programming instance, for our method (left)
and DMD (right). Resource~1 is the core resource. Action~1 rejects the
request, and actions~2 and~3 select the first and second service options,
respectively. The final column in panel~(a) shows the capacity remaining
at the end of the horizon.}
\label{fig:resource-preservation-case2}
\end{figure}

\paragraph{Insight 2: Early restraint can increase total capacity utilization.}
Our policy uses less capacity early but more over the horizon than DMD. In
the small online linear programming instance
(Figures~\ref{fig:adversarial-arrivals}b
and~\ref{fig:resource-preservation-case2}), the adversary first sends
flexible requests, which DMD serves with lucrative options that consume the
core resource heavily, and then sends hybrid requests, which need that
resource however they are served. Capacity in the other resource cannot
compensate, and $60\%$ of it is still unused at the end. Our policy rejects
some early requests and spreads accepted consumption across both resources,
preserving the combinations that later requests need.

The large instance separates available capacity from actual utilization
(Figure~\ref{fig:study2-stage-behavior}\subref{fig:study2-resource-consumption}--\subref{fig:study2-olp-reward}). DMD allocates almost nothing in period~4, although about $86\%$ of service options remain feasible, and its allocation falls sharply again in period~9. Physical feasibility therefore cannot explain its inactivity: under DMD's acceptance rule, a feasible option is rejected when its reward does not cover the current dual cost. Our policy gives up about $30$ in revenue during periods~1--3 but gains more than $110$ during periods~4--10. It keeps allocating throughout and ends with a lower feasible allocation ratio, $0.43$ against $0.62$, because more of its capacity has been converted into revenue. Early restraint thus pays off through later utilization, not through a larger stock of unused resources.

\paragraph{Managerial implication.}
Resource protection should be judged by how the protected resources are
eventually used. In assortment, lower-revenue products can channel current
purchases toward abundant inventory and keep scarce products for later
customers. In capacity allocation, keeping options feasible creates value
only if the allocation policy later uses them. Heuristics whose scarcity
signals react to past consumption invite either premature depletion, as
under IB, or prolonged rejection, as under DMD; policies that account for
substitution and resource combinations avoid both.

\FloatBarrier

\section{Conclusion}
We study online resource allocation against an adaptive adversary and develop a
reinforcement learning method that jointly learns allocation and request
policies using one shared critic per epoch. For the entropy-regularized
tabular model, we establish global last-iterate convergence under exact
natural updates and an explicit revenue loss bound under evaluation and
policy update errors. For the neural implementation, these guarantees apply
to the policies the networks represent whenever the critic and update
directions are uniformly accurate. Near a saddle point that the networks
represent exactly, exact natural updates in the network parameters converge
locally at the linear rates of the tabular model, and Bellman residuals,
value-network errors, and Fisher-solve errors certify the required accuracy
on average over visited decisions. Experiments in personalized assortment and
online linear programming show how preserving resource flexibility protects
revenue against the adaptive adversary.
Extending these guarantees to
fixed statewise divergence budgets remains a direction for future research.

\bibliographystyle{informs2014} 
\bibliography{sample} 




\clearpage
\ECSwitch


\ECHead{Reinforcement Learning for Online Allocation against an Adaptive Adversary}

\section{Experimental Configurations}
\label{app:experiment-config}

Tables~\ref{tab:assort-design} and~\ref{tab:olp-design} report the
environment parameters. For each application, the tabular and neural
methods use the same small instance. Each large instance is generated
once and held fixed across methods.
Table~\ref{tab:evaluation-design} summarizes the training and
evaluation procedures.

\begin{table}[htbp]
\centering
\caption{Environment configuration for personalized assortment allocation.}
\label{tab:assort-design}
\begingroup
\small
\renewcommand{\arraystretch}{1.15}
\begin{tabularx}{\textwidth}{@{}p{0.25\textwidth}YY@{}}
\toprule
Parameter & Small instance & Large instance \\
\midrule
Products & 2 & 10 \\
Customer types & 3 & 6 \\
Periods & 9 & 10 \\
Initial inventory & $(3,6)$ & Independently drawn from $\{1,2,3,4\}$ \\
Unit revenues & $(10,4)$ & Independently drawn from $\mathrm{Unif}(0,1)$ \\
Choice model & Multinomial logit & Multinomial logit \\
Attraction parameters & $(v_{z1},v_{z2},v_{z0})$ by type: & Drawn from $\mathrm{Unif}(0,1)$ \\
\quad Type 1 & $(2.131,0.515,0.556)$ & --- \\
\quad Type 2 & $(2.243,1.461,0.194)$ & --- \\
\quad Type 3 & $(0.348,2.238,0.227)$ & --- \\
Candidate assortments & $2^2=4$ & $2^{10}=1{,}024$ \\
Feasibility & \multicolumn{2}{p{0.69\textwidth}}{Products with zero inventory cannot be offered; the empty assortment is allowed.} \\
Baseline & Inventory Balancing & Inventory Balancing \\
\bottomrule
\end{tabularx}
\endgroup
\end{table}

\begin{table}[htbp]
\centering
\caption{Environment configuration for online linear programming.}
\label{tab:olp-design}
\begingroup
\small
\renewcommand{\arraystretch}{1.15}
\begin{tabularx}{\textwidth}{@{}p{0.25\textwidth}YY@{}}
\toprule
Parameter & Small instance & Large instance \\
\midrule
Resources & 2 & 10 \\
Request types & 3 & 20 \\
Options per request & 2 & 20 \\
Periods & 6 & 10 \\
Initial capacity & $(3,3)$ & Independently drawn from $\{4,5,6\}$ \\
Option specification & Reward $r$ and resource consumption $c$ below & Each option uses 1--3 resources; reward depends on total consumption and a random value density \\
\quad Type 1 & $(r,c)=(8,(2,0))$ or $(6,(0,1))$ & --- \\
\quad Type 2 & $(r,c)=(8,(0,2))$ or $(6,(1,0))$ & --- \\
\quad Type 3 & $(r,c)=(7,(1,1))$ or $(5,(1,0))$ & --- \\
Feasibility & \multicolumn{2}{p{0.69\textwidth}}{An option must satisfy all remaining-capacity constraints; rejection consumes no resources and earns zero reward.} \\
Baseline & Dual Mirror Descent; Euclidean updates with step size $1.5$ in the neural comparison & Dual Mirror Descent \\
\bottomrule
\end{tabularx}
\endgroup
\end{table}

\begin{table}[htbp]
\centering
\caption{Training and evaluation configurations.}
\label{tab:evaluation-design}
\begingroup
\small
\renewcommand{\arraystretch}{1.15}
\begin{tabularx}{\textwidth}{@{}p{0.22\textwidth}YYY@{}}
\toprule
Component
& \shortstack[l]{Small instance\\Tabular}
& \shortstack[l]{Small instance\\Neural}
& \shortstack[l]{Large instance\\Neural} \\
\midrule
Policy representation & Tabular allocation and adversary policies & Neural allocation and adversary policies & Neural allocation and adversary policies \\
Value evaluation & Exact backward dynamic programming & Sampled trajectories with fitted value networks & Sampled trajectories with fitted value networks \\
Policy updates & Exact gradients & Proximal policy optimization & Proximal policy optimization \\
Baseline training & Fixed decision policy & Fixed decision rule; train its adversary & Fixed decision rule; train its adversary \\
Payoff evaluation & Exact worst response and robust optimum & Independent out-of-sample evaluation; additional exact worst-response evaluation & Independent out-of-sample evaluation \\
Response diagnostics & Exact FR, LR, and ES & Finite-budget response probes and envelope stability & Finite-budget response probes and envelope stability \\
Selection threshold & $1\%$ & $1\%$ & $1\%$ \\
CR evaluation & Enumerate all arrival sequences & Uniformly sampled arrival sequences & Uniformly sampled arrival sequences \\
Offline benchmark & Exact sequence-specific optimum & Integer sequence-specific optimum & Sequence-specific optimum \\
\bottomrule
\end{tabularx}
\endgroup
\end{table}
\section{Distributionally Robust Extension}
\label{app:kl-extension}

Let $\bar\nu_h(\cdot\mid s)\in\Delta(\mathcal B_h(s))$ be a request
forecast with a positive minimum probability $\bar\nu_{\min}$ over
admissible types at structurally reachable states. For a Markov allocation
policy $\mu$, define the forecast-penalized value
\begin{equation}
\operatorname{Rev}_{\bar\nu}(\mu)
=\min_{\nu\in\mathcal N_{\rm M}}
\left\{J(\mu,\nu)+\alpha\mathbb E_{\rho,\mu,\nu}
\left[\sum_{h=1}^H\operatorname{KL}\!\left(
\nu_h(\cdot\mid s_h)\,\middle\|\,
\bar\nu_h(\cdot\mid s_h)\right)\right]\right\},
\label{eq:forecast-robust-revenue}
\end{equation}
and let $V_{\bar\nu}^*=\sup_{\mu\in\mathcal M_{\rm M}}
\operatorname{Rev}_{\bar\nu}(\mu)$. The minimum is attained because the
product of adversary policy simplices is compact and the objective is
continuous, with $0\log 0=0$.

Evaluating the objective at any admissible Markov response $\nu$ gives
\[
J(\mu,\nu)\ge\operatorname{Rev}_{\bar\nu}(\mu)
-\alpha\mathbb E_{\rho,\mu,\nu}
\left[\sum_{h=1}^H\operatorname{KL}\!\left(
\nu_h(\cdot\mid s_h)\,\middle\|\,
\bar\nu_h(\cdot\mid s_h)\right)\right].
\]
Thus the revenue bound subtracts the forecast divergence accumulated
along trajectories generated by $\mu$ and $\nu$.

\begin{proposition}[Forecast-centered request]
\label{prop:forecast-centered}
Set $r_h^{\bar\nu}(s,a,b)=r_h(s,a,b)-\alpha\log\bar\nu_h(b\mid s)$.
The regularized return of the shifted game is
\[
J_\alpha^{\bar\nu}(\theta,\phi)
=\mathbb E_{\rho,\mu_\theta,\nu_\phi}
\left[\sum_{h=1}^H\left\{
r_h-\alpha\log\mu_{\theta,h}(a_h\mid s_h,b_h)
+\alpha\log\frac{\nu_{\phi,h}(b_h\mid s_h)}
{\bar\nu_h(b_h\mid s_h)}\right\}\right].
\]
Theorem~\ref{thm:sampled} and Proposition~\ref{prop:ridge-accuracy}
apply with absolute reward bound $R_{\max}+\alpha\log(1/\bar\nu_{\min})$.
Compute all value, Fisher, and coverage quantities for the shifted game.

For the shifted-game allocation gap $\Delta_A^{\bar\nu}(\theta)$
defined as in \eqref{eq:gaps},
$V_{\bar\nu}^*-\operatorname{Rev}_{\bar\nu}(\mu_\theta)
\le\Delta_A^{\bar\nu}(\theta)+\alpha H\log A_{\max}$.
On the shifted-game accuracy event of Theorem~\ref{thm:sampled}, for
$0\le t\le T$,
\[
V_{\bar\nu}^*-\operatorname{Rev}_{\bar\nu}(\mu_{\theta_t})
\le\frac{\alpha H}{2}\left(
C_0\varrho_{\rm gl}^{\,t}E_0+\frac{4\varepsilon_{\rm est}}{\alpha}
\right)^2+\alpha H\log A_{\max}.
\]
Here $C_0$ and $E_0$ are computed for the shifted game. If the accuracy
bounds hold simultaneously at every epoch, the limit superior of the
gap is at most $8H\varepsilon_{\rm est}^2/\alpha+\alpha H\log A_{\max}$.
\end{proposition}

\begin{proof}{Proof}
Combining the reward shift with the request entropy term gives
$\alpha\log(\nu_{\phi,h}(b_h\mid s_h)/\bar\nu_h(b_h\mid s_h))$, which
yields the displayed return. Conditional on $s_h=s$, its expectation
is $\alpha\operatorname{KL}(\nu_{\phi,h}(\cdot\mid s)
\|\bar\nu_h(\cdot\mid s))$ by the definition of KL divergence.
The allocation entropy term has conditional expectation
$\alpha\mathcal H(\mu_{\theta,h}(\cdot\mid s,b))$ given $s_h=s$, $b_h=b$.
The shift preserves the game structure, and
$0\le-\log\bar\nu_h(b\mid s)\le\log(1/\bar\nu_{\min})$ gives the stated
absolute reward bound. The convergence and sampling results therefore
apply with the shifted-game constants.

Let $L_{\bar\nu}(\mu,\nu)$ be the objective inside the braces in
\eqref{eq:forecast-robust-revenue} and set $c_A=\alpha H\log A_{\max}$.
Writing $J_\alpha^{\bar\nu}(\mu,\nu)$ for the shifted regularized return
of the policy pair, the bound $0\le\mathcal H\le\log A_{\max}$ gives
\[
0\le J_\alpha^{\bar\nu}(\mu,\nu)-L_{\bar\nu}(\mu,\nu)
=\alpha\mathbb E_{\rho,\mu,\nu}
\left[\sum_h\mathcal H(\mu_h(\cdot\mid s_h,b_h))\right]
\le c_A.
\]
For fixed $\mu_\theta$, set
$h_\alpha^{\bar\nu}(\theta)=\min_\phi J_\alpha^{\bar\nu}(\theta,\phi)$.
For fixed allocation, the request Bellman step minimizes a linear
function of the type probabilities plus $\alpha$ times their KL
divergence from $\bar\nu_h(\cdot\mid s)$. For the finite coefficient $c_b$
of type $b$, the minimizing probability is proportional to
$\bar\nu_h(b\mid s)e^{-c_b/\alpha}$. The minimizer is positive at each
reachable state and is therefore attainable with finite logits.
Minimizing the two bounds over adversary policies therefore gives
$\operatorname{Rev}_{\bar\nu}(\mu_\theta)\le h_\alpha^{\bar\nu}(\theta)
\le\operatorname{Rev}_{\bar\nu}(\mu_\theta)+c_A$.
Continuity of the forecast-penalized payoff and compactness of the request
simplices imply continuity of $\operatorname{Rev}_{\bar\nu}$ in $\mu$.
Positive allocation policies are dense in the full allocation simplices,
so taking the supremum over $\theta$ in the lower bound gives
$\sup_{\theta'}h_\alpha^{\bar\nu}(\theta')\ge V_{\bar\nu}^*$. Subtracting
$\operatorname{Rev}_{\bar\nu}(\mu_\theta)$ yields
\[
\begin{aligned}
V_{\bar\nu}^*-\operatorname{Rev}_{\bar\nu}(\mu_\theta)
&\le \sup_{\theta'}h_\alpha^{\bar\nu}(\theta')
-h_\alpha^{\bar\nu}(\theta)
+h_\alpha^{\bar\nu}(\theta)-\operatorname{Rev}_{\bar\nu}(\mu_\theta)\\
&\le\Delta_A^{\bar\nu}(\theta)+c_A.
\end{aligned}
\]
Applying the allocation-gap bound from Theorem~\ref{thm:sampled} gives
the finite-time assertion. On the all-epoch accuracy event,
$\limsup_{t\to\infty} E_t\le4\varepsilon_{\rm est}/\alpha$, which gives
the stated limit superior.
\end{proof}

To use the extension, add $-\alpha\log\bar\nu_h(b_h\mid s_h)$ to the
critic reward. Tabular request logits may start at
$\phi_{0,0}=\Pi\log\bar\nu$. The same $\alpha$ controls the forecast
penalty and allocation entropy. For a uniform forecast, the shift is
$\alpha\log|\mathcal B_h(s)|$, a policy-independent stage constant if
the menu size depends only on the period $h$.
Fixed statewise KL budgets generally introduce state-dependent dual
penalties \citep{iyengar2005robust,nilim2005robust}. Their learning
guarantees require a separate analysis.

\section{Proofs for Section~\ref{sec:model}}
\label{app:model}
Use the policy classes in Section~\ref{sec:model} and write
$P_hV(s,a,b)=\sum_{s'}P_h(s'\mid s,a,b)V(s')$. Sums run over feasible
choices. Under rectangularity, a type choice at one history imposes no
additional restriction on the choices available at later histories.
This permits backward conditional optimization.

\subsection{Robust MDP representation and Markov policies}
\label{app:model-representation}
\begin{proof}{Proof of Proposition~\ref{prop:lifting}}
Fix a history before the period-$h$ request and set
$\varphi_b(a)=\mu_h(a\mid\mathcal F_h,b)$ and
$\eta(b)=\nu_h(b\mid\mathcal F_h)$. Since adversary selects $b$ before
the firm draws $a$ from $\varphi_b$, the conditional probability
of $(b,a)$ is $\eta(b)\varphi_b(a)$. For any bounded state function $f$,
\[
\mathbb E[f(s_{h+1})\mid\mathcal F_h]
=\sum_{b,a}\eta(b)\varphi_b(a)
  \sum_{s'}P_h(s'\mid s_h,a,b)f(s').
\]
The same averaging for rewards gives \eqref{eq:plan-model}. Conversely,
draw $b\sim\eta$ and then $a\sim\varphi_b$. Under history-dependent policies,
couple the type, action, reward, and next-state draws under the same
physical law. Both constructions reveal the same type and action at
each period, so induction gives identical observable histories and
expected revenue. In both constructions, the adversary selects its type before
the firm draws its realized action from the corresponding plan entry.

At $(h,s)$, a single $\eta$ determines the reward and transition rows
for all plans. Choosing $\eta$ independently at each stage and state
yields a product ambiguity set, which is $s$-rectangular. There are
$\prod_b|\mathcal A_h(s,b)|$ deterministic plans. The optimization below
works directly with actions at each type.
\end{proof}

\begin{proof}{Proof of Remark~\ref{rem:markov-reduction}}
First fix a Markov allocation policy $\mu$. Set $V_{H+1}^\mu=0$ and define
\begin{equation}
V_h^\mu(s)=\min_{b\in\mathcal B_h(s)}
\sum_{a\in\mathcal A_h(s,b)}\mu_h(a\mid s,b)
\{r_h(s,a,b)+P_hV_{h+1}^\mu(s,a,b)\}.
\label{eq:model-fixed-policy-recursion}
\end{equation}
At each $(h,s)$ the type set is finite, so a deterministic Markov
policy can select a minimizing type. Given a history $\mathcal F_h$,
any adversary policy averages the typewise expressions in
\eqref{eq:model-fixed-policy-recursion} with weights
$\nu_h(\cdot\mid\mathcal F_h)$. Thus
$\mathbb E_{\mu,\nu}[r_h+V_{h+1}^\mu(s_{h+1})\mid\mathcal F_h]
\ge V_h^\mu(s_h)$.
Taking expectations and summing over $h$ cancels the intermediate
value terms. Since $V_{H+1}^\mu=0$, this gives
$J(\mu,\nu)\ge\mathbb E_\rho[V_1^\mu(s_1)]$. The deterministic
Markov minimizer attains equality at every stage. Hence
$\inf_{\nu\in\mathcal N_{\rm H}}J(\mu,\nu)
=\min_{\nu\in\mathcal N_{\rm M}}J(\mu,\nu)
=\mathbb E_\rho[V_1^\mu(s_1)]$.

For the optimal allocation problem, fix $(h,s)$ and write
$q_{ba}=r_h(s,a,b)+P_hV_{h+1}(s,a,b)$. The optimal value over
type-contingent plans is
\[
\max_{(\varphi_b)_b}\min_{\eta}
\sum_b\eta(b)\sum_a\varphi_b(a)q_{ba}
=\max_{(\varphi_b)_b}\min_b\sum_a\varphi_b(a)q_{ba}
=\min_b\max_a q_{ba}.
\]
For a fixed plan, the objective is linear in $\eta$, so a point mass
on a minimizing type attains the request minimum. For each type $b$,
$\sum_a\varphi_b(a)q_{ba}\le\max_a q_{ba}$. Assigning probability one
to a maximizing action for each type attains all these upper bounds
simultaneously, which proves the second equality.

Let $V_h$ satisfy \eqref{eq:bellman}. For every observed type choose a
maximizing action $a_h^\star(s,b)$, and let $b_h^\star(s)$ minimize the
resulting typewise maxima. These choices define deterministic Markov
policies $\mu^\star$ and $\nu^\star$. Against any history-dependent request
policy $\nu$, the conditional expectation of $r_h+V_{h+1}(s_{h+1})$
under $\mu^\star$ averages the typewise maxima and is therefore at least
$V_h(s_h)$. Against any history-dependent allocation policy $\mu$,
$\nu^\star$ selects the type whose maximum equals $V_h(s_h)$, so the
corresponding conditional expectation is at most $V_h(s_h)$. Taking
expectations, telescoping, and using $V_{H+1}=0$ yields
\begin{equation}
J(\mu,\nu^\star)\le\mathbb E_\rho[V_1(s_1)]
=J(\mu^\star,\nu^\star)\le J(\mu^\star,\nu),
\quad \mu\in\mathcal M_{\rm H},\ \nu\in\mathcal N_{\rm H}.
\label{eq:model-markov-saddle}
\end{equation}
Consequently $V^*=\mathbb E_\rho[V_1(s_1)]$, attained by deterministic
Markov policies. The argument applies to arbitrary initial
distributions $\rho$ and allows irreversible resource depletion.
\end{proof}

\begin{proof}{Proof of Proposition~\ref{prop:inventory-information}}
Both subgames have continuation value $V_{h+1}$. The firm can attain
$q_h(s,b)$ at every state--type pair with a single allocation policy
that chooses a maximizing action for each $(s,b)$. By definition,
$q_h(s,b)$ also bounds the payoff the firm can guarantee at that pair.

With state observation, the minimum is taken separately at each
realized $s_h$, giving $\E_{\rho_h}[\min_{b\in\cB_h}q_h(s_h,b)]$.
When only $\rho_h$ is known, the type distribution is independent
of $s_h$. The expected payoff is a weighted average of
$\E_{\rho_h}[q_h(s_h,b)]$ over $b\in\cB_h$, so a minimizing pure
type attains the minimum.

Subtracting the two values gives
\[
V_h^{\mathrm{hid}}(\rho_h)-V_h^{\mathrm{obs}}(\rho_h)
=\min_{b\in\cB_h}\E_{\rho_h}\!\left[
q_h(s_h,b)-\min_{b'\in\cB_h}q_h(s_h,b')
\right]\ge 0.
\]
The minimum is attained because $\cB_h$ is finite. Since the integrand
is nonnegative, the gap is zero exactly when some type minimizes
$q_h(s,b)$ at every state with $\rho_h(s)>0$.
\end{proof}

\subsection{Competitive ratios and recovery loss}
\label{app:model-benchmarks}
\begin{proof}{Proof of Proposition~\ref{pro:cr}}
Fix $c>0$, set $\varepsilon=\min\{c/2,1/2\}$, and use
Example~\ref{ex:two}. Any randomized, history-dependent allocation
policy accepts the first request with some probability $p\in[0,1]$.
Acceptance earns $\varepsilon$, leaves zero inventory, and produces
the request sequence $(\varepsilon,1)$. Rejection earns zero and
produces $(\varepsilon,0)$. The offline optimum evaluates each
realized sequence as given, earning $1$ in the first case and
$\varepsilon$ in the second. Averaging these outcomes gives
$J(\mu,\nu)=p\varepsilon$ and
$\mathbb E[\mathrm{OPT}(U^{\mu,\nu})]=p+(1-p)\varepsilon>0$.
When $p>0$, the denominator is at least $p$, so the ratio is at most
$\varepsilon<c$. When $p=0$, the ratio is zero. The same deterministic
Markov adversary policy applies to every allocation policy.
\end{proof}

\begin{proof}{Proof of Proposition~\ref{prop:chase}}
Start with $H-1$ units of one resource. The only request in period
one consumes $H-1$ units if accepted and pays zero. In each later
period, accepting the only request consumes one unit and earns one.
The firm may accept whenever it has sufficient capacity and may
always reject. Transitions are deterministic.

The Markov policy that accepts every feasible request uses all its
capacity in period one, so every continuation, including randomized
recovery, earns zero. The policy that rejects the first request and
accepts the remaining requests keeps $H-1$ units after period one
and earns $H-1$ in the remaining periods. A uniform recovery-loss
bound must therefore be at least $H-1$, which rules out an $o(H)$
bound.
\end{proof}

\section{Analysis of the Learning Framework}
\label{app:algorithm}
We use the policies, centering operator, critic, advantages, and sampled
directions from Section~\ref{sec:framework}. We center each tabular
block over its feasible choices and omit blocks with a single choice.
We write $V$ for state values.

\subsection{Regularization and soft responses}
\label{app:representation}\label{app:tabular-geometry}
\begin{proposition}[Regularization bias]
\label{prop:regularization-bias}
For every allocation policy $\mu_\theta$,
$V^*-\operatorname{Rev}(\mu_\theta)
\le\Delta_A(\theta)+\alpha H\log(A_{\max}B_{\max})$.
\end{proposition}
\begin{proof}{Proof}
Let $J_0$ denote the unregularized return. Conditional averaging of the
log-policy terms adds allocation entropy and subtracts request entropy.
Since $0\le\mathcal H(p)\le\log m$ on an $m$-point simplex,
\[
J_0(\mu,\nu)-\alpha H\log B_{\max}
\le J_\alpha(\mu,\nu)
\le J_0(\mu,\nu)+\alpha H\log A_{\max}.
\]
For a fixed allocation policy, taking the infimum of the upper bound
over request gives $h_\alpha(\theta)\le\operatorname{Rev}(\mu_\theta)
+\alpha H\log A_{\max}$. Taking the request infimum and then the
allocation supremum of the lower bound gives
$V_\alpha^\star\ge V^*-\alpha H\log B_{\max}$.
Combining these inequalities and using
$\Delta_A=V_\alpha^\star-h_\alpha$ proves the result.
The finite-horizon return and entropy extend continuously to the boundary
of the policy simplices, with $0\log0=0$. Since positive softmax policies
are dense, taking the request infimum over positive policies gives the
same value as minimizing over the compact adversary policy space. The
resulting allocation envelope is continuous, so the allocation supremum
is also unchanged by restricting it to positive policies.
\end{proof}

\paragraph{Reachable support and value notation.}
The sets $\mathcal S_h^{\rm reach}$ defined in the main text satisfy
\begin{equation}
\mathcal S_1^{\rm reach}=\operatorname{supp}\rho,\qquad
\mathcal S_{h+1}^{\rm reach}
=\bigcup_{\substack{s\in\mathcal S_h^{\rm reach},\ b\in\mathcal B_h(s)\\
 a\in\mathcal A_h(s,b)}}\{s':P_h(s'\mid s,a,b)>0\}.
\label{eq:structural-state-support}
\end{equation}
For finite logits, the forward recursion
$d_{h+1}^z(s')=\sum_{s,b,a}d_h^z(s)\nu_h(b\mid s)\mu_h(a\mid s,b)
P_h(s'\mid s,a,b)$ sums nonnegative terms over feasible choices. Each
policy factor is strictly positive on those choices. Thus a term is
positive exactly when its predecessor state has positive probability
and its transition probability to $s'$ is positive. Starting from
$d_1^z=\rho$, induction gives
\begin{equation}
d_h^z(s)>0\quad\Longleftrightarrow\quad s\in\mathcal S_h^{\rm reach}.
\label{eq:structural-support-equivalence}
\end{equation}
A uniform lower bound on these positive state probabilities follows
once the logits are bounded along training.

For a policy pair $z$ with finite logits, define the regularized state value
$V_h^z$, the immediate reward plus continuation $U_h^z$, and the
type target $Q_{\nu,h}^z$ by
\begin{equation}
\begin{aligned}
V_h^z(s)&=\mathbb E_z\!\left[\sum_{\ell=h}^H
 \widetilde r_{\alpha,\ell}^z\,\middle|\,s_h=s\right],
&V_{H+1}^z&=0,\\
U_h^z(s,a,b)&=r_h(s,a,b)+P_hV_{h+1}^z(s,a,b),\\
Q_{\nu,h}^z(s,b)&=\sum_a\mu_h(a\mid s,b)
 \{U_h^z(s,a,b)-\alpha\log\mu_h(a\mid s,b)\}.
\end{aligned}
\label{eq:conditional-value-definitions}
\end{equation}
The action value in \eqref{eq:fixedpair} includes both entropy terms,
so $Q_h^z=U_h^z-\alpha\log\mu_h+\alpha\log\nu_h$. Averaging over
allocation at a given state and type gives
$\sum_a\mu_h Q_h^z=Q_{\nu,h}^z+\alpha\log\nu_h$.
When $\widehat Q_t=Q^{z_t}$, the allocation baseline is therefore
$\bar V^\theta_{t,h}(s,b)
=Q_{\nu,h}^{z_t}(s,b)+\alpha\log\nu_{t,0,h}(b\mid s)$.
Averaging once more over the reference adversary policy gives
$\bar V^\phi_{t,0,h}(s)=V_h^{z_t}(s)$.

\begin{proposition}[Soft responses and the regularized saddle]
\label{ass:local}
In the complete finite tabular model with $\alpha>0$, the response to a
finite-logit opponent is unique on structurally reachable decisions
and varies smoothly with the opponent's logits. The regularized game has
a unique saddle on these decisions, with a unique centered representation
$z^\star$. If $|r_h|\le R_{\max}$, the fixed-opponent response logits satisfy
\begin{equation}
\|\theta_\alpha(\phi)\|_\infty,\ \|\phi_\alpha(\theta)\|_\infty
\le\frac{2H}{\alpha}
 \{R_{\max}+\alpha\max(\log A_{\max},\log B_{\max})\}.
\label{eq:tabular-response-logit-bound}
\end{equation}
\end{proposition}
\begin{proof}{Proof}
For a finite vector $q$, define the soft responses
$p_i^\pm=e^{\pm q_i/\alpha}/\sum_j e^{\pm q_j/\alpha}$.
Substituting $\log p_i^\pm=\pm q_i/\alpha-\log\sum_j e^{\pm q_j/\alpha}$
into the definition of KL divergence gives, for every probability vector
$p$,
\[
\begin{aligned}
\sum_i p_i(q_i-\alpha\log p_i)
 &=\alpha\log\sum_i e^{q_i/\alpha}
      -\alpha\operatorname{KL}(p\|p^+),\\
\sum_i p_i(q_i+\alpha\log p_i)
 &=-\alpha\log\sum_i e^{-q_i/\alpha}
      +\alpha\operatorname{KL}(p\|p^-).
\end{aligned}
\]
The KL terms vanish only at $p^+$ and $p^-$, respectively, so each
conditional optimum is unique and strictly positive. At each
backward step, the continuation value is fixed before optimizing the
current policy \citep[Proposition~1 and Theorem~1]{geist2019regularized}.

For fixed allocation $\mu$, set $V_{H+1}^{\mu,\min}=0$. The request
response recursion is
\[
\begin{aligned}
Q_{\nu,h}^{\mu,\min}(s,b)
&=\sum_a\mu_h(a\mid s,b)
 \{r_h(s,a,b)-\alpha\log\mu_h(a\mid s,b)
                   +P_hV_{h+1}^{\mu,\min}(s,a,b)\},\\
V_h^{\mu,\min}(s)
&=-\alpha\log\sum_b\exp\{-Q_{\nu,h}^{\mu,\min}(s,b)/\alpha\},\\
\nu_{\alpha,h}(b\mid s)
&=\exp\{[V_h^{\mu,\min}(s)-Q_{\nu,h}^{\mu,\min}(s,b)]/\alpha\}.
\end{aligned}
\]
For fixed request $\nu$, set $V_{H+1}^{\max,\nu}=0$ and put
$U_h^{\max,\nu}=r_h+P_hV_{h+1}^{\max,\nu}$. The allocation response
uses $\mu_{\alpha,h}(a\mid s,b)\propto
e^{U_h^{\max,\nu}(s,a,b)/\alpha}$. Averaging its conditional optimum
and the request entropy term over types gives
\[
V_h^{\max,\nu}(s)
=\sum_b\nu_h(b\mid s)
 \left\{\alpha\log\sum_a e^{U_h^{\max,\nu}(s,a,b)/\alpha}
                          +\alpha\log\nu_h(b\mid s)\right\}.
\]

For request, the soft-minimum identity applied at $s_h=s$ gives
\[
\mathbb E_{\mu,\nu}
[\widetilde r_{\alpha,h}+V_{h+1}^{\mu,\min}(s_{h+1})
                    -V_h^{\mu,\min}(s_h)\mid s_h=s]
=\alpha\operatorname{KL}(\nu_h\|\nu_{\alpha,h}).
\]
For allocation, the soft-maximum identity at $(s_h,b_h)=(s,b)$
expresses the expected action target net of the log-policy term as
\begin{align*}
    \sum_a\mu_h(a\mid s,b)
 \{U_h^{\max,\nu}(s,a,b)-\alpha\log\mu_h(a\mid s,b)\}\\
=\alpha(\log\sum_a e^{U_h^{\max,\nu}(s,a,b)/\alpha}
 -\operatorname{KL}(\mu_h\|\mu_{\alpha,h})).
\end{align*}
Sum the request identity after taking expectations under $(\mu,\nu)$.
The value terms telescope to
$J_\alpha(\mu,\nu)-\mathbb E_\rho[V_1^{\mu,\min}]$.
Since the response attains the conditional minimum at every state,
$\mathbb E_\rho[V_1^{\mu,\min}]=J_\alpha(\mu,\nu_\alpha)$.
For allocation, add the request log-policy term and average over the fixed
type distribution. Subtracting $V_h^{\max,\nu}$ leaves the negative
allocation KL term, averaged over types. Summing over periods and
using $\mathbb E_\rho[V_1^{\max,\nu}]=J_\alpha(\mu_\alpha,\nu)$ gives
the second identity below.
\begin{equation}
\begin{aligned}
J_\alpha(\mu,\nu)-J_\alpha(\mu,\nu_\alpha)
&=\alpha\sum_{h,s}d_h^{\mu,\nu}(s)
                      \operatorname{KL}(\nu_h\|\nu_{\alpha,h}),\\
J_\alpha(\mu_\alpha,\nu)-J_\alpha(\mu,\nu)
&=\alpha\sum_{h,s,b}d_h^{\mu,\nu}(s)\nu_h(b\mid s)
                      \operatorname{KL}(\mu_h\|\mu_{\alpha,h}).
\end{aligned}
\label{eq:response-kl-identities}
\end{equation}
These are the finite-horizon versions of the soft performance-difference
identities \citep[Lemmas~25--26]{mei2020global}. The first is
\eqref{eq:kl-gap}.

The backward recursions compose smooth functions of the finite opponent
logits, which gives smooth responses. For uniqueness, any optimal
candidate must have zero KL at each decision with positive weight in
\eqref{eq:response-kl-identities}. In period one this forces the candidate
to equal the positive soft response on $\operatorname{supp}\rho$ and,
for allocation, at every feasible type because the fixed adversary policy
is positive there. The positive response and opponent then assign
positive probability to every state in $\mathcal S_2^{\rm reach}$.
Repeating the argument at each period forces equality on the full
reachable support, including for candidates on the boundary.

For the saddle, start from $V_{H+1}^\star=0$ and apply the two conditional
optima at every stage, with allocation chosen conditional on the type.
The resulting recursion is
\[
\begin{aligned}
U_h^\star&=r_h+P_hV_{h+1}^\star,\qquad
Q_{\nu,h}^\star(s,b)=\alpha\log\sum_a e^{U_h^\star(s,a,b)/\alpha},\\
V_h^\star(s)&=-\alpha\log\sum_b e^{-Q_{\nu,h}^\star(s,b)/\alpha},\\
\mu_h^\star(a\mid s,b)&\propto e^{U_h^\star(s,a,b)/\alpha},\qquad
\nu_h^\star(b\mid s)\propto e^{-Q_{\nu,h}^\star(s,b)/\alpha}.
\end{aligned}
\]
With $\mu^\star$ fixed, any adversary policy gives a conditional expected
regularized reward plus continuation at least $V_h^\star$. With
$\nu^\star$ fixed, every allocation policy gives at most $V_h^\star$.
Summing these inequalities over periods telescopes the continuation
terms and proves the saddle inequalities with value
$V_\alpha^\star=\mathbb E_\rho[V_1^\star]$.
If $(\widetilde\mu,\widetilde\nu)$ is another saddle, both pairs attain
$V_\alpha^\star$. Because $\widetilde\mu$ guarantees this value and
$\nu^\star$ caps the payoff at this value,
$J_\alpha(\widetilde\mu,\nu^\star)=V_\alpha^\star$.
The same argument gives $J_\alpha(\mu^\star,\widetilde\nu)=V_\alpha^\star$.
Thus $\widetilde\mu$ is a best response to $\nu^\star$ and
$\widetilde\nu$ is a best response to $\mu^\star$. Fixed-opponent
uniqueness identifies both policies. A positive probability vector $p$
has unique centered logits $\Pi\log p$.

Put $C=R_{\max}+\alpha\max(\log A_{\max},\log B_{\max})$.
The expected regularized reward conditional on a state lies in
$[-C,C]$, since its entropy contribution lies between
$-\alpha\log B_{\max}$ and $\alpha\log A_{\max}$.
Consequently
$\|V_h^{\mu,\min}\|_\infty,\|V_h^{\max,\nu}\|_\infty\le(H-h+1)C$.
The conditional targets have the same bound. In $U_h^{\max,\nu}$, the
immediate reward has absolute value at most $R_{\max}$. In
$Q_{\nu,h}^{\mu,\min}$, its allocation average includes a bonus of at
most $\alpha\log A_{\max}$. Both immediate terms are bounded in absolute
value by $C$, and the continuation contributes at most $(H-h)C$.
Dividing the targets by $\alpha$ and using
$\|\Pi x\|_\infty\le2\|x\|_\infty$ proves
\eqref{eq:tabular-response-logit-bound}.
\end{proof}

\paragraph{Natural directions.}
Differentiating the trajectory likelihood gives a score term for each
decision. A stage-$h$ score $\psi_h^j$ has conditional mean zero given
the history just before the actor chooses. Thus any earlier reward
$Z$ satisfies
$\mathbb E_z[Z\psi_h^j]
=\mathbb E_z[Z\,\mathbb E_z[\psi_h^j\mid\text{decision history}]]=0$.
The direct derivatives of the entropy terms also have zero expectation
because they are sums of $-\alpha\psi_h^\theta$ or
$+\alpha\psi_h^\phi$. The remaining score terms involve only the return
from stage $h$ onward. Conditioning that return on $X_h$ gives
\begin{equation}
\nabla_jJ_\alpha(z)
=\sum_{h=1}^H\mathbb E_z\!\left[
 \psi_h^j\sum_{\ell=h}^H\widetilde r_{\alpha,\ell}^{z}\right]
=\bar S_j(z;Q^z),\qquad j\in\{\theta,\phi\}.
\label{eq:actor-score-identity}
\end{equation}
The same zero-mean property allows a baseline that depends on the
decision history alone. The differentiation is valid because the
expected return is a finite sum of smooth functions of the logits.

\begin{lemma}[Conditional mirror updates]
\label{lem:allocation-fisher}
For a fixed finite table $Q$, exact Fisher preconditioning gives the
allocation direction $\Pi Q$ and the request direction
$\Pi\sum_a\mu_h Q_h$. At $Q=Q^z$, these are the natural gradients of
$J_\alpha$. Each nonempty population Fisher matrix is positive definite
on centered reachable coordinates.
\end{lemma}
\begin{proof}{Proof}
In a decision block with choice probabilities $p$, the score of choice
$i$ is $\mathbf e_i-p$. Write $C(p)=\operatorname{Diag}(p)-pp^\top$.
For any baseline $v$ common to all choices, the score average is
$\sum_i p_i(\mathbf e_i-p)(Q_i-v)=C(p)Q=C(p)\Pi Q$.
The baseline cancels because $\sum_i p_i\mathbf e_i=p$, and centering
preserves the product because $C(p)\mathbf 1=0$. This covariance is also
the Hessian of $L(u)=\log\sum_i e^{u_i}$, since $\nabla L(u)=p$ and
$\nabla^2L(u)=C(p)$ for $p=\operatorname{softmax}(u)$.
Weighting each covariance by the probability of reaching its decision
block gives
\begin{equation}
\begin{aligned}
F_{\theta,h,s,b}(z)
&=d_h^z(s)\nu_{\phi,h}(b\mid s)
 \{\operatorname{Diag}(\mu_h)-\mu_h\mu_h^\top\},\\
F_{\phi,h,s}(z)
&=d_h^z(s)\{\operatorname{Diag}(\nu_h)-\nu_h\nu_h^\top\}.
\end{aligned}
\label{eq:actor-fisher-blocks}
\end{equation}
Here $\mu_h$ is evaluated at $(s,b)$ and $\nu_h$ at $s$.
The allocation score average uses the action values $Q_h(s,a,b)$.
The request score average first conditions on $(s,b)$ and uses the
type values $\sum_a\mu_h(a\mid s,b)Q_h(s,a,b)$, since the adversary selects
$b$ before allocation selects $a$. Multiplying both score averages
by their block probabilities proves \eqref{eq:frozen-score-identity}.
At $Q=Q^z$, \eqref{eq:actor-score-identity} then yields
\begin{equation}
\nabla_\theta J_\alpha(z)=F_\theta(z)\Pi Q^z,\qquad
\nabla_\phi J_\alpha(z)=F_\phi(z)\Pi\sum_a\mu_h Q_h^z.
\label{eq:natural-gradient-identity}
\end{equation}
On any retained block, $v^\top C(p)v=\operatorname{Var}_p(v)>0$
for a nonzero centered vector $v$. Indeed, all feasible choices have
$p_i>0$, so zero variance would force $v$ to be constant and hence zero.
The tabular Fisher is block diagonal, and each reachable block has
positive probability. It is therefore invertible on the retained
centered coordinates. Preconditioning cancels both the block
probability and the conditional covariance. Estimating the blocks
from samples still requires visiting them.

The conditional KL step maximizes
$\lambda\langle u,Q\rangle-\operatorname{KL}(u\|p)$ over
$\sum_i u_i=1$, $u_i\ge0$. The simplex multiplier gives
$\log(u_i/p_i)=\lambda Q_i+c$ at an interior stationary point, with
$c$ common to all choices. Normalization gives
$u_i=p_i e^{\lambda Q_i}/\sum_k p_k e^{\lambda Q_k}$, which is positive.
Strict concavity makes this stationary point the unique maximizer.
Taking centered logarithms yields
$\Pi\log u=\Pi\log p+\lambda\Pi Q$. This proves the allocation step in
\eqref{eq:allocation-mirror} and \eqref{eq:natural-logit-update}.
For request, applying the same calculation to the negative type values
gives the minimizing actor's update.
\end{proof}
\subsection{Conditional performance difference}
\label{app:bilateral-geometry}

\begin{lemma}[Conditional saddle KL identity]
\label{lem:saddle-kl-curvature}
For every finite-logit policy pair $z$ and every structurally reachable
starting pair $(h,s)$,
\begin{equation}
\begin{aligned}
V_h^z(s)-V_h^\star(s)
=\alpha\mathbb E_z\!\left[\sum_{\ell=h}^H\left\{
\operatorname{KL}(\nu_\ell\|\nu_\ell^\star)
-\sum_b\nu_\ell(b\mid s_\ell)
 \operatorname{KL}(\mu_\ell\|\mu_\ell^\star)\right\}
\,\middle|\,s_h=s\right].
\end{aligned}
\label{eq:saddle-weighted-kl}
\end{equation}
Each request KL is evaluated at $s_\ell$ and each allocation KL at
$(s_\ell,b)$. The expectation uses the trajectory law under $z$
starting at $(h,s)$.
\end{lemma}

\begin{proof}{Proof}
Proposition~\ref{ass:local} gives
$U_\ell^\star=\alpha\log\mu_\ell^\star+Q_{\nu,\ell}^\star$ and
$Q_{\nu,\ell}^\star=V_\ell^\star-\alpha\log\nu_\ell^\star$.
Conditioning on the realized state, type, and allocation at stage $\ell$
gives the one-period identity
\[
\begin{aligned}
&\mathbb E_z\!\left[\widetilde r_{\alpha,\ell}^z
 +V_{\ell+1}^\star(s_{\ell+1})-V_\ell^\star(s_\ell)
 \,\middle|\,s_\ell=s',b_\ell=b,a_\ell=a\right]\\
&\qquad = U_\ell^\star(s',a,b)-V_\ell^\star(s')
 -\alpha\log\mu_\ell(a\mid s',b)+\alpha\log\nu_\ell(b\mid s')\\
&\qquad =\alpha\log\frac{\nu_\ell(b\mid s')}{\nu_\ell^\star(b\mid s')}
 -\alpha\log\frac{\mu_\ell(a\mid s',b)}{\mu_\ell^\star(a\mid s',b)}.
\end{aligned}
\]
The first equality uses $U_\ell^\star=r_\ell+P_\ell V_{\ell+1}^\star$.
Averaging over $a$ given $(s',b)$ turns the allocation log-ratio into
the conditional allocation KL. Averaging over $b$ given $s'$ gives
the request KL and weights each allocation KL by $\nu_\ell(b\mid s')$.

By the tower property, conditioning this identity on $s_h=s$ gives
the expected stage-$\ell$ KL contribution in
\eqref{eq:saddle-weighted-kl}. Summing over $\ell=h,\ldots,H$, the
regularized rewards give $V_h^z(s)$ in expectation, and the saddle
values telescope to $-V_h^\star(s)$ because $V_{H+1}^\star=0$.
This proves \eqref{eq:saddle-weighted-kl}.
\end{proof}


\begin{lemma}[Critic linearization at the saddle]
\label{lem:saddle-critic-linearization}
For every structurally reachable $(h,s)$, $DV_h(z^\star)=0$ with
respect to the centered logits. At each feasible $(s,a,b)$,
$Q_h^{z^\star}(s,a,b)=V_h^\star(s)$. For a centered-logit perturbation
$\delta z=(\delta\theta,\delta\phi)$, the critic differential at this
tuple is
\begin{equation}
\begin{aligned}
DQ_h(z^\star)[\delta z]
={}&-\alpha\{\delta\theta_{h,s,b,a}
                   -\langle\mu_h^\star,\delta\theta_{h,s,b}\rangle\}\\
&+\alpha\{\delta\phi_{h,s,b}
                   -\langle\nu_h^\star,\delta\phi_{h,s}\rangle\}.
\end{aligned}
\label{eq:saddle-critic-differential}
\end{equation}
\end{lemma}

\begin{proof}{Proof}
Apply the softmax KL bound used to derive
\eqref{eq:bilateral-continuation-quadratic} to each policy block in
\eqref{eq:saddle-weighted-kl}. Each KL is at most
$\|z-z^\star\|_\infty^2/2$, where the norm is over all retained logit
coordinates. The type weights sum to one, so
$|V_h^z(s)-V_h^\star(s)|
\le\alpha(H-h+1)\|z-z^\star\|_\infty^2$. This quadratic bound gives
$DV_h(z^\star)=0$.

The saddle soft responses give
$Q_h^{z^\star}(s,a,b)=Q_{\nu,h}^\star(s,b)
+\alpha\log\nu_h^\star(b\mid s)=V_h^\star(s)$.
For general $z$, the critic is
$Q_h^z=U_h^z-\alpha\log\mu_h+\alpha\log\nu_h$.
The derivative of $U_h^z=r_h+P_hV_{h+1}^z$ vanishes at the saddle.
For $h<H$, every successor with positive transition probability is
structurally reachable and has $DV_{h+1}(z^\star)=0$. At $h=H$,
$V_{H+1}^z=0$ for every $z$.

If $p=\operatorname{softmax}(x)$, the derivative of
$\log p_i=x_i-\log\sum_c e^{x_c}$ in direction $u$ is
$u_i-\langle p,u\rangle$. Applying this formula to the allocation and
request log-probabilities gives \eqref{eq:saddle-critic-differential}.
\end{proof}
\subsection{Global convergence with a frozen critic}
\label{app:global-convergence}
We work on the simultaneous accuracy event of Theorem~\ref{thm:sampled}.
On this event, the argument is deterministic. The norms below are
supremum norms over structurally reachable tuples or centered decision
blocks. By \eqref{eq:conditional-value-definitions}, the exact reference
critic satisfies
\[
\begin{aligned}
Q_h^{z_t}&=U_h^{z_t}-\alpha\log\mu_t+\alpha\log\nu_{t,0},\\
\sum_a\mu_{t,h}(a\mid s,b)Q_h^{z_t}(s,a,b)
&=Q_{\nu,h}^{z_t}(s,b)+\alpha\log\nu_{t,0,h}(b\mid s).
\end{aligned}
\]

\paragraph{Bounded values and logits.}
Conditional on the state, the allocation entropy contributes between
$0$ and $\alpha\log A_{\max}$, while the request entropy contributes
between $-\alpha\log B_{\max}$ and $0$. Summing these expected reward
bounds from $h$ through $H$ gives
\[
-(H-h+1)(R_{\max}+\alpha\log B_{\max})
\le V_h^z(s)
\le(H-h+1)(R_{\max}+\alpha\log A_{\max}).
\]
For $U_h^z$, the current reward is bounded by $R_{\max}$ in absolute
value. For $Q_{\nu,h}^z$, averaging over allocation adds at most
$\alpha\log A_{\max}$ to the current reward. Combining these current
terms with the bounds on $V_{h+1}^z$ gives
\[
\|U_h^z\|_\infty,\ \|Q_{\nu,h}^z\|_\infty
\le(H-h+1)\{R_{\max}+\alpha\max(\log A_{\max},\log B_{\max})\}.
\]
These bounds use conditional expected entropy and hold for any finite
logits.

\begin{lemma}[Bounded logits under approximate natural updates]
\label{lem:global-boundedness}\label{lem:noisy-invariant}
Under the step conditions of Theorem~\ref{thm:sampled}, define
\begin{equation}
B_0\coloneqq\max\left\{\|z_0\|_\infty,
\frac{2H}{\alpha}
 \{R_{\max}+\alpha\max(\log A_{\max},\log B_{\max})\}
+\frac{2\varepsilon_{\rm est}}\alpha\right\}.
\label{eq:invariant-logit-bound}
\end{equation}
For any $T\ge1$, if the critic and direction accuracy bounds hold at
every evaluation in epochs $0,\ldots,T-1$, then
$\|z_t\|_\infty\le B_0$ for $0\le t\le T$ and
$\|\phi_{t,k}\|_\infty\le B_0$ for $0\le t<T$ and $0\le k\le K$.
The bound also holds along the line segments joining successive
updates in these epochs. If the accuracy bounds hold at every epoch,
so do the logit bounds.
\end{lemma}
\begin{proof}{Proof}
Within each allocation block, the request log-probability is constant
across actions. Centering removes it and gives
$\Pi Q^{z_t}=\Pi U^{z_t}-\alpha\theta_t$.
Substituting into the allocation update gives
$\theta_{t+1}=(1-\alpha\lambda)\theta_t
+\lambda(\Pi U^{z_t}+e_t^\theta)$, where
$e_t^\theta=\Pi(\widehat Q_t-Q^{z_t})
+(\widehat G_t^\theta-G_t^\theta)$.
Since $\|\Pi x\|_\infty\le2\|x\|_\infty$,
$\|e_t^\theta\|_\infty\le2\varepsilon_Q+\varepsilon_{\rm act}
\le2\varepsilon_{\rm est}$.

During the request updates, both $\mu_t$ and $\widehat Q_t$ are fixed,
so the ideal request direction is the same at every inner step.
For the exact critic, centering the second reference identity gives
$\Pi Q_\nu^{z_t}+\alpha\phi_{t,0}$. Summing the first $k$ request steps
yields
\begin{equation}
\phi_{t,k}
=(1-\alpha k\eta)\phi_{t,0}
 -k\eta\Pi Q_\nu^{z_t}+k\eta e_{t,k}^\phi,
\qquad \|e_{t,k}^\phi\|_\infty\le2\varepsilon_{\rm est}.
\label{eq:bilateral-affine-prefix}
\end{equation}
For $k\ge1$, the error is
\[
e_{t,k}^\phi
=-\Pi\sum_a\mu_t(a\mid s,b)(\widehat Q_t-Q^{z_t})
 -\frac1k\sum_{i=0}^{k-1}
       (\widehat G^\phi_{t,i}-G^\phi_{t,i}).
\]
Set $e_{t,0}^\phi=0$. Averaging over allocation preserves the
supremum-norm critic bound, centering costs at most a factor of two,
and averaging the direction errors costs at most $\varepsilon_{\rm act}$.
The damping term uses $\phi_{t,0}$ because the critic retains the
request entropy of the reference policy throughout the epoch.

The allocation update moves a fraction $\alpha\lambda$ of the way
from $\theta_t$ to $(\Pi U^{z_t}+e_t^\theta)/\alpha$. The request
prefix moves a fraction $\alpha k\eta$ of the way from $\phi_{t,0}$ to
$(-\Pi Q_\nu^{z_t}+e_{t,k}^\phi)/\alpha$. The value and error bounds
place both targets in the box of radius $B_0$. The step conditions
$0<\alpha\lambda\le1$ and $0\le\alpha k\eta\le\alpha K\eta\le1$ make
both moves convex combinations. Starting from $\|z_0\|_\infty\le B_0$,
induction keeps all stated iterates inside the box. Convexity also
keeps the joining segments inside the box. Each step uses only the
accuracy bounds before and at that update, which permits the first-exit
argument below.
\end{proof}

The prefactor in the global convergence bound is
\begin{equation}
C_0\coloneqq(1+8B_0)^{2H-1},
\label{eq:global-rate-prefactor}
\end{equation}
where $B_0$ is defined in \eqref{eq:invariant-logit-bound}.

\paragraph{A finite recursion for stagewise errors.}
Let $\epsilon_h^\theta(t)=\|\theta_{t,h}-\theta_h^\star\|_\infty$
and $\epsilon_h^\phi(t)=\|\phi_{t,0,h}-\phi_h^\star\|_\infty$ be the
largest stage-$h$ allocation and epoch-start request logit errors,
respectively. Set either error to zero if the corresponding actor has no
free stage-$h$ coordinates, and write
$\beta=\varepsilon_{\rm est}/\alpha$. Define the continuation error by
\[
D_{h+1}(t)
=\max_{s\in\mathcal S_{h+1}^{\rm reach}}
 |V_{h+1}^{z_t}(s)-V_{h+1}^\star(s)|,
\qquad D_{H+1}(t)=0.
\]
For finite vectors, the standard softmax bound \citep{cen2024fast}
follows directly from Hoeffding's lemma. With
$p=\operatorname{softmax}(x)$ and $a_c=y_c-x_c$,
\[
\begin{aligned}
\operatorname{KL}(p\|\operatorname{softmax}(y))
&=\log\sum_c p_c e^{a_c}-\sum_c p_c a_c\\
&\le\tfrac18(\max_c a_c-\min_c a_c)^2
\le\tfrac12\|x-y\|_\infty^2.
\end{aligned}
\]
Applying Lemma~\ref{lem:saddle-kl-curvature} to the subgame beginning at
$(h+1,s)$ gives
\[
\begin{aligned}
|V_{h+1}^{z_t}(s)-V_{h+1}^\star(s)|
&\le\alpha\mathbb E_{z_t}\!\left[
 \sum_{\ell>h}\left\{
 \operatorname{KL}(\nu_{t,0,\ell}\|\nu_\ell^\star)
 +\sum_b\nu_{t,0,\ell}(b\mid s_\ell)
       \operatorname{KL}(\mu_{t,\ell}\|\mu_\ell^\star)
 \right\}\,\middle|\,s_{h+1}=s\right]\\
&\le\frac\alpha2\sum_{\ell>h}
 \{\epsilon_\ell^\theta(t)^2+\epsilon_\ell^\phi(t)^2\}.
\end{aligned}
\]
Each KL is evaluated at the corresponding realized state and type.
Maximizing over $s$ therefore yields
\begin{equation}
\frac{2D_{h+1}(t)}\alpha
\le\sum_{\ell>h}
 \{\epsilon_\ell^\theta(t)^2+\epsilon_\ell^\phi(t)^2\}.
\label{eq:bilateral-continuation-quadratic}
\end{equation}
The conditional probabilities sum to one, so this bound requires no
lower bound on state visitation probabilities.

Entropy conjugacy at each type also gives the exact identity
\[
Q_{\nu,h}^{z_t}-Q_{\nu,h}^\star
=\sum_a\mu_{t,h}P_h(V_{h+1}^{z_t}-V_{h+1}^\star)
 -\alpha\operatorname{KL}(\mu_{t,h}\|\mu_h^\star).
\]
The transition and allocation averages preserve the bound $D_{h+1}(t)$.
The same softmax inequality bounds the allocation KL by
$\epsilon_h^\theta(t)^2/2$. Centering multiplies each bound by at most
two, which gives
\[
\begin{aligned}
\|\Pi(U_h^{z_t}-U_h^\star)\|_\infty
&\le2D_{h+1}(t),\\
\|\Pi(Q_{\nu,h}^{z_t}-Q_{\nu,h}^\star)\|_\infty
&\le2D_{h+1}(t)+\alpha\epsilon_h^\theta(t)^2.
\end{aligned}
\]
Subtract the saddle targets
$\theta_h^\star=\Pi U_h^\star/\alpha$ and
$\phi_h^\star=-\Pi Q_{\nu,h}^\star/\alpha$ from the affine updates,
take norms, and apply \eqref{eq:bilateral-continuation-quadratic} to obtain
\begin{equation}
\begin{aligned}
\epsilon_h^\theta(t+1)
&\le(1-\alpha\lambda)\epsilon_h^\theta(t)
 +\alpha\lambda\left[
 \sum_{\ell>h}\{\epsilon_\ell^\theta(t)^2
                  +\epsilon_\ell^\phi(t)^2\}+2\beta\right],\\
\epsilon_h^\phi(t+1)
&\le(1-\alpha K\eta)\epsilon_h^\phi(t)
 +\alpha K\eta\left[
 \sum_{\ell>h}\{\epsilon_\ell^\theta(t)^2
                  +\epsilon_\ell^\phi(t)^2\}
 +\epsilon_h^\theta(t)^2+2\beta\right].
\end{aligned}
\label{eq:bilateral-global-perturbed-recursion}
\end{equation}
For a partial request update, replace $K\eta$ by $k\eta$ and the left side
by $\|\phi_{t,k,h}-\phi_h^\star\|_\infty$. The accuracy cap is used
below to bound the comparison levels. Apart from its own damped error,
the allocation block at stage $h$ receives errors from later periods.
The request block also receives the allocation error from the same
period.

\begin{proof}{Proof of Theorem~\ref{thm:sampled}}
\emph{Finite-time control of the triangular recursion.}
Work on the simultaneous accuracy event through epoch $T-1$. Order
the stagewise errors backward from the terminal period as
\[
x(t)=\bigl(
\epsilon_H^\theta(t),\epsilon_H^\phi(t),
\epsilon_{H-1}^\theta(t),\epsilon_{H-1}^\phi(t),
\ldots,\epsilon_1^\theta(t),\epsilon_1^\phi(t)
\bigr).
\]
An absent actor block has error zero. Set
$\gamma_{2(H-h)+1}=\alpha\lambda$ and
$\gamma_{2(H-h)+2}=\alpha K\eta$. Then
\eqref{eq:bilateral-global-perturbed-recursion} becomes, for $0\le t<T$,
\begin{equation}
x_i(t+1)
\le(1-\gamma_i)x_i(t)
+\gamma_i\left\{\sum_{j<i}x_j(t)^2+2\beta\right\},
\qquad 1\le i\le2H.
\label{eq:global-ordered-recursion}
\end{equation}
The step conditions give $0<\gamma_i\le1$.
Lemma~\ref{lem:global-boundedness} and the saddle response bound
\eqref{eq:tabular-response-logit-bound} imply
$\|z_t\|_\infty,\|z^\star\|_\infty\le B_0$, with $B_0$ defined in
\eqref{eq:invariant-logit-bound}. Hence $0\le x_i(t)\le2B_0$ through
epoch $T$.

Define comparison levels by $\bar x_1=2\beta$ and
$\bar x_i=2\beta+\sum_{j<i}\bar x_j^2$. Since
$\beta\le1/(16H)$, induction gives $0\le\bar x_i\le4\beta$ for every
$i\le2H$. The base case is $\bar x_1=2\beta$. If the preceding levels
are at most $4\beta$, then
$\bar x_i\le2\beta+16(i-1)\beta^2\le4\beta$, where the last inequality
uses $i-1<2H$.
Also $B_0\ge2\beta$, so $\bar x_i\le2B_0$.

Let $p_i(t)=(x_i(t)-\bar x_i)_+$, where $(u)_+=\max\{u,0\}$.
For $u,v\in[0,2B_0]$,
$(u^2-v^2)_+\le4B_0(u-v)_+$, because
$u^2-v^2=(u-v)(u+v)$ when $u>v$.
Subtracting the comparison recursion from
\eqref{eq:global-ordered-recursion}, taking positive parts, and using
$1-\gamma_i\ge0$ therefore gives
\begin{equation}
p_i(t+1)
\le(1-\gamma_i)p_i(t)
+4B_0\gamma_i\sum_{j<i}p_j(t).
\label{eq:global-excess-recursion}
\end{equation}
Choose $w_i=(1+8B_0)^{i-1}$, so $w_{2H}=C_0$ by
\eqref{eq:global-rate-prefactor}. The geometric-series identity gives
$4B_0\sum_{j<i}w_j=(w_i-1)/2\le w_i/2$, also when $B_0=0$.
Consequently,
\[
\begin{aligned}
\frac{p_i(t+1)}{w_i}
&\le\left(
1-\gamma_i+
4B_0\gamma_i\frac{\sum_{j<i}w_j}{w_i}
\right)\max_j\frac{p_j(t)}{w_j}\\
&\le\left(1-\frac{\gamma_i}{2}\right)
\max_j\frac{p_j(t)}{w_j}.
\end{aligned}
\]
Taking the maximum over $i$ yields
$\max_i p_i(t+1)/w_i\le\varrho_{\rm gl}\max_i p_i(t)/w_i$, since
$\min_i\gamma_i=\alpha\min\{\lambda,K\eta\}$.
Because $w_i\ge1$ and $p_i(0)\le x_i(0)\le E_0$, iteration gives
\[
E_t=\max_i x_i(t)
\le4\beta+w_{2H}\varrho_{\rm gl}^{\,t}E_0
=\frac{4\varepsilon_{\rm est}}{\alpha}
+C_0\varrho_{\rm gl}^{\,t}E_0,
\qquad 0\le t\le T.
\]
This argument uses accuracy only during the first $T$ epochs.

\emph{Gap bounds and all-time conclusions.}
For any adversary policy $\nu$, the saddle KL identity gives
\[
J_\alpha(\mu_t,\nu)-V_\alpha^\star
\ge-\alpha\sum_{h,s,b}d_h^{\mu_t,\nu}(s)\nu_h(b\mid s)
       \operatorname{KL}(\mu_{t,h}\|\mu_h^\star)
\ge-\frac\alpha2\sum_h\epsilon_h^\theta(t)^2.
\]
The bound is uniform in $\nu$ and therefore remains valid after taking
its infimum. Boundary adversary policies are covered by continuity, with
$0\log0=0$. At the current pair, dropping the nonpositive allocation
KL term gives $J_\alpha(z_t)-V_\alpha^\star
\le(\alpha/2)\sum_h\epsilon_h^\phi(t)^2$.
Combining the lower bound on $h_\alpha(\theta_t)$ with this upper bound
on $J_\alpha(z_t)$ yields
\begin{equation}
\begin{aligned}
\Delta_A(\theta_t)
&\le\frac\alpha2\sum_{h=1}^H\epsilon_h^\theta(t)^2,\\
\Delta_R(z_t)
&\le\frac\alpha2\sum_{h=1}^H
       \{\epsilon_h^\theta(t)^2+\epsilon_h^\phi(t)^2\}.
\end{aligned}
\label{eq:gaps-from-stage-errors}
\end{equation}
Each stagewise error is at most $E_t$, proving the finite-time gap bounds
in \eqref{eq:global-error-floor-main}. All estimates hold on the theorem's
simultaneous accuracy event, which has probability at least $1-\delta$.

If accuracy holds over all epochs, the same argument applies for every
$T$. Since $0<\varrho_{\rm gl}<1$, it gives
$\limsup_{t\to\infty}E_t\le4\varepsilon_{\rm est}/\alpha$ and
\begin{equation}
\limsup_t\Delta_A(\theta_t)
\le\frac{8H}{\alpha}\varepsilon_{\rm est}^2,
\qquad
\limsup_t\Delta_R(z_t)
\le\frac{16H}{\alpha}\varepsilon_{\rm est}^2.
\label{eq:global-damped-error-floor}
\end{equation}
With zero errors, the finite-time estimate gives geometric convergence
$z_t\to z^\star$. The partial-update version of
\eqref{eq:bilateral-global-perturbed-recursion} also gives
$\phi_{t,k}\to\phi^\star$ for each $k\le K$; the joining trajectories
then converge by convexity.
\end{proof}

\begin{proof}{Proof of Corollary~\ref{cor:sampled-revenue}}
Proposition~\ref{prop:regularization-bias} bounds the unregularized revenue
loss at every epoch by
$\Delta_A(\theta_t)+\alpha H\log(A_{\max}B_{\max})$.
For $0\le t\le T$, substitute \eqref{eq:gaps-from-stage-errors} and the
logit bound in \eqref{eq:global-error-floor-main} to obtain
\eqref{eq:sampled-revenue}. Under all-time accuracy, taking $t\to\infty$
gives the stated limiting learning error.
\end{proof}

\paragraph{The $2H$-epoch backward propagation.}
For the admissible full steps $\alpha\lambda=\alpha K\eta=1$,
\eqref{eq:global-ordered-recursion} reduces to
$x_i(t+1)\le2\beta+\sum_{j<i}x_j(t)^2$.
Induction over $i$ gives $x_i(t)\le\bar x_i\le4\beta$ whenever
$i\le t\le T$. For $i=1$, the sum is empty. For the induction step,
if $t+1\ge i$, then $t\ge j$ for every $j<i$, so the preceding claims
bound the right side by $2\beta+\sum_{j<i}\bar x_j^2=\bar x_i$.
Thus, on a $T$-epoch accuracy event with $T\ge2H$,
\[
\|z_t-z^\star\|_\infty\le4\varepsilon_{\rm est}/\alpha,
\qquad 2H\le t\le T.
\]
With exact evaluation and ideal directions, $\beta=0$, so the same
induction gives $z_t=z^\star$ by epoch $2H$. Under all-time accuracy,
the respective conclusions hold for every $t\ge2H$.

In the exact case, terminal allocation reaches its saddle value first,
followed by terminal request, then allocation and request one period
earlier. Demand takes its inner steps before the allocation update
within each epoch, but its target uses the allocation policy at the
start of the epoch because the critic is frozen. The $2H$ bound is
measured in epochs of the ideal output-logit iteration.

\subsection{Reference critics and direction estimation}
\label{app:sampled-construction}\label{app:direction-estimation}
Section~\ref{subsec:sampled} defines the critic regression and the
sampled Fisher system. We first fix the closed centered-logit box
$\mathcal Z$ of Lemma~\ref{lem:global-boundedness}, using the maximal
admitted tolerance $\alpha/(16H)$ before drawing any data. The same
arguments apply on any prescribed compact convex region of finite
logits. Let $c_0$ be the minimum feasible choice probability on
$\mathcal Z$, and define the minimum probability of a reachable tuple by
\begin{equation}
\zeta_0:=
\min_{\substack{z=(\theta,\phi)\in\mathcal Z,\ 1\le h\le H\\
 s\in\mathcal S_h^{\rm reach},\ b\in\mathcal B_h(s),\
 a\in\mathcal A_h(s,b)}}
d_h^z(s)\nu_{\phi,h}(b\mid s)\mu_{\theta,h}(a\mid s,b)>0.
\label{eq:tuple-probability-floor}
\end{equation}
Each retained tuple has a continuous probability that is positive at
every finite $z$ by \eqref{eq:structural-support-equivalence}. The same
holds for each feasible choice. Finiteness and compactness then give
$c_0,\zeta_0>0$. This argument allows $\rho$ to lack full support and
the inventory process to be nonrecurrent. The probability floors can
be small and may deteriorate with the horizon.

With $|R_h|\le R_{\max}$, set
\begin{equation}
Q_{\max}=H\{R_{\max}+\alpha\log(1/c_0)\},\qquad
Q_{\max,h}=\frac{H-h+1}{H}Q_{\max},\qquad Q_{\max,H+1}=0.
\label{eq:projection}
\end{equation}
Each feasible log-probability lies in $[\log c_0,0]$. The two entropy
terms enter the reward with opposite signs, so their sum has absolute
value at most $\alpha\log(1/c_0)$. Hence $|Q_h^z|\le Q_{\max,h}$.
We clip fitted values to $[-Q_{\max,h},Q_{\max,h}]$. Their conditional
averages and baselines then lie in the same interval.

\paragraph{The regression target.}
Let $\mathcal F^Q_{t,h}$ record the preceding epochs, the reference
pair $z_t$, and the downstream fitting data. Conditional on this
information, $\widehat Q_{t,h+1}$ is fixed and the stage-$h$ paths are
drawn from $z_t$ using a fresh batch. The target in \eqref{eq:targets}
therefore satisfies
\begin{equation}
\mathbb E[Y_{t,h}\mid\mathcal F^Q_{t,h},X_h=x]
=\mathcal T_{\alpha,h}^{z_t}\widehat Q_{t,h+1}(x),\qquad
\mathcal T_{\alpha,h}^{z}q(x)
=\widetilde r_{\alpha,h}^{z}(x)
 +\mathbb E_z[q(X_{h+1})\mid X_h=x].
\label{eq:conditional-critic-target}
\end{equation}
Linearity of conditional expectation suffices even when revenue and
the next state are dependent. Denote the conditional mean by $m_h(X_h)$.
The cross term vanishes after conditioning on $X_h$ because
$\mathbb E[Y_{t,h}-m_h(X_h)\mid\mathcal F^Q_{t,h},X_h]=0$. Thus
\[
\mathbb E[(q(X_h)-Y_{t,h})^2\mid\mathcal F^Q_{t,h}]
=\|q-m_h\|_{L^2(\zeta_h^{z_t})}^2
 +\mathbb E[(Y_{t,h}-m_h(X_h))^2\mid\mathcal F^Q_{t,h}].
\]
An unrestricted tabular fit minimizes the empirical loss by taking
the mean target in each visited cell. Set the fitted value to zero
in empty cells, which the count event below rules out. Each target
is bounded by $Q_{\max,h}$ because the fitted continuation is
clipped at $Q_{\max,h+1}$. The conditional mean is therefore in the
clipping interval, so clipping cannot increase the absolute error.
Restricted critic classes additionally require approximation and
fitting error bounds.

\paragraph{Score identities and baselines.}
In a block with choice probabilities $p$, the score for choice $c$ is
$\psi=\mathbf e_c-p$. Its inner product with a centered value vector
is the value of choice $c$ minus the $p$-average value. The additive
constant removed by $\Pi$ cancels in this difference. For allocation,
the choice values are the fitted action values. For request, they
are the allocation averages at each type. With the advantage signs
in \eqref{eq:advantages}, these give
\[
\begin{aligned}
(\psi_h^\theta)^\top G^\theta
&=\widehat Q_{t,h}(X_h)-\bar V^\theta_{t,h}(s_h,b_h)
 =\widehat A^\theta_{t,h}(X_h),\\
(\psi_h^\phi)^\top G^\phi
&=\bar V^\theta_{t,h}(s_h,b_h)-\bar V^\phi_{t,k,h}(s_h)
 =-\widehat A^\phi_{t,k,h}(s_h,b_h).
\end{aligned}
\]
Multiplying by the corresponding score and taking expectations gives
\eqref{eq:frozen-score-identity}. Using the same observations for the
score and Fisher estimates gives the pathwise identity below. The
critic outputs and baselines are fixed when differentiating.

\paragraph{The accept/reject calculation.}
For a one-unit accept/reject decision evaluated with $Q^{z_t}$, let
$p=\mu_h(\mathrm{accept}\mid s,b)$ and $\ell=\log\{p/(1-p)\}$.
Acceptance earns $q_b$ and reduces inventory from $s$ to $s-1$, so
its value relative to rejection is
$\Delta Q=q_b-\{V_{h+1}^{z_t}(s)-V_{h+1}^{z_t}(s-1)\}-\alpha\ell$.
The last term is the allocation log-probability difference. The request
log-probability cancels because the request type is the same for both
choices. Multiplying the centered action values by the Fisher block in
\eqref{eq:actor-fisher-blocks} gives the score mean
\begin{equation}
\bar S_{\theta,h,s,b}
=d_h^z(s)\nu_{\phi,h}(b\mid s)p(1-p)
 \begin{pmatrix}\Delta Q\\-\Delta Q\end{pmatrix},\qquad
G^\theta_{h,s,b}=\frac12\begin{pmatrix}\Delta Q\\-\Delta Q\end{pmatrix}.
\label{eq:bid-price-natural-update}
\end{equation}
Taking the difference of the two logit increments yields
\begin{equation}
\begin{aligned}
\ell_{\rm ordinary}^{+}
&=\ell+2\lambda d_h^z(s)\nu_{\phi,h}(b\mid s)p(1-p)\Delta Q,\\
\ell_{\rm natural}^{+}&=\ell+\lambda\Delta Q.
\end{aligned}
\label{eq:logodds-updates}
\end{equation}
At the regularized saddle the natural direction vanishes. Setting
$\Delta Q=0$ therefore gives the soft bid-price formula in the main-text
remark.

\subsection{Finite sample accuracy}
\label{app:sampled-accuracy}

Let $N_Q$ count reachable stage--tuple cells, and let $\zeta_0$ be the
probability floor in \eqref{eq:tuple-probability-floor} on the fixed logit
box of Lemma~\ref{lem:global-boundedness}. At an actor evaluation $z$, write
$\sigma_j=\lambda_{\min}(F_j(z))$ on centered reachable coordinates,
$m_\theta=A_{\max}$, $m_\phi=B_{\max}$, and
$L_t=\log(8N_Q/\delta_t)$ for $\delta_t\in(0,1)$.

\begin{proposition}[Sampling cost of an epoch]
\label{prop:ridge-accuracy}
In the complete tabular model, suppose the reference and evaluated
policies have logits in $\mathcal Z$. Fit the critic by clipped cell means
using the bounds $Q_{\max,h}$ in \eqref{eq:projection}, with fresh
independent paths for every batch. Each actor uses the same paths for
its score and Fisher estimates and computes the conditional advantages
\eqref{eq:neural-ppo-advantages} exactly. For positive tolerances
$\varepsilon_Q,\varepsilon_{\rm act}$ and $Q_{\max}>0$, suppose
\begin{equation}
\begin{aligned}
B_{Q,t}&\ge\frac{4L_t}{\zeta_0}
 \max\left\{2,\frac{H^2Q_{\max}^2}{\varepsilon_Q^2}\right\},\\
B_{j,t}&\ge\frac{16L_t}{\sigma_j},\qquad
\varepsilon_{\rm rid}\le
 \frac{\sigma_j\varepsilon_{\rm act}}{4Q_{\max}\sqrt{m_j}},
\qquad j\in\{\theta,\phi\}.
\end{aligned}
\label{eq:main-critic-sampling}
\end{equation}
Then $\|\widehat Q_t-Q^{z_t}\|_\infty\le\varepsilon_Q$ with probability
at least $1-\delta_t$, and at each actor evaluation
$\|\widehat G^j-G^j\|_\infty\le\varepsilon_{\rm act}$ with conditional
probability at least $1-\delta_t$. Actors with no free coordinates
require no estimation.
\end{proposition}

\begin{lemma}[Bellman error propagation]
\label{lem:critic-drift}
For zero terminal values, simultaneous residual bounds imply
\begin{equation}
\begin{aligned}
\|\widehat Q_h-\mathcal T_{\alpha,h}^z\widehat Q_{h+1}\|_\infty
&\le\epsilon_h\\
&\Longrightarrow\quad
\|\widehat Q_h-Q_h^z\|_\infty\le\sum_{\ell=h}^H\epsilon_\ell.
\end{aligned}
\label{eq:bellmanchain}
\end{equation}
\end{lemma}
\begin{proof}{Proof}
Set $e_h=\|\widehat Q_h-Q_h^z\|_\infty$. Add and subtract
$\mathcal T_{\alpha,h}^z\widehat Q_{h+1}$ and use the Bellman equation for
$Q_h^z$ to obtain
\[
e_h\le\epsilon_h+
\left\|\mathbb E_z[(\widehat Q_{h+1}-Q_{h+1}^z)(X_{h+1})\mid X_h]
\right\|_\infty\le\epsilon_h+e_{h+1}.
\]
The last inequality uses the supremum-norm bound for conditional
expectation. Backward substitution from $e_{H+1}=0$ proves the claim.
\end{proof}

\begin{lemma}[Paired tabular reconstruction]
\label{lem:paired-tabular-accuracy}
At a fixed tabular evaluation, use the exact conditional advantages and
the same paths for the score and Fisher estimates. Write
$\widehat F_{j,0}=\widehat F_j-\varepsilon_{\rm rid}I$. Whenever
$\widehat F_{j,0}+\varepsilon_{\rm rid}I$ is invertible, the exact solve
satisfies the pathwise identities
\begin{equation}
\widehat S^j=\widehat F_{j,0}G^j,\qquad
\widehat G^j-G^j=-\varepsilon_{\rm rid}
 (\widehat F_{j,0}+\varepsilon_{\rm rid}I)^{-1}G^j.
\label{eq:paired-tabular-estimation}
\end{equation}
In particular, a full-rank solve without a ridge recovers $G^j$ exactly.
\end{lemma}
\begin{proof}{Proof}
For either actor, each active-block score satisfies
$(\psi_h^j)^\top G^j=$ its signed advantage, as established above.
Multiplying by $\psi_h^j$, summing over the same paths and periods, and
dividing by the batch size gives
$\widehat S^j=\widehat F_{j,0}G^j$. Hence
$\widehat G^j=(\widehat F_{j,0}+\varepsilon_{\rm rid}I)^{-1}
\widehat F_{j,0}G^j$. Subtracting $G^j$ gives the second identity.
\end{proof}

\begin{proof}{Proof of Proposition~\ref{prop:ridge-accuracy}}
Condition on the reference pair, previously fitted continuation values,
and preceding updates, as appropriate. Use $L_t$ and the batch conditions
in \eqref{eq:main-critic-sampling}.

\emph{Critic.} At stage $h$, let $n_x$ count the reference paths visiting a
reachable cell $x$, whose probability is $p_x\ge\zeta_0$.
Conditional on $\mathcal F^Q_{t,h}$, this count is binomial. Its lower-tail
bound gives
$\Pr\{n_x<B_{Q,t}\zeta_0/2\mid\mathcal F^Q_{t,h}\}
\le e^{-B_{Q,t}\zeta_0/8}$.
Conditional also on the cell indicators, the selected targets remain
independent, lie in $[-Q_{\max},Q_{\max}]$, and have mean
$m_x=\mathcal T_{\alpha,h}^{z_t}\widehat Q_{t,h+1}(x)$.
On the count event, Hoeffding's inequality therefore gives
\[
\Pr\!\left\{
 \left|\frac1{n_x}\sum_{i:X_h^i=x}Y_{t,h}^i-m_x\right|
 >\frac{\varepsilon_Q}{H}
 \,\middle|\,\mathcal F^Q_{t,h},(\mathbf1\{X_h^i=x\})_i\right\}
\le2\exp\!\left\{-\frac{B_{Q,t}\zeta_0\varepsilon_Q^2}
                         {4H^2Q_{\max}^2}\right\}.
\]
The critic batch condition bounds the total probability of a count or
target error in a cell by
\[
e^{-B_{Q,t}\zeta_0/8}
+2e^{-B_{Q,t}\zeta_0\varepsilon_Q^2/(4H^2Q_{\max}^2)}
\le3e^{-L_t}=\frac{3\delta_t}{8N_Q}.
\]
The bound holds for each fitted continuation. Taking expectations
over the downstream fitting data leaves the same bound conditional on
the reference policy and preceding epochs. A union bound over the $N_Q$
stage--tuple cells gives simultaneous Bellman residuals at most
$\varepsilon_Q/H$ with failure probability at most $3\delta_t/8$.
Clipping cannot increase distance to $m_x$, which lies in the prescribed
stage return interval. Lemma~\ref{lem:critic-drift} now yields
$\|\widehat Q_t-Q^{z_t}\|_\infty\le\varepsilon_Q$.

\emph{Actors.} Set $\widehat F_{j,0}=\widehat F_j-\varepsilon_{\rm rid}I$.
Lemma~\ref{lem:paired-tabular-accuracy} holds for every frozen fitted
critic, so it remains to establish empirical coverage and bound ridge bias.
Let $c$ index a decision--choice cell in an active actor block with at least
two feasible choices. Write $p_c$ for its probability, $\widehat p_c$ for
its empirical frequency, and $\psi_c$ for its score.
The centered covariance bounds in
Appendix~\ref{app:fisher-interpretation} give $\sigma_j\le2p_c$ for
every such cell. There are at most $N_Q$ cells, so a binomial lower-tail
bound and a union bound give
\[
\Pr\{\exists c:\widehat p_c<p_c/2\mid\mathrm{past}\}
\le\sum_c e^{-B_{j,t}p_c/8}
\le N_Qe^{-B_{j,t}\sigma_j/16}
\le\delta_t/8.
\]
On the complementary event, every centered vector $v$ satisfies
\[
v^\top\widehat F_{j,0}v
=\sum_c\widehat p_c(v^\top\psi_c)^2
\ge\tfrac12\sum_c p_c(v^\top\psi_c)^2
=\tfrac12v^\top F_j(z)v.
\]
Thus $\lambda_{\min}(\widehat F_{j,0})\ge\sigma_j/2$.
Within each allocation block, $G^\theta=\Pi\widehat Q_t$. Within each
request block, $G^\phi=\Pi\bar V_t^\theta$. The uncentered entries lie in
$[-Q_{\max},Q_{\max}]$, so orthogonal projection gives
$\|G^j_{\rm block}\|_2\le Q_{\max}\sqrt m$ for a block with $m$ choices.
The matrices are block diagonal. Applying the inverse bound
$\|(\widehat F_{j,0}+\varepsilon_{\rm rid}I)^{-1}\|_{\rm op}
\le(\sigma_j/2+\varepsilon_{\rm rid})^{-1}$ to
\eqref{eq:paired-tabular-estimation}, block by block, gives
\begin{equation}
\begin{aligned}
\|\widehat G^\theta-G^\theta\|_\infty
&\le\frac{\varepsilon_{\rm rid}Q_{\max}\sqrt{A_{\max}}}
              {\sigma_\theta/2+\varepsilon_{\rm rid}},\\
\|\widehat G^\phi-G^\phi\|_\infty
&\le\frac{\varepsilon_{\rm rid}Q_{\max}\sqrt{B_{\max}}}
              {\sigma_\phi/2+\varepsilon_{\rm rid}}.
\end{aligned}
\label{eq:paired-ridge-direction-error}
\end{equation}
For actor $j$, the right side is at most
$2\varepsilon_{\rm rid}Q_{\max}\sqrt{m_j}/\sigma_j
\le\varepsilon_{\rm act}/2$ by \eqref{eq:main-critic-sampling}.
This proves the claimed tolerance and leaves room for an additional
numerical direction error of at most $\varepsilon_{\rm act}/2$ if the
linear system is solved approximately. The proposition itself uses exact
solves.
\end{proof}

The pathwise identity requires the exact allocation-averaged type value
and paired score and Fisher observations. With a sampled type value or
a separate Fisher batch, the score mismatch
$\widehat S^j-\widehat F_{j,0}G^j$ must be controlled separately, as in
the solve certificates below.

\paragraph{Simultaneous accuracy under adaptive updates.}
Proposition~\ref{prop:ridge-accuracy} applies separately to the critic fit
and each actor evaluation. For an epoch failure budget $\delta_t$, use
$\delta_t/2$ in the critic logarithm and $\delta_t/[2(K+1)]$ in each
actor logarithm. Sufficient batch sizes are then
\begin{equation}
\begin{aligned}
B_{Q,t}
&\ge\frac4{\zeta_0}
 \max\!\left\{2,\frac{H^2Q_{\max}^2}{\varepsilon_Q^2}\right\}
 \log\frac{16N_Q}{\delta_t},\\
B_{j,t}
&\ge\frac{16}{\sigma_j(z)}
 \log\frac{16(K+1)N_Q}{\delta_t}.
\end{aligned}
\label{eq:samples}
\end{equation}
The critic bound is uniform over reachable tuples, and the actor bound
is uniform over centered output coordinates.

On the box $\mathcal Z$ fixed before sampling, let
$\underline\sigma_j=\min_{z\in\mathcal Z}\lambda_{\min}(F_j(z))>0$
for each actor with free coordinates. Positive definiteness, continuity,
and compactness justify this minimum. Replacing $\sigma_j(z)$ by
$\underline\sigma_j$ in \eqref{eq:samples} gives fixed sufficient batch
sizes. A fixed ridge satisfying
\[
0<\varepsilon_{\rm rid}\le\frac{\varepsilon_{\rm act}}{4Q_{\max}}
\min\!\left\{
 \frac{\underline\sigma_\theta}{\sqrt{A_{\max}}},\,
 \frac{\underline\sigma_\phi}{\sqrt{B_{\max}}}\right\}
\]
suffices at every solve. The minimum uses only actors with free
coordinates. If
$Q_{\max}=0$, the ideal directions vanish and the ridge restriction is
unnecessary. If neither actor has free coordinates, no actor batches are
needed.

For the adaptive union bound, index batches in sampling order. Let $A_n$
be the event of no previous failure or exit from $\mathcal Z$, and let
$B_n$ be the current accuracy failure. Since $A_n$ is measurable before
batch $n$, its conditional failure budget $\delta_n$ gives
\[
\Pr(A_n\cap B_n)
=\mathbb E[\mathbf1_{A_n}\Pr(B_n\mid\mathrm{past})]\le\delta_n,
\qquad
\Pr\!\left(\bigcup_n(A_n\cap B_n)\right)\le\sum_n\delta_n.
\]
This includes the evaluation that could propose a first exit. On the
complementary event every such evaluation is accurate, and
Lemma~\ref{lem:global-boundedness} keeps its proposed iterate in
$\mathcal Z$. Induction therefore excludes a first exit.
For $T$ epochs take $\delta_t=\delta/T$. For all epochs take
$\delta_t=6\delta/\{\pi^2(t+1)^2\}$, whose sum is $\delta$.
At fixed positive tolerances, \eqref{eq:samples} grows logarithmically
with $t$, with constants depending on coverage and initialization.
Uniform accuracy may require exponentially many paths in the horizon,
because the probability floors can decay exponentially.

For any $0<\varepsilon_{\rm est}\le\alpha/(16H)$, take
$\varepsilon_Q=\varepsilon_{\rm act}=\varepsilon_{\rm est}/2$ in the
complete tabular class. This constructs the simultaneous accuracy event
of Theorem~\ref{thm:sampled}. Zero-error convergence instead concerns
exact evaluation and ideal updates, since these finite-batch bounds
require positive tolerances.

\subsection{Demand budget and fresh batches}
\label{app:bilateral-budget}
The frozen critic and reference allocation fix the ideal request direction
throughout an epoch. Thus the complete request increment is exactly
\[
\phi_{t,K}-\phi_{t,0}
=-K\eta G_t^\phi
 -\eta\sum_{k<K}(\widehat G_{t,k}^\phi-G_t^\phi).
\]
The exact part depends on $K$ only through $K\eta$. The rate comparison in Section~\ref{subsec:sampled} therefore measures
progress per critic fit, rather than per actor trajectory.

Fresh actor batches can reduce sampling fluctuations, but they do not
remove the common error in the frozen critic or systematic fitting bias.
At fixed $K\eta$, splitting a fixed total number of paths among more
inner steps does not by itself reduce the variance of a conditionally
centered estimate. Thus a comparison of $K$ should report total request
trajectories as well as critic fits. Appendix~\ref{app:bilateral-parameterized}
gives a corresponding local calculation for parameterized actors.

\subsection{Proofs for Fisher conditioning and its operational interpretation}
\label{app:fisher-interpretation}

\begin{proposition}[Operational determinants of Fisher conditioning]
\label{prop:sigma}
Let $\mathcal S_h^\star=\{s:d_h^{z^\star}(s)>0,\,
|\mathcal B_h(s)|\ge2\}$. If these sets are not all empty, let
$\sigma_\phi=\lambda_{\min}(F_\phi(z^\star))$ on their centered request
blocks. The type target in \eqref{eq:conditional-value-definitions}
satisfies
$Q_{\nu,h}^\star(s,b)
=\sum_a\mu_h^\star(a\mid s,b)Q_h^{z^\star}(s,a,b)
-\alpha\log\nu_h^\star(b\mid s)$.
Define its spread by
$\Gamma_h(s)=\max_b Q_{\nu,h}^\star(s,b)
-\min_b Q_{\nu,h}^\star(s,b)$. Then
\begin{equation}
\min_{h,s,b}d_h^{z^\star}(s)\nu_h^\star(b\mid s)
\le\sigma_\phi\le
2\min_{h,s,b}d_h^{z^\star}(s)\nu_h^\star(b\mid s),
\label{eq:sigma-explicit}
\end{equation}
where both minima range over $s\in\mathcal S_h^\star$ and
$b\in\mathcal B_h(s)$. Moreover,
\begin{equation}
\frac1{B_{\max}}\min_{h,\,s\in\mathcal S_h^\star}
d_h^{z^\star}(s)e^{-\Gamma_h(s)/\alpha}
\le\sigma_\phi\le
2\min_{h,\,s\in\mathcal S_h^\star}
d_h^{z^\star}(s)e^{-\Gamma_h(s)/\alpha}
\le\frac2{\max_h|\mathcal S_h^\star|}.
\label{eq:sigma-decomposition}
\end{equation}
\end{proposition}

\begin{proof}{Proof of Proposition~\ref{prop:sigma}}
For a positive probability vector $p\in\Delta_m$, $m\ge2$, write
$p_{\min}=\min_c p_c$ and
$C(p)=\operatorname{Diag}(p)-pp^\top$. If $x\perp\mathbf1$, then
\[
\begin{aligned}
x^\top C(p)x
&=\min_a\sum_c p_c(x_c-a)^2\\
&\ge p_{\min}\min_a\sum_c(x_c-a)^2
=p_{\min}\|x\|_2^2.
\end{aligned}
\]
For a coordinate $i$ with $p_i=p_{\min}$, test
$x=\mathbf e_i-\mathbf1/m$. Its squared norm is $(m-1)/m$ and its
variance is $p_{\min}(1-p_{\min})$, so
\[
p_{\min}
\le\lambda_{\min}\!\left(
 \left.C(p)\right|_{\mathbf1^\perp}\right)
\le\frac{m}{m-1}p_{\min}(1-p_{\min})
\le2p_{\min}.
\]
The centered request Fisher is block diagonal, with block
$d_h^{z^\star}(s)C(\nu_h^\star(\cdot\mid s))$. Multiplying the preceding
bounds by the state probability and minimizing over the retained
blocks proves \eqref{eq:sigma-explicit}. Finite saddle logits and
structural reachability give positive choice and state probabilities
on these blocks.

The soft adversary policy satisfies
$\nu_h^\star(b\mid s)\propto
\exp\{-Q_{\nu,h}^\star(s,b)/\alpha\}$.
Subtract the smallest type value and put $m=|\mathcal B_h(s)|$.
The least likely type has the largest value, giving
\[
\min_b\nu_h^\star(b\mid s)
=\frac{e^{-\Gamma_h(s)/\alpha}}
       {\sum_{b'}\exp\{-[Q_{\nu,h}^\star(s,b')
                         -\min_bQ_{\nu,h}^\star(s,b)]/\alpha\}}.
\]
The denominator lies in $[1,m]$, so
$e^{-\Gamma_h(s)/\alpha}/m
\le\min_b\nu_h^\star(b\mid s)
\le e^{-\Gamma_h(s)/\alpha}$.
Combining this with \eqref{eq:sigma-explicit} and $m\le B_{\max}$
proves the first two inequalities in \eqref{eq:sigma-decomposition}.
For each stage with $\mathcal S_h^\star\ne\varnothing$, the state
probabilities over $\mathcal S_h^\star$ sum to at most one; hence
$\min_{s\in\mathcal S_h^\star}d_h^{z^\star}(s)
\le1/|\mathcal S_h^\star|$.
Using $e^{-\Gamma_h(s)/\alpha}\le1$ and choosing a stage of maximum
cardinality proves the final inequality.
\end{proof}

\paragraph{Counting depletion states in assortment.}
Suppose each sale consumes one unit of one product, no-sale is possible,
and every available product can be purchased with positive probability
under some feasible assortment and admissible type. Starting from
$s_1$ with $s_{1,i}\ge H-1$ for every product, the stage-$h$ reachable
inventories are exactly
$\mathcal S_h^{\rm reach}
=\{s_1-u:u\in\mathbb Z_+^n,\ \sum_i u_i\le h-1\}$.
Necessity follows from at most one sale per period. Conversely, each
such vector is attained by a feasible sequence of the indicated sales
and no-sales, all with positive probability. Adding the slack coordinate
$u_0=h-1-\sum_i u_i$ gives a bijection with nonnegative integer solutions
of $\sum_{i=0}^n u_i=h-1$, so
$|\mathcal S_h^{\rm reach}|=\binom{h-1+n}{n}$.
If every reachable terminal inventory has at least two admissible types,
then $\mathcal S_H^\star=\mathcal S_H^{\rm reach}$ and
\eqref{eq:sigma-decomposition} gives
$\sigma_\phi\le2/\binom{H-1+n}{n}$.
At $n=H=10$, the denominator is $\binom{19}{10}=92{,}378$.
This count identifies how many states share a unit probability mass;
it does not prescribe their individual probabilities.

\paragraph{Local effect of Fisher regularization.}
Fix exact reference values and expected scores. At
$z=(\theta_t,\phi_{t,k})$, the population ridged request direction is
$\{F_\phi(z)+\varepsilon_{\rm rid}I\}^{-1}
F_\phi(z)\Pi\bar V_t^\theta$, where $\bar V_t^\theta$ averages the exact
reference critic. At the saddle, $\Pi\bar V^\theta=0$, and
\eqref{eq:saddle-critic-differential} gives
$D(\Pi\bar V^\theta)(z^\star)[\delta z]=\alpha\delta\phi$.
Indeed, differentiating the allocation average contributes
$\sum_aD\mu_a\,Q_h^{z^\star}(s,a,b)=0$, because the saddle critic is
constant over feasible allocations and $\sum_aD\mu_a=0$.
The averaged allocation entropy derivative also vanishes; centering
the remaining request term leaves $\alpha\delta\phi$.

Derivatives of the current Fisher multiply a zero direction and
therefore vanish. Each inner step consequently has the same first-order
reference contribution, and the request block of the epoch derivative
is $I-\alpha K\eta
(F_\phi^\star+\varepsilon_{\rm rid}I)^{-1}F_\phi^\star$.
An equilibrium Fisher eigenvector with eigenvalue $x>0$ has factor
$1-\alpha K\eta x/(x+\varepsilon_{\rm rid})$.
Under the step condition these factors are nonnegative and decreasing
in $x$, so their maximum is
\begin{equation}
1-\alpha K\eta
\frac{\sigma_\phi}{\sigma_\phi+\varepsilon_{\rm rid}}.
\label{eq:ridged-request-factor}
\end{equation}
Allocation has the same formula with $\lambda$ and $F_\theta^\star$.
This calculation uses exact values and population scores; sampled
inverse errors are controlled separately.

\paragraph{Small probabilities even at equilibrium.}
Consider one resource, initial capacity $H$, $H$ periods, zero fares,
two identical request types, and unit accept/reject decisions. Both
choices remain feasible at every reachable stage. Backward induction
gives $V_h^\star=0$ and uniform policies: the allocation entropy bonus
$\alpha\log2$ and request entropy term $-\alpha\log2$ cancel.
The number of acceptances in the first $h-1$ periods is therefore
binomial, giving
\begin{equation}
d_h^{z^\star}(H-j)
=2^{-(h-1)}\binom{h-1}{j},\quad 0\le j<h,
\qquad \sigma_\phi=2^{-H}.
\label{eq:coverage-horizon-example}
\end{equation}
The extreme terminal inventories have probability $2^{-(H-1)}$.
The centered covariance eigenvalue for two equally likely types is
$1/2$, so the smallest request-Fisher eigenvalue is exactly $2^{-H}$.
All physical transitions are deterministic. Thus equilibrium coverage
constants in sufficient uniform-estimation bounds can be exponential
in the horizon, even though ideal natural-update factors contain no
visitation probability.

\subsection{Neural natural updates with a reference critic}
\label{app:neural-natural}\label{app:neural-errors}\label{app:bilateral-parameterized}
This subsection first derives operating-distribution certificates for
neural value and direction errors, then states what is additionally needed
for the uniform requirements of Theorem~\ref{thm:sampled}. The policies, baselines, and sampled scores
are those of Section~\ref{subsec:sampled}; their definitions are not repeated.
Write $w=(w^\theta,w^\phi)$ for the two actors' network parameters and
$\ell_j(w^j)$ for actor $j$'s centered reachable output logits.
The induced tabular pair is $z(w)=(\ell_\theta(w^\theta),
\ell_\phi(w^\phi))$. We abbreviate $Q^{z(w)}$, $V^{z(w)}$,
$\mathcal T_{\alpha,h}^{z(w)}$, and $\zeta_h^{z(w)}$ by
$Q^w$, $V^w$, $\mathcal T_{\alpha,h}^w$, and $\zeta_h^w$, respectively;
their definitions remain those of the induced policies.
All feasibility masks depend on the decision and not on the parameters.

\paragraph{Parameter directions and output directions.}
Write $D\ell_j$ for the output-logit Jacobian. The population
parameter scores and Fishers are
\begin{equation}
\begin{aligned}
\psi_h^{\theta,{\rm N}}&=\nabla_{w^\theta}\log\mu_{w^\theta,h},\qquad
\psi_h^{\phi,{\rm N}}=\nabla_{w^\phi}\log\nu_{w^\phi,h},\\
F_j^{\rm N}(w)&=\sum_h\mathbb E_w
[\psi_h^{j,{\rm N}}(\psi_h^{j,{\rm N}})^\top].
\end{aligned}
\label{eq:neural-parameter-fisher}
\end{equation}
Use $\bar S_j^{\rm N}(w;Q)=\sum_h\mathbb E_w[
\psi_h^{j,{\rm N}}Q_h(X_h)]$ for the parameter score mean.
If $F_j$ denotes the corresponding tabular Fisher on output logits,
the chain rule and \eqref{eq:frozen-score-identity} give
\[
F_j^{\rm N}=(D\ell_j)^\top F_jD\ell_j,\qquad
\bar S_j^{\rm N}(w;Q)=(D\ell_j)^\top F_jG^j.
\]
Consequently, when $F_j^{\rm N}$ is invertible, the induced first-order
output direction is
\[
D\ell_j(F_j^{\rm N})^{-1}\bar S_j^{\rm N}
=D\ell_j((D\ell_j)^\top F_jD\ell_j)^{-1}(D\ell_j)^\top F_jG^j.
\]
The matrix on the right is the $F_j$-orthogonal projection onto the range
of $D\ell_j$. Thus a neural natural step need not realize every conditional
tabular step. Its output discrepancy belongs to the actor accuracy
requirement of Theorem~\ref{thm:sampled}.
Redundant parameters can make $F_j^{\rm N}$ singular even at finite
logits. Statements involving its inverse are restricted to a fixed
identifiable coordinate system on which it is positive definite.

\begin{proposition}[Local natural updates in a parameterized class]
\label{prop:neural-natural}
Suppose the centered logit maps are twice continuously differentiable
near $w^\star$, their induced policies equal the full regularized saddle,
and $F_j^{\rm N}(w^\star)\succ0$ on the chosen identifiable coordinates.
Consider one exact-reference epoch: freeze $Q^{w_t}$, take $K$ request
steps and one allocation step using population parameter natural
directions. Put
$\varrho_K=\max\{1-\alpha\lambda,1-\alpha K\eta\}$. Its epoch map $\mathcal E_{\rm N}$ satisfies
\begin{equation}
D\mathcal E_{\rm N}(w^\star)
=\operatorname{diag}\{(1-\alpha\lambda)I,
                       (1-\alpha K\eta)I\}.
\label{eq:parameterized-bilateral-jacobian}
\end{equation}
In particular, if \eqref{eq:global-step-conditions} holds, then for every
$\varrho\in(\varrho_K,1)$ there is a neighborhood in which the exact-reference
epoch contracts Euclidean parameter error by at most $\varrho$.
\end{proposition}
\begin{proof}{Proof}
The parameter updates are
\begin{equation}
\begin{aligned}
w^\phi_{t,k+1}&=w^\phi_{t,k}-\eta
 [F_\phi^{\rm N}(w_{t,k})]^{-1}
 \bar S_\phi^{\rm N}(w_{t,k};Q^{w_t}),\\
w^\theta_{t+1}&=w^\theta_t+\lambda
 [F_\theta^{\rm N}(w_{t,K})]^{-1}
 \bar S_\theta^{\rm N}(w_{t,K};Q^{w_t}).
\end{aligned}
\label{eq:neural-natural-updates}
\end{equation}
where $w_{t,k}=(w^\theta_t,w^\phi_{t,k})$ and $w^\phi_{t+1,0}=w^\phi_{t,K}$.
At the saddle $Q_h^{w^\star}(s,a,b)=V_h^\star(s)$. For this fixed table,
the conditional zero-mean score identity implies
\[
\bar S_j^{\rm N}(w;Q^{w^\star})=0
\quad\text{for every current pair }w.
\]
Hence derivatives with respect to the current pair, and derivatives of
the inverse Fisher multiplying this zero score mean, contribute zero.
Only changes in the reference critic contribute to the first derivative.
Pulling back \eqref{eq:saddle-critic-differential} gives
\[
DQ_h^{w^\star}[\delta w^\theta,\delta w^\phi]
=-\alpha(\psi_h^{\theta,{\rm N}})^\top\delta w^\theta
 +\alpha(\psi_h^{\phi,{\rm N}})^\top\delta w^\phi.
\]
Allocation scores have zero conditional mean given $(s_h,b_h)$, so
$\mathbb E[\psi_h^{\theta,{\rm N}}
(\psi_h^{\phi,{\rm N}})^\top]=0$. Therefore
\[
\begin{aligned}
D_{\rm ref}\bar S_\theta^{\rm N}[\delta w^\theta,\delta w^\phi]
&=-\alpha F_\theta^{\rm N}(w^\star)\delta w^\theta,\\
D_{\rm ref}\bar S_\phi^{\rm N}[\delta w^\theta,\delta w^\phi]
&= \alpha F_\phi^{\rm N}(w^\star)\delta w^\phi.
\end{aligned}
\]
Every frozen-reference request step has this same first-order reference
contribution. Preconditioning, adding the $K$ request increments, and
adding the allocation increment proves
\eqref{eq:parameterized-bilateral-jacobian}.
The epoch map is continuously differentiable near $w^\star$ because the
Fishers remain positive definite there. Its derivative at the saddle
has Euclidean operator norm $\varrho_K$. Continuity gives a convex
neighborhood on which $\|D\mathcal E_{\rm N}\|_{\rm op}\le\varrho$; integrating
the derivative along the line to $w^\star$ proves the contraction claim.
\end{proof}
These are local claims under saddle representation and identifiability,
not consequences of finite logits alone. ReLU actors satisfy the stated
local smoothness condition only on neighborhoods where the relevant
activation patterns do not change; frozen encoders and smooth heads
provide another admissible class.

\paragraph{Accuracy of a backward critic fit.}
\begin{proposition}[Accuracy of a fitted target]
\label{prop:neural-regression}
At epoch $t$, condition on the reference pair $z=z_t$ and an independently fitted downstream
critic $\widehat Q_{h+1}$. Let $\mathcal Q_h$ be a class with
$|Q|\le Q_{\max,h}$, and draw $B_{Q,t}$ independent paths with labels
$Y_h=\widetilde R_{\alpha,h}+\widehat Q_{h+1}(X_{h+1})$ satisfying
$|Y_h|\le Q_{\max,h}$. Suppose $\widehat Q_h\in\mathcal Q_h$ has empirical
loss at most $\epsilon_{{\rm fit},h}$ above the class infimum in
\eqref{eq:critic-fit}. Write
\[
\epsilon_h=\|\widehat Q_h-\mathcal T_{\alpha,h}^{z}
\widehat Q_{h+1}\|_{L^2(\zeta_h^z)},\qquad
\epsilon_{{\rm rep},h}^2=\inf_{Q\in\mathcal Q_h}
\|Q-\mathcal T_{\alpha,h}^{z}\widehat Q_{h+1}\|_{L^2(\zeta_h^z)}^2.
\]
If $N_h(\varepsilon_{\rm cov})$ is the size of a supremum-norm
$\varepsilon_{\rm cov}$ cover with centers in $\mathcal Q_h$, then with
conditional probability at least $1-\delta_h$,
\begin{equation}
\epsilon_h^2\le\epsilon_{{\rm rep},h}^2+\epsilon_{{\rm fit},h}
+8Q_{\max,h}^2
 \sqrt{\frac{\log\{2N_h(\varepsilon_{\rm cov})/\delta_h\}}{2B_{Q,t}}}
+16Q_{\max,h}\varepsilon_{\rm cov}.
\label{eq:neural-regression}
\end{equation}
\end{proposition}
\begin{proof}{Proof}
Let $m_h(X_h)=\mathbb E[Y_h\mid X_h]
=\mathcal T_{\alpha,h}^{z}\widehat Q_{h+1}(X_h)$.
The loss decomposition in
Appendix~\ref{app:direction-estimation} identifies population excess
risk with $\|Q-m_h\|_{L^2(\zeta_h^z)}^2$. Squared losses lie in $[0,4Q_{\max,h}^2]$. Moreover,
$|(Q-Y_h)^2-(Q'-Y_h)^2|\le4Q_{\max,h}\|Q-Q'\|_\infty$.
Hoeffding's inequality \citep{hoeffding1963probability}, a union bound on
the cover, and approximation by a cover center give
\[
\sup_{Q\in\mathcal Q_h}|\mathcal L_h(Q)-\mathcal L_h^{\rm pop}(Q)|
\le4Q_{\max,h}^2
 \sqrt{\frac{\log\{2N_h(\varepsilon_{\rm cov})/\delta_h\}}{2B_{Q,t}}}
 +8Q_{\max,h}\varepsilon_{\rm cov}.
\]
The fitted predictor's population excess risk is at most its
representation error, its empirical fitting tolerance, and twice this
uniform deviation. Subtracting the common conditional-variance term
proves \eqref{eq:neural-regression}.
\end{proof}
Compact continuous critic classes have finite covers, including fixed
ReLU architectures with bounded parameters and inputs. Independent stage
batches permit conditional application at each backward fit, and their
failure probabilities add.

\begin{lemma}[Reference residuals control both natural directions]
\label{lem:neural-reference-natural-error}
Suppose $\widehat Q_{t,H+1}=0$ and
\begin{equation}
\|\widehat Q_{t,h}-\mathcal T_{\alpha,h}^{w_t}
\widehat Q_{t,h+1}\|_{L^2(\zeta_h^{w_t})}\le\epsilon_{t,h}.
\label{eq:main-neural-reference-residual}
\end{equation}
At an actor evaluation $w$, assume
$\zeta_h^w\ll\zeta_h^{w_t}$ and
$d\zeta_h^w/d\zeta_h^{w_t}\le\Lambda_{t,h}$. Define
\begin{equation}
E_{Q,t}^2=\sum_{h=1}^H\Lambda_{t,h}
\left(\sum_{\ell=h}^H\epsilon_{t,\ell}\right)^2.
\label{eq:neural-reference-error-budget}
\end{equation}
Then
\begin{equation}
\begin{aligned}
\|\widehat Q_{t,h}-Q_h^{w_t}\|_{L^2(\zeta_h^{w_t})}
&\le\sum_{\ell=h}^H\epsilon_{t,\ell},\\
\|\bar S_j^{\rm N}(w;\widehat Q_t)
-\bar S_j^{\rm N}(w;Q^{w_t})\|_{[F_j^{\rm N}(w)]^{-1}}
&\le E_{Q,t}.
\end{aligned}
\label{eq:neural-natural-critic-bias}
\end{equation}
The second claim uses identifiable coordinates with positive definite
current Fisher; the first needs no Fisher assumption or visitation floor.
\end{lemma}
\begin{proof}{Proof}
Put $e_h=\widehat Q_{t,h}-Q_h^{w_t}$ and let $\mathcal R_h$ be the
reference Bellman residual. The exact decomposition is
\[
e_h(X_h)=\mathcal R_h(X_h)
+\mathbb E_{w_t}[e_{h+1}(X_{h+1})\mid X_h].
\]
Conditional Jensen and Minkowski give
\[
\|e_h\|_{L^2(\zeta_h^{w_t})}
\le\epsilon_{t,h}+\|e_{h+1}\|_{L^2(\zeta_{h+1}^{w_t})}.
\]
Induction from $e_{H+1}=0$ proves the first claim. The density bound gives
\[
\|e_h\|_{L^2(\zeta_h^w)}^2
\le\Lambda_{t,h}(\sum_{\ell\ge h}\epsilon_{t,\ell})^2.
\]
For any parameter vector $a$, Cauchy--Schwarz over the stage-tuple measure
then yields
\[
\begin{aligned}
\left|a^\top\sum_h\mathbb E_w[\psi_h^{j,{\rm N}}e_h(X_h)]\right|
&\le\left\{\sum_h\mathbb E_w[
       (a^\top\psi_h^{j,{\rm N}})^2]\right\}^{1/2}
       \left\{\sum_h\|e_h\|_{L^2(\zeta_h^w)}^2\right\}^{1/2}\\
&\le\sqrt{a^\top F_j^{\rm N}(w)a}\,E_{Q,t}.
\end{aligned}
\]
Duality proves the second claim. For the type-averaged request advantage
in \eqref{eq:advantages}, averaging $e_h$ over the allocation policy
preserves this bound by conditional Jensen. The argument also permits
parameters shared across periods.
\end{proof}

If a separate type-value network is used in place of the exact conditional
mean $\bar V^\theta$, its discrepancy from
$\sum_a\mu_{t,h}(a\mid s,b)\widehat Q_{t,h}(s,a,b)$ is an additional actor
error. It is not certified by the action-value residual alone. For
example, a uniform discrepancy at most $\varepsilon_{\rm type}$ produces
a tabular centered request-direction error at most
$2\varepsilon_{\rm type}$ and parameter score error at most
$\sqrt H\varepsilon_{\rm type}$ in the inverse-Fisher norm, by the same
Cauchy--Schwarz argument.

For a finite tuple support with reference probabilities at least
$\zeta_0>0$, a uniform value certificate follows:
\begin{equation}
\|\widehat Q_t-Q^{w_t}\|_\infty
\le\zeta_0^{-1/2}\max_h\sum_{\ell=h}^H\epsilon_{t,\ell}.
\label{eq:neural-global-critic-certificate}
\end{equation}
Indeed, each pointwise squared error is bounded by its $L^2$ norm squared
divided by the tuple probability. This supplies the global theorem's
critic tolerance without saddle representation. The factor
$\zeta_0^{-1/2}$ is the cost of converting operating-distribution
accuracy into uniform accuracy; actor approximation and optimization
require separate certificates.

\paragraph{Distribution change during critic reuse.}
Within an epoch the allocation parameters stay fixed. Suppose request
has separate period parameters, and its period score norm is bounded
by $\Psi_\phi$ along the parameter paths from $w^\phi_{t,0}$ to $w^\phi$. Then
\begin{equation}
\Lambda_{t,h}\le
\exp\{\Psi_\phi\sqrt h\,\|w^\phi-w^\phi_{t,0}\|_2\}.
\label{eq:neural-reference-density-ratio}
\end{equation}
To see this, the path-prefix likelihood ratio through $X_h$ is
\[
\frac{d\mathbb P_w^{1:h}}{d\mathbb P_{w_t}^{1:h}}
=\prod_{\ell=1}^h
\frac{\nu_{w^\phi_\ell,\ell}(b_\ell\mid s_\ell)}
     {\nu_{w^\phi_{t,0,\ell},\ell}(b_\ell\mid s_\ell)}
\le\exp\left\{\Psi_\phi\sum_{\ell=1}^h
\|w^\phi_\ell-w^\phi_{t,0,\ell}\|_2\right\}.
\]
Each log ratio is the integral of the score along the corresponding
parameter path; allocation and physical transition factors cancel.
Cauchy--Schwarz gives the stated exponent. Marginalization preserves
the upper bound because
\[
\frac{d\zeta_h^w}{d\zeta_h^{w_t}}(X_h)
=\mathbb E_{w_t}\left[
\frac{d\mathbb P_w^{1:h}}{d\mathbb P_{w_t}^{1:h}}\,\middle|\,X_h\right].
\]
Thus $\Psi_\phi\sqrt H\max_{k\le K}\|w^\phi_{t,k}-w^\phi_{t,0}\|_2\le\log2$
suffices for every $\Lambda_{t,h}\le2$. For shared period parameters,
replace $\sqrt h$ by $h$. A verified policy likelihood-ratio bound gives
an alternative certificate. Clipping a PPO surrogate alone is not such
a bound: it does not constrain all policy ratios or guarantee a KL radius.

\begin{lemma}[A sampled parameter-Fisher solve]
\label{lem:neural-relative-fisher-solve}
At a fixed evaluation $w$, abbreviate $F=F_j^{\rm N}(w)\succ0$ and
$G_{\rm par}=F^{-1}\bar S_j^{\rm N}(w;\widehat Q_t)$.
Let $\widehat F$ and $\widehat S$ be the empirical parameter Fisher,
including its ridge, and the corresponding score mean. Suppose
\begin{equation}
\|F^{-1/2}(\widehat F-F)F^{-1/2}\|_{\rm op}\le\delta_F<1,
\label{eq:main-neural-relative-fisher}
\end{equation}
$\|\widehat S-\bar S_j^{\rm N}(w;\widehat Q_t)\|_{F^{-1}}
\le\varepsilon_{\rm sample}$, and a computed parameter direction
$\widehat G_{\rm par}$ has residual
$\|\widehat F\widehat G_{\rm par}-\widehat S\|_{F^{-1}}
\le\varepsilon_{\rm solve}$. Then
\begin{equation}
\|\widehat G_{\rm par}-G_{\rm par}\|_F
\le\frac{\varepsilon_{\rm sample}+\varepsilon_{\rm solve}
                 +\delta_F\|G_{\rm par}\|_F}{1-\delta_F}.
\label{eq:neural-natural-solve-error}
\end{equation}
The score and Fisher may use the same fresh batch. If a fitted value
network replaces an exact conditional average, its score discrepancy
must be added separately, as in Appendix~\ref{app:neural-refinement}.
\end{lemma}
\begin{proof}{Proof}
Put $A=F^{-1/2}\widehat FF^{-1/2}$ and
$r=\widehat F\widehat G_{\rm par}-\widehat S$. Exact algebra gives
\[
\begin{aligned}
F^{1/2}(\widehat G_{\rm par}-G_{\rm par})
=A^{-1}\bigl\{&F^{-1/2}
 [\widehat S-\bar S_j^{\rm N}(w;\widehat Q_t)+r]\\
 &-(A-I)F^{1/2}G_{\rm par}\bigr\}.
\end{aligned}
\]
The relative covariance condition implies
$(1-\delta_F)I\preceq A\preceq(1+\delta_F)I$ and
$\|A^{-1}\|_{\rm op}\le(1-\delta_F)^{-1}$. Taking norms proves the bound.
It targets the same frozen fitted critic as the main-text ideal direction;
reference evaluation error is controlled separately by
Lemma~\ref{lem:neural-reference-natural-error}, which gives
\[
\|F^{-1}\bar S_j^{\rm N}(w;\widehat Q_t)
 -F^{-1}\bar S_j^{\rm N}(w;Q^{w_t})\|_F\le E_{Q,t}.
\]
\end{proof}

For a finite neural step, let $s_\theta=\lambda$, $s_\phi=-\eta$, and
$\widehat G_{\rm par}^j$ denote its computed parameter direction. The
realized direction used in the main text is exactly
\[
\widehat G^j
=\frac{\ell_j(w^j+s_j\widehat G_{\rm par}^j)-\ell_j(w^j)}{s_j}.
\]
If $\|D^2\ell_j\|_{2,2\to\infty}\le L_\ell$ on the parameter step,
Taylor's inequality yields the explicit output certificate
\begin{equation}
\begin{aligned}
\|\widehat G^j-G^j\|_\infty
\le{}&\underbrace{\|D\ell_jG_{\rm par}^j-G^j\|_\infty}
 _{\text{class projection error}}\\
&+\|D\ell_jF^{-1/2}\|_{2\to\infty}
       \|\widehat G_{\rm par}^j-G_{\rm par}^j\|_F
 +\frac{|s_j|L_\ell}{2}\|\widehat G_{\rm par}^j\|_2^2.
\end{aligned}
\label{eq:neural-output-direction-certificate}
\end{equation}
The three terms are class restriction, estimation/solve error, and finite-step
curvature. They concern realizing the fitted-critic direction $G^j$;
the reference value tolerance $\varepsilon_Q$ remains a separate requirement.
A PPO update must also be checked against the conditional KL target; its
surrogate loss alone does not certify these output increments.

\paragraph{Sufficient sampling bounds.}
Apply Proposition~\ref{prop:neural-regression} conditionally to each
independent backward fit and substitute its residual bounds into
\eqref{eq:neural-reference-error-budget}. For actor sampling, suppose
$\|F^{-1/2}\psi_h^{j,{\rm N}}\|_2\le\kappa_h$, and critic values and
baselines are bounded by $Q_{\max,h}$. Put
$L_S=2\sum_h\kappa_hQ_{\max,h}$ and $M_F=\sum_h\kappa_h^2$.
Condition on the current policy, fitted critic, and fitted baseline
networks. Draw $B$ fresh current-policy paths independently of their
fitting data. With exact conditional type averages, bounded-vector
concentration gives
\begin{equation}
\begin{aligned}
\Pr\{\|\widehat S-\bar S_j^{\rm N}(w;\widehat Q_t)\|_{F^{-1}}
>\varepsilon_{\rm sc}\mid\text{past}\}
&\le2\exp\{-B\varepsilon_{\rm sc}^2/(8L_S^2)\},\\
\Pr\{\|F^{-1/2}(\widehat F-\varepsilon_{\rm rid}I-F)
F^{-1/2}\|_{\rm op}>\varepsilon_F\mid\text{past}\}
&\le2\exp\{-B\varepsilon_F^2/(8M_F^2)\}.
\end{aligned}
\label{eq:neural-whitened-sample-bounds}
\end{equation}
For completeness, let independent Hilbert-space vectors satisfy
$\|X_i\|\le R_X$. Their centered mean has expected norm at most
$R_X/\sqrt B$, and replacing one path changes its norm by at most
$2R_X/B$. The bounded-differences inequality therefore gives, for
$\varepsilon\ge2R_X/\sqrt B$,
\[
\Pr\left\{\left\|B^{-1}\sum_i(X_i-\mathbb EX_i)\right\|
>\varepsilon\right\}
\le\exp\{-B\varepsilon^2/(8R_X^2)\}.
\]
For smaller $\varepsilon$, the asserted upper bound with prefactor two
exceeds one. Apply this with $R_X=L_S$ to whitened path scores. For the
second line, use the Hilbert space of matrices with Frobenius norm and
$R_X=M_F$; Frobenius norm
dominates operator norm. The ridge contributes at most
$\varepsilon_{\rm rid}/\lambda_{\min}(F)$ to the full relative error.
With separate period parameters, orthogonality sharpens the bounds to
$L_S=2(\sum_h\kappa_h^2Q_{\max,h}^2)^{1/2}$ and
$M_F=(\sum_h\kappa_h^4)^{1/2}$. Conditional failure budgets accommodate
adaptive policy evaluations. These leverage bounds can deteriorate when
the parameter Fisher is poorly conditioned.

The preceding reference-critic bounds control error relative to the
frozen epoch-start value. They do not automatically bound a critic's
residual against a later current policy; the fixed-opponent refinement
result below assumes the latter residual separately.

\subsection{Refinement against a fixed opponent}
\label{app:neural-refinement}
Freeze the opponent and minimize
$J_{\rm ref}(w)=J_\alpha(\bar\theta,w)$ for request, or
$J_{\rm ref}(w)=-J_\alpha(w,\bar\phi)$ for allocation.
Here $w$ denotes only the refined actor's parameters, and $W$ is its compact
convex feasible parameter set. Use the unpreconditioned
score estimate of \eqref{eq:advantages}, with the allocation sign reversed,
and let
\begin{equation}
w_{t+1}=\operatorname{proj}_W(w_t-\eta\widehat S_t),\qquad
\mathcal G_\eta(w)=\eta^{-1}
\{w-\operatorname{proj}_W(w-\eta\nabla J_{\rm ref}(w))\}.
\label{eq:neural-refinement-update}
\end{equation}
Assume current-pair residuals satisfy
\eqref{eq:main-neural-reference-residual} under $\zeta_h^{z_t}$,
stage scores have norm at most $\Psi_j$, the fresh-batch score error
is at most $\varepsilon_{{\rm sample},t}$, and any additional
population score error from the fitted type-value network is at most
$\varepsilon_{{\rm value},t}$. Then
\begin{equation}
\|\widehat S_t-\nabla J_{\rm ref}(w_t)\|_2
\le\varepsilon_{{\rm grad},t}
\coloneqq \Psi_j\sum_{h=1}^Hh\epsilon_{t,h}
+\varepsilon_{{\rm sample},t}+\varepsilon_{{\rm value},t}.
\label{eq:neural-response-gradient-error}
\end{equation}
In fact, Lemma~\ref{lem:neural-reference-natural-error} at $z_t$ gives
$\|\widehat Q_{t,h}-Q_h^{z_t}\|_{L^2(\zeta_h^{z_t})}
\le\sum_{\ell\ge h}\epsilon_{t,\ell}$. Cauchy--Schwarz bounds the score
bias by
\[
\sum_h\Psi_j\|\widehat Q_{t,h}-Q_h^{z_t}\|_{L^2(\zeta_h^{z_t})}
\le\Psi_j\sum_{\ell=1}^H\ell\epsilon_{t,\ell}.
\]
For separate period parameters, orthogonality improves this to
$\Psi_j\{\sum_h(\sum_{\ell\ge h}\epsilon_{t,\ell})^2\}^{1/2}$.
The fresh-path error is measured around the population score computed
with the fitted critic and actual value network. If the type-value network
replaces the exact allocation average, let $d^{\rm value}_{t,h}(s,b)$
be their difference. Its additional request-score bias is bounded by
$\Psi_j\sum_h\|d^{\rm value}_{t,h}\|_{L^2(d_h^{z_t}\nu_h^{z_t})}$.
An allocation baseline depending only on $(s,b)$ has zero population
score contribution. Exact conditional type averages give
$\varepsilon_{{\rm value},t}=0$. A reused critic must satisfy these
current-pair residual conditions; reference-pair accuracy alone does not
imply them.

\begin{proposition}[Stationarity of neural response refinement]
\label{thm:neural-response-stationarity}\label{cor:fa}
Fix the opponent. Suppose $\nabla J_{\rm ref}$ is
$L_{\rm ref}$-Lipschitz on a neighborhood of the compact convex set $W$,
stage scores are bounded by $\Psi_j$, and use
\eqref{eq:neural-refinement-update} from any $w_0\in W$ with
$0<\eta\le L_{\rm ref}^{-1}$. If the current-pair residual and fresh-path
sampling bounds, and the stated value-network score-error bound hold
simultaneously for $t=0,\ldots,T-1$ with probability at least
$1-\delta$, then on that event
\begin{equation}
\frac1T\sum_{t=0}^{T-1}\|\mathcal G_\eta(w_t)\|_2^2
\le\frac{8\{J_{\rm ref}(w_0)-\min_WJ_{\rm ref}\}}{\eta T}
+\frac{10}{T}\sum_{t=0}^{T-1}\varepsilon_{{\rm grad},t}^2.
\label{eq:neural-response-stationarity}
\end{equation}
\end{proposition}
\begin{proof}{Proof}
Put $\Delta w_t=w_{t+1}-w_t$. Projection optimality and feasibility of
$w_t$ give
$\langle\widehat S_t,\Delta w_t\rangle
\le-\|\Delta w_t\|_2^2/\eta$.
Smoothness, \eqref{eq:neural-response-gradient-error}, and
$\eta L_{\rm ref}\le1$ imply
\[
\begin{aligned}
J_{\rm ref}(w_{t+1})-J_{\rm ref}(w_t)
&\le-\frac{\|\Delta w_t\|_2^2}{2\eta}
 +\varepsilon_{{\rm grad},t}\|\Delta w_t\|_2\\
&\le-\frac{\|\Delta w_t\|_2^2}{4\eta}
 +\eta\varepsilon_{{\rm grad},t}^2.
\end{aligned}
\]
Nonexpansiveness of projection also gives
\[
\|\mathcal G_\eta(w_t)\|_2
\le\frac{\|\Delta w_t\|_2}{\eta}
 +\varepsilon_{{\rm grad},t}.
\]
Squaring, summing, and using the preceding descent inequality yields
\[
\begin{aligned}
\frac1T\sum_{t<T}\|\mathcal G_\eta(w_t)\|_2^2
&\le\frac2{\eta^2T}\sum_{t<T}\|\Delta w_t\|_2^2
 +\frac2T\sum_{t<T}\varepsilon_{{\rm grad},t}^2\\
&\le\frac{8\{J_{\rm ref}(w_0)-J_{\rm ref}(w_T)\}}{\eta T}
 +\frac{10}{T}\sum_{t<T}\varepsilon_{{\rm grad},t}^2.
\end{aligned}
\]
Finally $J_{\rm ref}(w_T)\ge\min_WJ_{\rm ref}$.
\end{proof}
Frozen encoders with bounded linear softmax heads satisfy the score and
smoothness requirements; smooth full networks are also admissible when
these bounds hold. Dropout is disabled during analyzed updates so the
sampled and differentiated policies agree. This stationarity certificate
concerns a fixed opponent and does not certify a globally worst response.

\subsection{Shared-feature sampling}
\label{app:neural-shared-sampling}

\paragraph{Proof of the shared-feature sampling bound.}
If $M=0$, the claim is immediate; suppose $M>0$.
Condition on the reference policies, the independently fitted
continuation, and an encoder $g$ fixed in advance or fitted on data
independent of the current regression paths. Let $n_g$ be the group count,
$p_g=\Pr\{g(X)=g\}$, and $\bar m_g=\mathbb E[Y\mid g(X)=g]$.
The $B$ paths are independent. A binomial lower-tail bound gives
\[
\Pr\{n_g<Bp_g/2\}\le\exp(-Bp_g/8).
\]
Conditional on the group indicators, the labels in a group are
independent, lie in $[-M,M]$, and have mean $\bar m_g$. On the count event,
Hoeffding's inequality therefore gives
\[
\Pr\{|\widehat m_g-\bar m_g|>u\mid\text{group indicators}\}
\le2\exp\{-Bp_{\min}u^2/(4M^2)\}.
\]
With $L=\log(4G/\delta)$, the batch in
\eqref{eq:shared-group-sampling} bounds the union failure probability by
$Ge^{-L}+2Ge^{-L}=3\delta/4\le\delta$.
On the complementary event, all groups are observed and
\[
|\widehat Q_h(x)-m(x)|
\le|\widehat m_{g(x)}-\bar m_{g(x)}|
  +|\bar m_{g(x)}-m(x)|\le u+a_{\rm rep}.
\]
Empty groups can be assigned zero before the event is imposed.
If every fitted downstream continuation and encoder is independent of the
data used for the current stage and satisfies the grouping assumption,
the same argument applies conditionally at each stage. For $H$ backward
fits, take failure budgets $\delta_h$ summing to
$\delta$. Conditional application at each independent stage fit gives
uniform Bellman residual tolerances $a_{{\rm rep},h}+u_h$, and
Lemma~\ref{lem:critic-drift} yields
\[
\|\widehat Q_{t,h}-Q_h^{w_t}\|_\infty
\le\sum_{\ell=h}^H(a_{{\rm rep},\ell}+u_\ell).
\]
This example concerns a critic with a frozen shared encoder and a fitted
linear head. It does not assert target grouping for arbitrary learned
encoders or certify actor accuracy by itself.

\end{document}